\documentclass{article}

\usepackage{microtype}
\usepackage{graphicx}
\usepackage{subcaption}
\usepackage{booktabs} 

\usepackage{hyperref}

\usepackage[preprint]{icml2026}

\usepackage{amsmath}
\usepackage{amssymb}
\usepackage{mathtools}
\usepackage{amsthm}

\usepackage[capitalize,noabbrev]{cleveref}

\theoremstyle{plain}

\theoremstyle{definition}

\theoremstyle{remark}

\usepackage[textsize=tiny]{todonotes}

\usepackage{url}

\usepackage[utf8]{inputenc} 
\usepackage[T1]{fontenc}    
\usepackage{booktabs}       
\usepackage{amsfonts}       
\usepackage{nicefrac}       
\usepackage{microtype}      
\usepackage[table,dvipsnames]{xcolor}

\usepackage{amsmath}
\usepackage{amsfonts}
\usepackage{amssymb}
\usepackage{amsthm}
\usepackage[most]{tcolorbox}
\usepackage{graphicx}
\usepackage{subcaption}
\usepackage{enumitem}
\usepackage{array}
\usepackage{multirow}
\usepackage{tabularx}
\usepackage{arydshln}
\usepackage{wrapfig}
\usepackage{makecell}
\usepackage{colortbl}
\definecolor{softred}{RGB}{253,236,233}

\usepackage{titletoc}

\icmltitlerunning{Agentic RL via Fine-Grained Reward Structure and Credit Assignment}

\begin{document}

\twocolumn[
  \icmltitle{Reinforcing Multi-Turn Reasoning in LLM Agents via Fine-Grained \\Reward Structure and Credit Assignment}



  \icmlsetsymbol{equal}{*}
    \begin{icmlauthorlist}
    \icmlauthor{Quan Wei}{equal,umn}
    \icmlauthor{Siliang Zeng}{equal,umn}
    \icmlauthor{Chenliang Li}{tamu}
    \icmlauthor{Zhongruo Wang}{ind}
    \icmlauthor{William Brown}{pi}
    \icmlauthor{Oana Frunza}{ms}
    \icmlauthor{Wei Deng}{ms}
    \icmlauthor{Anderson Schneider}{ms}
    \icmlauthor{Yuriy Nevmyvaka}{ms}
    \icmlauthor{Yang (Katie) Zhao}{umn}
    \icmlauthor{Alfredo Garcia}{tamu}
    \icmlauthor{Mingyi Hong}{umn}
    
    \end{icmlauthorlist}
    
    \icmlaffiliation{umn}{University of Minnesota, USA}
    \icmlaffiliation{tamu}{Texas A\&M University, USA}
    \icmlaffiliation{ind}{Independent Researcher}
    \icmlaffiliation{pi}{Prime Intellect, USA}
    \icmlaffiliation{ms}{Morgan Stanley, USA}
    
    \icmlcorrespondingauthor{Mingyi Hong}{mhong@umn.edu}

  \icmlkeywords{Machine Learning, ICML}

  \vskip 0.3in
]



\printAffiliationsAndNotice{}  

\begin{abstract}

Reinforcement Learning (RL) approaches have been wildly used to enhance the reasoning capabilities of Large Language Model (LLM) agents in long-horizon, multi-turn scenarios.
Such interactions can be formalized as turn-level Markov decision processes (MDPs), where intermediate rewards are often available.
However, most prior work relies on sparse trajectory-level rewards, resulting in poor credit assignment, while dense turn-level rewards remain underexplored.
In this paper, we investigate how to effectively leverage dense turn-level reward structures in RL algorithms, specifically Group Relative Policy Optimization (GRPO) and Proximal Policy Optimization (PPO), to enable fine-grained credit assignment.
We categorize reward structures into three types based on their granularity: (1) terminal reward; (2) delayed reward; (3) per-turn reward, each corresponding to a distinct turn-level MDP formulation, and derive GRPO and PPO algorithms tailored to each case, respectively.
Experiments on multi-turn search and game agent tasks show that, for both GRPO and PPO, dense per-turn reward structures consistently outperform sparse terminal and delayed reward structures in terms of training dynamics and numerical results.
Furthermore, on search tasks, PPO with dense per-turn rewards achieves greater training stability and faster convergence, and delivers the highest answer correctness among all state-of-the-art methods across diverse question-answering datasets.

\end{abstract}

\section{Introduction}

Reinforcement Learning (RL) has recently emerged as a powerful approach for improving the reasoning capabilities of Large Language Models (LLMs), allowing them to explore and refine long Chains of Thought (CoT) \citep{wei2022chain} in complex decision-making tasks. Building on this paradigm, reasoning-based LLMs, such as OpenAI's o1 \citep{jaech2024openai} and DeepSeek's R1 \citep{guo2025deepseek}, demonstrate remarkable performance in textual reasoning tasks by learning analytical thinking and self-reflection.
Despite these advancements, LLMs that rely solely on textual reasoning remain limited in tasks that require precise and complex numerical computation, information retrieval from web pages or local databases, or code execution. 
Equipping LLMs with access to external tools, such as search engines, scientific calculators, or code interpreters, as autonomous agents for multi-turn interactions can significantly extend their capabilities beyond pure text-based reasoning \citep{gou2023tora}.

Training LLMs to operate as autonomous agents in interactive environments presents challenges that extend beyond single-turn inference.
In these settings, models must perform sequential decision-making, generating actions across multiple interaction turns to solve complex reasoning tasks.
This sequential structure naturally situates LLM agents in multi-turn, long-horizon regimes, which can be formally modeled as turn-level Markov decision processes (MDPs).
Within this framework, two types of rewards are commonly considered: the outcome (terminal) reward, which reflects the correctness of the final answer, and intermediate rewards, which correspond to the correctness of individual turns and are available at each interaction step with the environment.
To optimize LLM agents under this MDP framework, RL algorithms, including Group Relative Policy Optimization (GRPO) \citep{shao2024deepseekmath} and Proximal Policy Optimization (PPO) \citep{schulman2017proximal}, are commonly employed.
GRPO leverages group-based normalization for advantage estimation, whereas PPO relies on a learned critic to compute advantages via Generalized Advantage Estimation (GAE) \citep{schulman2015high}.

One line of research \citep{chen2025learning, jin2025search,feng2025retool} formulate multi-turn interactive tasks as single-turn problems, relying solely on final outcome-level rewards such as answer correctness.
However, such single-turn formulation is inadequate for long-horizon multi-turn reasoning as it treats the entire trajectory as a single decision step, ignoring the multi-turn structure of the interactive tasks. In particular, it ignores intermediate signals that evaluate each complete agent-environment interaction, such as a tool call and its result, providing feedback at the granularity of a single turn in multi-turn tasks \citep{lightman2023let,zhang2025lessons,ma2023let,choudhury2025process}. 
Without access to dense turn-level feedback, agents struggle to refine their behavior, making it difficult to interact effectively with dynamic environments over multiple steps.
For example, in a search agent \citep{chen2025learning,jin2025empirical}, selecting a good query early on is crucial for retrieving relevant information; without turn-level retrieval feedback, the agent may not learn which queries contribute to correct answers.

Another line of research \citep{li2025torl,qian2025toolrl,wang2025otc,bespoke_improving_multi_turn_tool_use,wang2025ragen,singh2025agentic,zhang2025nemotron,jin2025empirical} incorporate intermediate rewards, such as signals derived from tool execution.
However, these approaches suffer from a credit assignment problem, as outcome-level and intermediate rewards are merged into a trajectory-level training signal for reinforcement learning.
This reward aggregation leads to inaccurate advantage estimation and limits the ability of RL algorithms to provide fine-grained supervision at intermediate interaction steps, ultimately hindering effective learning in multi-turn, long-horizon settings \citep{guo2025segment, feng2025group, zhang2025rlvmr}.

Overall, most prior work relies on sparse trajectory-level rewards, resulting in poor credit assignment, while dense turn-level rewards remain underexplored.
This observation raises a fundamental question:
\begin{center}
\textit{How can dense turn-level reward structures be effectively leveraged within RL algorithms to enable fine-grained credit assignment?}
\end{center}
Motivated by this question, we investigate the use of dense turn-level rewards in this paper.
Our key contributions are summarized as follows:
\begin{itemize}[itemsep=0mm, topsep=0mm]
    \item We categorize reward structures into three types based on their granularity: (1) terminal reward; (2) delayed reward; (3) per-turn reward, each corresponding to a distinct turn-level MDP formulation. We derive GRPO and PPO algorithms tailored to each case, respectively.
    \item For GRPO, we introduce a novel variant based on per-turn rollouts.
    The resulting advantage function enables credit assignment at both global and local scales by aggregating current and future advantages.
    Experiments across three GRPO-based algorithms, each corresponding to a different MDP formulation, demonstrate that dense per-turn reward structures consistently outperform sparse terminal and delayed rewards, both in terms of training dynamics and final numerical performance.
    \item For PPO, we present the reward assignment strategy within GAE for three MDP cases. Experiments on multi-turn search and game agent tasks demonstrate that dense per-turn reward structures consistently outperform sparse terminal and delayed rewards, both in terms of training dynamics and numerical results, with PPO trained on dense per-turn rewards exhibiting greater training stability and faster convergence. Furthermore, benchmarks on both in-domain and out-of-domain search tasks demonstrates PPO with dense per-turn rewards derived from LLM-based judges achieves the highest answer correctness among all state-of-the-art methods. These findings verify that dense turn-level reward structure and fined-grained credit assignment can improve the performance of LLM agents. 
\end{itemize}

\section{Related Work}

\subsection{Process Reward and Credit Assignment in RL}

Process rewards provide fine-grained credit assignment and enhance both training efficiency and optimization stability in RL.
Such dense rewards have proven effective in classical RL domains such as games and robotic control~\citep{schrittwieser2020mastering, liu2022meta, sun2025large}.
Process reward models (PRMs) have also been extensively explored for inference-time scaling in LLMs~\citep{lightman2023let, uesato2022solving}.
Recent studies have further highlighted the importance of effective credit assignment in RL~\citep{pignatelli2023survey, shao2024deepseekmath, cui2025process, cheng2025stop, feng2025group, guo2025segment}, particularly for textual reasoning tasks such as mathematical problem solving.
In multi-turn agent interaction settings, turn-level rewards evaluate each complete agent–environment interaction, such as a tool invocation and its resulting output, providing feedback at the granularity of a single turn.
This setting naturally emphasizes the advantages of process-level rewards and fine-grained credit assignment.
However, dense turn-level reward structures and the corresponding RL algorithms for multi-turn settings remain underexplored.

\subsection{RL for LLM Agents}

RL has been applied to train long-horizon multi-turn LLM agents in diverse domains, including search \citep{chen2025learning, jin2025search, jin2025empirical}, tool use \citep{feng2025retool, li2025torl, qian2025toolrl, wang2025otc, bespoke_improving_multi_turn_tool_use, zhang2025nemotron, singh2025agentic}, text-based games \citep{yao2020keep, carta2023grounding, zhai2024fine, wang2025ragen}, web shopping \citep{yao2022webshop}, digital app interaction \citep{chen2025reinforcement}, and mobile device control \citep{bai2024digirl}.
A number of these studies \citep{jin2025empirical, feng2025retool, li2025torl, qian2025toolrl, wang2025otc, bespoke_improving_multi_turn_tool_use, zhang2025nemotron, singh2025agentic} apply RL algorithms such as GRPO and PPO to train tool-using LLM agents, including calculators, code interpreters, and search engines, thus enabling reasoning with external tools. 
However, these methods generally use terminal rewards or collapse outcome and intermediate signals into a single trajectory-level reward, limiting fine-grained credit assignment.
The most closely related work includes StepSearch \citep{wang2025stepsearch} and PPR \citep{xu2025hybrid}, both of which apply PPO with turn-level rewards for multi-turn search tasks. Our work differs in that we focus on a systematic comparison of different reward structures, which is not explored in detail in these prior studies.
In particular, StepSearch uses information gain as the reward signal, while PPR relies on trainable reward models. In contrast, our approach employs an LLM-based judge as a PRM.

\section{Problem Setup for Multi-Turn Tasks}

\subsection{Multi-Turn Rollout as Turn-Level MDP}

\textbf{Multi-Turn Rollout.}
Let $x$ denote the input prompt sampled from the dataset $\mathcal{D}$.
A multi-turn rollout produces a complete response 
$y = [l_1,f_1,\ldots,l_K,f_K]$,
where $l_k$ is the response generated from an LLM policy $\pi_\theta$, and $f_k$ is the corresponding environment feedback at the $k$-th turn.

\textbf{Turn-Level MDP.}
LLM agents operate in interactive environments where each turn yields stochastic feedback.
To capture these dynamics, we formulate the multi-turn rollout as a turn-level MDP, which is formally defined as $\mathcal{M} = \{ \mathcal{S}, \mathcal{A},P, R, \gamma\}$.
Here, $\mathcal{S}$ denotes the state space, and $\mathcal{A}$ denotes the action space; A state $s\in \mathcal{S}$ typically corresponds to an interaction history, while an action $a\in \mathcal{A}$ often corresponds to a sequence of generated tokens; $P$ represents the transition dynamics; $R$ is the turn-level reward function; $\gamma$ is the discount factor. 
At the $k$-th turn, conditioned on the current state $s_k$, the agent makes an action $a_k$ according to the policy $\pi_\theta$, where $a_k = [l_k,f_k]$ if environment feedback exists, otherwise $a_k=l_k$. The agent then receives a turn-level reward $R_k=R(s_k,a_k)$, and transitions to the next state $s_{k+1}$. 
We formalize this process as follows:
$a_k \sim \pi_{\theta}(\cdot \mid s_k)$, $s_{k+1} \sim P(\cdot \mid s_k,a_k)$.
A multi-turn rollout trajectory is $\tau = \{ (s_1,a_1,R_1), (s_2,a_2,R_2),\ldots,(s_K,a_K,R_K)\}$.
Note that the outcome reward is denoted by $R(x,y)=R_K$ for a prompt--response pair $(x,y)$.
Under the turn-level MDP framework, the objective can be written as maximizing the cumulative reward at the turn level:
\begin{equation}
\max_{\pi_\theta} \;
\mathbb{E}_{\tau \sim \pi_\theta}
\left[\sum_{k=1}^{K} \gamma^k R(s_k,a_k)\right]
\label{eq:mdp_objective}
\end{equation}

\subsection{Turn-Level Reward Assignment}
\label{sec:reward-assignment}
\begin{figure}
    \centering
    \includegraphics[width=1\linewidth]{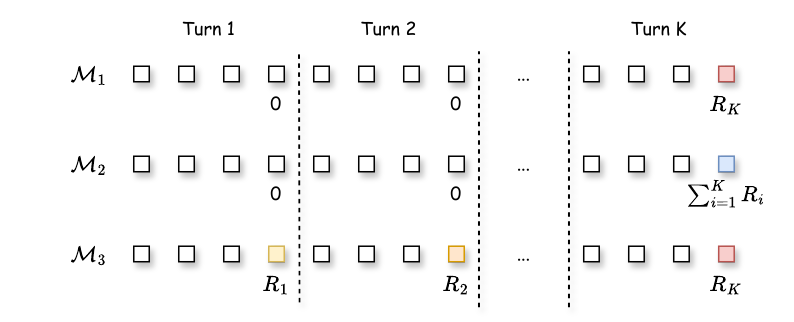}
    \caption{Comparison of three turn-level reward assignment strategies for multi-turn tasks.
    Squares represent tokens; color indicates whether the assigned reward is zero or nonzero.
    Turn-level rewards are assigned to the final token of each turn.}
    \label{fig:reward-assignment}
\end{figure}

In multi-turn tasks, the outcome reward typically corresponds to the correctness of the final answer, while intermediate rewards are often available from each round of interaction with the environment.
Based on the granularity of reward assignment, we categorize multi-turn formulations into three types of turn-level MDPs:
\begin{enumerate}[itemsep=0mm, topsep=0mm]
    \item Turn-level MDP with a \textbf{terminal} reward $\mathcal{M}_1$: provides only a final outcome (terminal) reward with no intermediate rewards.
    \item Turn-level MDP with a \textbf{delayed} reward $\mathcal{M}_2$: provides an accumulated reward that merges both intermediate and outcome rewards into a single delayed signal.
    \item Turn-level MDP with explicit \textbf{per-turn} rewards $\mathcal{M}_3$: provides explicit rewards at each turn.
\end{enumerate}
Here, $\mathcal{M}_1$ contains only \textit{outcome} rewards, whereas both $\mathcal{M}_2$ and $\mathcal{M}_3$ include \textit{intermediate} rewards but differ in how these rewards are distributed across the turns.
Moreover, $\mathcal{M}_1$ and $\mathcal{M}_2$ provide \textit{trajectory-level} rewards, whereas $\mathcal{M}_3$ provides explicit \textit{turn-level} rewards.
For $\mathcal{M}_1$, the objective reduces to maximizing the final outcome reward.
$\mathcal{M}_2$ and $\mathcal{M}_3$ are return-equivalent.
Figure~\ref{fig:reward-assignment} compares three turn-level reward assignment strategies for multi-turn tasks.

In the following section, we investigate how to effectively leverage dense turn-level reward structures in RL algorithms, specifically GPRO and PPO, to enable fine-grained credit assignment.
Table~\ref{tab:algo} summarizes the main RL algorithms discussed in this paper under three different MDP formulations.

\begin{table}
    \centering
    \caption{RL algorithms discussed in this paper under different MDP formulations.}
    \label{tab:algo}
    \resizebox{\linewidth}{!}{%
    \begin{tabular}{lccc}
        \toprule
        \textbf{Reward Structure}
        & \textbf{GPRO}
        & \textbf{PPO}
        & \makecell{\textbf{Turn-Decomposed} \\ \textbf{PPO}} \\
        \midrule
        Terminal reward $\mathcal{M}_1$
            & $\checkmark$ & $\checkmark$ & $\checkmark$ \\
        Delayed reward $\mathcal{M}_2$
            & $\checkmark$ & $\checkmark$ & $\checkmark$ \\
        Per-turn reward $\mathcal{M}_3$
            & $\checkmark$ & $\checkmark$ & $\checkmark$ \\
        \bottomrule
    \end{tabular}%
    }
\end{table}

\section{Credit Assignment for GRPO-Type Algorithms in Multi-Turn Tasks}
\label{sec:grpo}

\subsection{Preliminaries: GPRO}

\textbf{GRPO.}
The Group Relative Policy Optimization (GRPO) algorithm \citep{shao2024deepseekmath} has been widely used to enhance the reasoning capabilities of LLMs, which estimates the advantage in a group-relative manner.
Specifically, for each input question $x$, it samples a group of responses  $\{y_1, y_2, \ldots, y_G\}$ from the reference policy $\pi_{\text{ref}}$.
GRPO optimizes the policy by maximizing the following objective function:
\begin{align*}
\mathbb{E}_{x \sim \mathcal{D},\, \{y_i\}_{i=1}^G \sim \pi_{\text{old}}(\cdot \mid x)}
\Bigg[
\frac{1}{G} \sum_{i=1}^G \frac{1}{|y_i|}
\sum_{t=1}^{|y_i|}
\min\Big(
w_{i,t}(\theta) A_{i,t}, \\
\text{clip}\!\big(
w_{i,t}(\theta), 1-\epsilon, 1+\epsilon
\big) A_{i,t}
\Big)
- \beta\, \mathbb{D}_{\text{KL}}\!\left[
\pi_\theta \,\|\, \pi_{\text{ref}}
\right]
\Bigg]
\end{align*}
where $w_{i,t}(\theta)=\frac{\pi_\theta(y_{i,t} \mid x, y_{i,<t})}{\pi_{\text{old}}(y_{i,t} \mid x, y_{i,<t})}$ is the token-level importance sampling ratio between the current policy $\pi_\theta$ and the previous policy $\pi_{\text{old}}$, $\epsilon$ is the clipping parameter, and $\beta$ is the KL divergence coefficient.
Given a group of trajectory-level rewards $\{R^{\text{traj}}_i\}_{i=1}^G$, the advantage of the $i$-th response $A_{i,t}$ is calculated by
\begin{equation}
A_{i,t} = A^{\text{GRPO}}_{i} = \frac{R^{\text{traj}}_i - \text{mean}(\{R^{\text{traj}}_i\}_{i=1}^G)}{\text{std}(\{R^{\text{traj}}_i\}_{i=1}^G)}
\label{eq:grpo-adv}
\end{equation}
GRPO is naturally suited for MDPs with trajectory-level rewards, namely $\mathcal{M}_1$ and $\mathcal{M}_2$;
for $\mathcal{M}_1$, $R^{\text{traj}}_i=R(x,y_i)$,
and for $\mathcal{M}_2$, $R^{\text{traj}}_i = \sum_{k=1}^K \gamma^k R_{i,k}$ 
with $R_{i,k}=R(s_{i,k}, a_{i,k})$ denoting the intermediate reward given the state $s_{i,k}$ and action $a_{i,k}$ at the $k$-th turn.

\textbf{Limitations of GRPO in Multi-Turn Settings.}
For GPRO, the advantage function $A_{i,t}$ in Eq.~\eqref{eq:grpo-adv} is computed at the \textit{trajectory level}, which means the same advantage is assigned uniformly across the entire trajectory, without distinguishing the contributions of individual turns or tokens. 
For long-horizon multi-turn tasks, such coarse-grained credit assignment often leads to unstable training and suboptimal performance \citep{guo2025segment,feng2025group,zhang2025rlvmr}.

\subsection{GRPO with Fine-Grained Credit Assignment}
We now derive GRPO with fine-grained credit assignment under a dense reward structure for $\mathcal{M}_3$, which we refer to as multi-turn GRPO (MT-GRPO). 
We begin by defining two types of advantages: the outcome advantage and the intermediate advantage.
\begin{itemize}[itemsep=0mm, topsep=0mm]
    \item The outcome advantage captures global task completion signals. Given a group of outcome rewards $R^O_i=R(x,y_i)$, it is defined as 
    \begin{equation}
    A^{O}_{i} = \frac{R^{O}_i - \text{mean}(\{R^{O}_i\}_{i=1}^G)}{\text{std}(\{R^{O}_i\}_{i=1}^G)},
    \end{equation}
    \item The intermediate advantage captures local optimization signals by comparing returns across trajectories at the same timestep. 
    At the $k$-th turn ($k=1,\ldots,K-1$), given a state $s_k$, the algorithm samples $G$ actions $\{a_{i,(k)}\}_{i=1}^G$, resulting in a group of intermediate rewards $R^I_{i,(k)} = R(s_k, a_{i,(k)})$. 
    The intermediate advantage is defined as
    \begin{equation}
    A^{I}_{i,(k)} = \frac{R^I_{i,(k)} - \text{mean}(\{R^I_{i,(k)}\}_{i=1}^G)}{\text{std}(\{R^I_{i,(k)}\}_{i=1}^G)}
    \end{equation}
\end{itemize}
We combine these into a unified  advantage that assigns credit at both global and local scales by aggregating current and future advantages:
\begin{equation}
      A^{\text{MT-GPRO}}_{i,(k)} = \sum_{l=k}^{K-1} \alpha^{l-k} A^{I}_{i,(l)} + \alpha^{K-k} A^{O}_{i} 
\end{equation}
where $\alpha\in[0,1]$ is a discount coefficient controlling the relative weight of current and future terms. 
This aggregated advantage is uniformly assigned to all tokens generated within the $k$-th turn, i.e., $A_{i,1}  = \cdots = A_{i,t} = A^{\text{MT-GPRO}}_{i,(k)}$, where $t$ indexes tokens within the $k$-th turn.
It is easy to see that the per-turn rollout mechanism of MT-GRPO can also be applied to other group-normalization advantage algorithms, including DAPO~\citep{yu2025dapo}, Dr.GRPO~\citep{liu2025understanding}, and RLOO~\citep{kool2019buy,ahmadian2024back}.
In MT-GRPO, computing the intermediate advantages requires $G$ rollout samples at each turn for $k=1,\ldots,K-1$. 
Note that rollouts are not needed at the final turn; instead, the final advantage is computed after collecting all per-state rollout samples.
Therefore, over a horizon of $K$ turns, this results in $G^{K-1}$ rollout trajectories in total. 
When $k=2$, only $G$ rollout trajectories are required, which is the same as in the vanilla GRPO setting.

\begin{figure}
    \centering
    \includegraphics[width=0.95\linewidth]{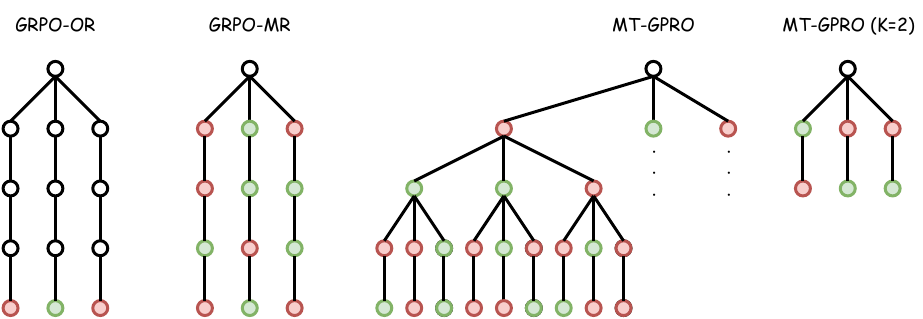}
    \caption{Comparison of rollout structures among GRPO variants.
    GRPO-OR for $\mathcal{M}_1$, GPRO-MR for $\mathcal{M}_2$, and MT-GRPO for $\mathcal{M}_3$
    The \textcolor{red}{red} and \textcolor{Green}{green} nodes denote the presence of turn-level rewards, while uncolored nodes indicate the absence of turn-level rewards.}
    \label{fig:grpo-tree}
\end{figure}

We compare GRPO variants under different turn-level MDP formulations in Section~\ref{sec:reward-assignment}, namely GRPO-OR for $\mathcal{M}_1$, GRPO-MR for $\mathcal{M}_2$, and MT-GRPO for $\mathcal{M}_3$.
Figure~\ref{fig:grpo-tree} illustrates the rollout structures of these GRPO variants.
We observe that GRPO-OR and GRPO-MR perform per-trajectory rollouts (chain-based structures), whereas MT-GPRO performs per-state (per-turn) rollouts (tree-based structures).
As a result, the computational complexity of GRPO scales linearly with the number of turns, while that of MT-GPRO grows exponentially with respect to the number of turns.

\textit{Remark.}
MT-GPRO has two limitations:
(1) In MT-GRPO, computing the intermediate advantages requires $G$ rollout samples at each turn.
Therefore, over a horizon of $K$ turns, this results in $G^{K-1}$ rollout trajectories in total. Such \textit{exponential} growth in complexity makes the approach computationally prohibitive for long-horizon multi-turn tasks.
(2) This strategy also assumes that all rollout samples in a group must contain the \textit{same number of turns}, which requires enforcing this constraint in the system prompt and leads to a fixed-turn setting. Such a restriction limits the flexibility and applicability of GRPO in more diverse scenarios.
For example, in a search task, one question may be resolved in a single tool call or require multiple calls to retrieve, filter, and refine results in a sampled group.

\subsection{Experiments for GPRO-Type Algorithms}

\textbf{Experiment Setup.}
We conduct experiments to evaluate three GRPO algorithms in a two-turn agent setting, where the agent first calls the search tool with reasoning in the initial turn and then produces the final answer in the subsequent turn (see Appendix~\ref{sec:grpo-exp} for details). 
We use Qwen2.5-7B \citep{yang2024qwen2} as the base model and use the TriviaQA dataset \citep{joshi2017triviaqa} to train the LLM agent for answering questions by interacting with a Wikipedia search engine.
Beyond the outcome-level exact-match reward, we design intermediate rewards based on tool-execution feedback.

\begin{figure}
    \centering
    \begin{subfigure}[b]{0.23\textwidth}
        \includegraphics[width=\linewidth]{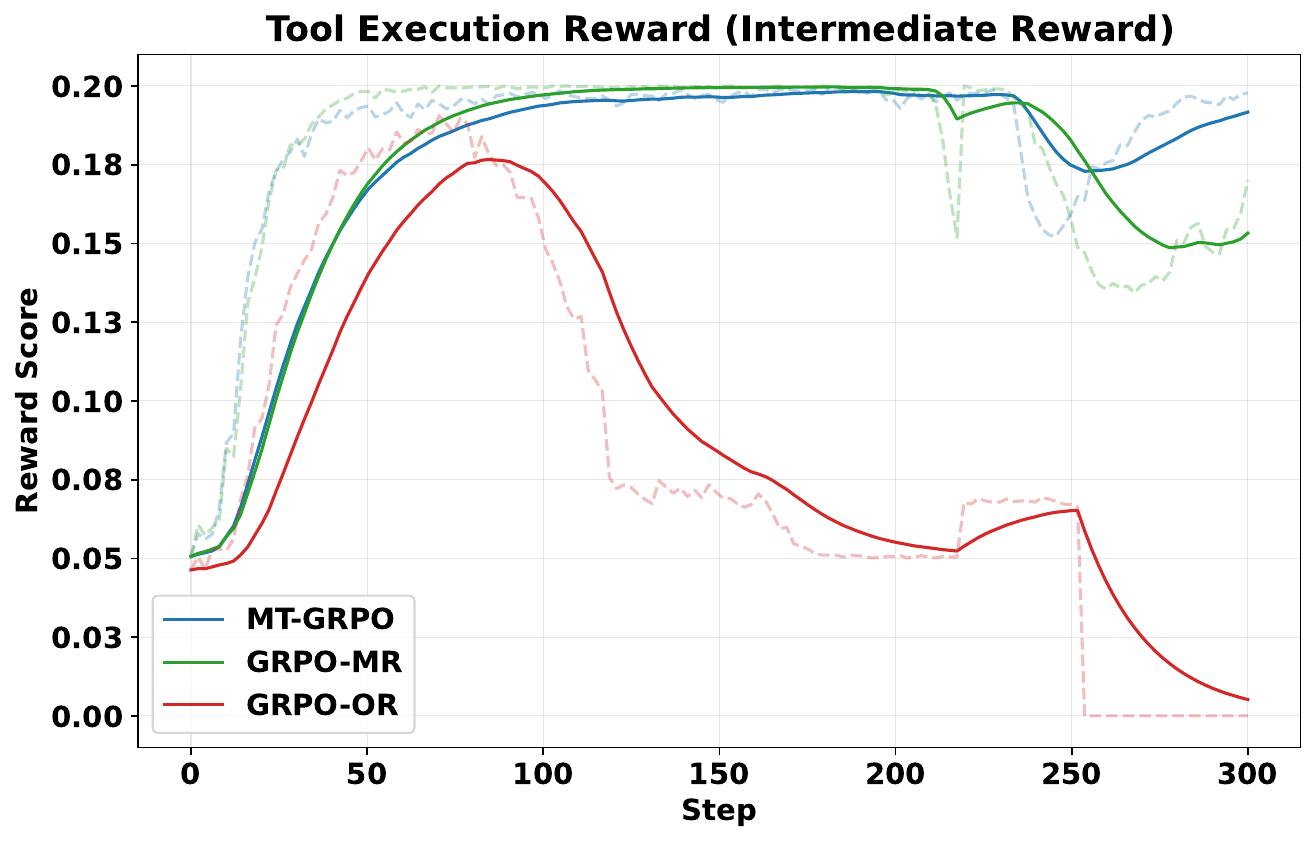}
    \end{subfigure}
    \hfill
    \begin{subfigure}[b]{0.23\textwidth}
        \includegraphics[width=\linewidth]{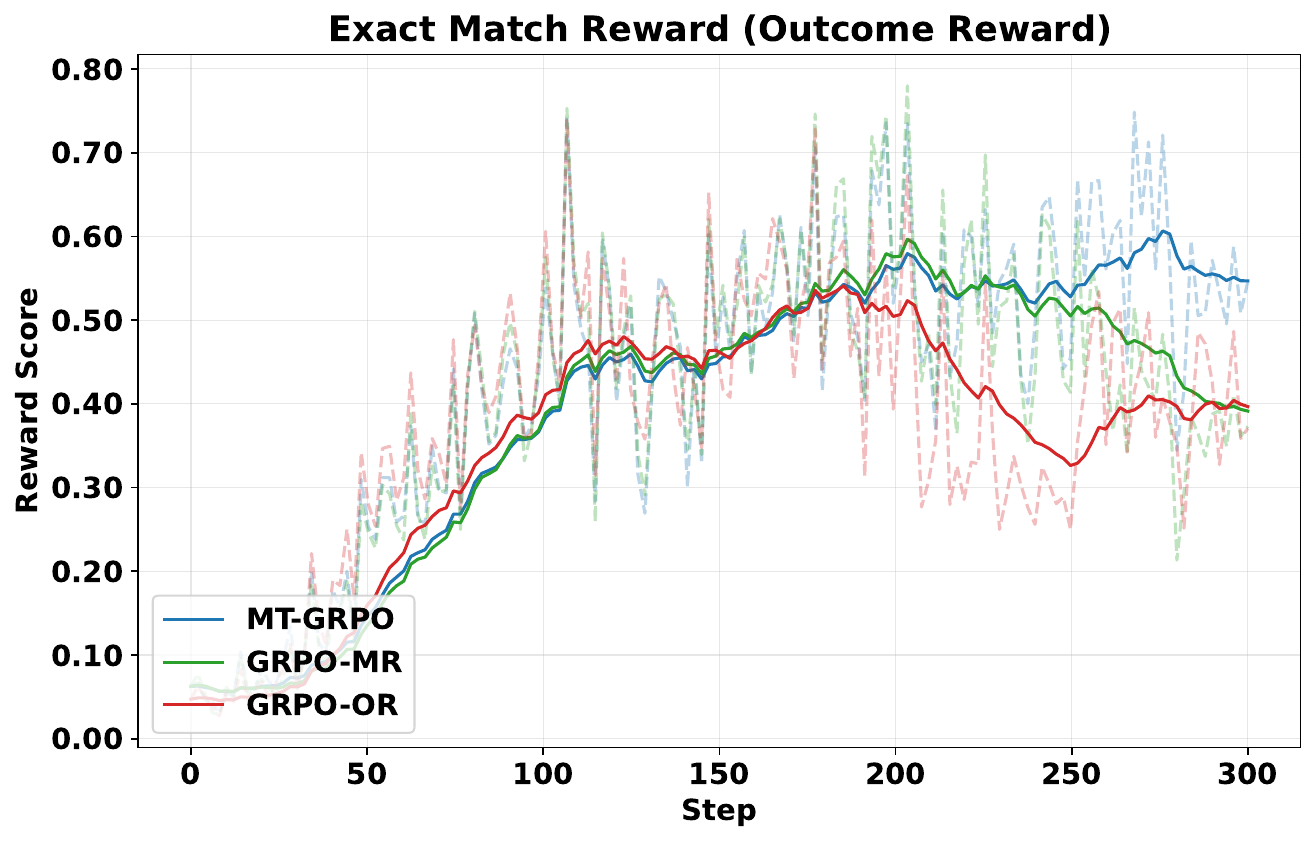}
    \end{subfigure}
    \caption{Curves for different training reward components during training with various GRPO algorithms.
    GPRO-OR, GPRO-MR, and MT-GPRO correspond to the MDPs $\mathcal{M}_1$, $\mathcal{M}_2$, and $\mathcal{M}_3$, respectively.
    Each plot shows the training reward score over training steps. Dotted lines represent the average reward across 10 runs, while solid lines show trends smoothed using the Exponential Moving Average (EMA). 
    }
    \label{fig:exp-grpo-1}
\end{figure}

\textbf{Experiment Results on Training Dynamics.}
Figure~\ref{fig:exp-grpo-1} presents training reward curves for different GPRO variants, which show that MT-GRPO achieves more stable tool usage (left figure), 100\% success in tool execution, and higher exact-match accuracy (right figure), highlighting the importance of fine-grained credit assignment in multi-turn agentic tasks. Other reward component curves during training can be found in Figure~\ref{fig:exp}.
Figures~\ref{fig:gpro-or}, \ref{fig:gpro-mr}, and \ref{fig:mtgpro} in Appendix~\ref{sec:grpo-exp} illustrate reward component curves during training with different algorithms, showcasing the variability in learning performance. Notably, the MT-GRPO method demonstrates lower variance during training, while GRPO-OR and GRPO-MR exhibit greater instability. 
An interesting observation is that the tool execution curve of MT-GRPO temporarily drops sharply around step 230–250 but subsequently recovers and stabilizes. This demonstrates that even if the agent forgets to call search tools in the middle of the training, it eventually learns to incorporate them in the final stages.
This finding further emphasizes the significance of credit assignment, contributing to more stable training.

\textbf{Experiment Results on Benchmark Performance.}
Table~\ref{tab:validation-results} presents the validation reward scores across different models. MT-GRPO achieves the highest performance in all reward metrics. Compared to GRPO-MR, which reaches 0.3724 in final search answer and 0.3346 in exact match, MT-GRPO demonstrates clear improvements, especially in exact match with a margin of +0.1664. In contrast, GRPO-OR performs poorly across all metrics, scoring 0 in intermediate rewards and only 0.04 in XML format. These results confirm that fine-grained, dense reward structure and turn-level credit assignment employed by MT-GRPO lead to superior turn-level decision-making and more accurate final outcomes in multi-turn tasks, compared to the sparse, trajectory-level reward structure and trajectory-level credit assignment used in vanilla GRPO.

\begin{table}
\caption{Performance comparison across different methods on reward scores evaluated on the validation set. Values in parentheses indicate the reward range for each metric. Bold numbers indicate the best performance for each reward type.}
\label{tab:validation-results}
\centering
\resizebox{\columnwidth}{!}{
\renewcommand{\arraystretch}{1.5}
\begin{tabular}{m{3cm} 
                >{\centering\arraybackslash}m{2.4cm}
                >{\centering\arraybackslash}m{2.4cm}
                >{\centering\arraybackslash}m{2cm} 
                >{\centering\arraybackslash}m{2cm}}
\toprule
\multirow{2}{*}{Model} & \multicolumn{2}{c}{Intermediate Reward} & \multicolumn{2}{c}{Outcome Reward} \\
\cmidrule(lr){2-3} \cmidrule(lr){4-5}
 & Tool Execution (0-0.2) & Search Answer (0-0.5) & XML Format (0-0.2) & Exact Match (0-1) \\
\midrule
Qwen2.5-7B-Base & 0.0559 & 0.0934 & 0.1562 & 0.0469 \\
Qwen2.5-7B-Instruct & 0.1626 & 0.2814 & 0.1982 & 0.1559 \\
GRPO-OR & 0 & 0 & 0.04 & 0 \\
GRPO-MR & 0.2 & 0.3724 & 0.1994 & 0.3346 \\
\rowcolor{softred}
MT-GRPO & 0.2 & \textbf{0.3926} & \textbf{0.1996} & \textbf{0.5010} \\
\bottomrule
\end{tabular}
}
\end{table}

\section{Credit Assignment for PPO-Type Algorithms in Multi-Turn Tasks}

\subsection{Preliminaries: PPO}

Proximal Policy Optimization (PPO) \citep{schulman2017proximal} is a popular actor-critic RL algorithm commonly used for LLM training \citep{ouyang2022training}. In PPO, the advantage estimate $A_t$ is computed using Generalized Advantage Estimation (GAE) \citep{schulman2015high}, based on rewards and a learned value function (critic model). 
Formally, for a trajectory of length \( T \), the GAE \( A_t \) at time step \( t \) is computed as: $A_t = \sum_{l=0}^{T-t-1} (\gamma \lambda)^l \delta_{t+l}$
where $\gamma$ is the discount factor, \( \lambda \in [0,1] \) is the GAE parameter, $\delta_t =  r_t + \gamma V_{t+1} - V_{t}$ is the temporal-difference (TD) error, $r_t$ is the token-level reward and $V_t$ is the token-level value at step $t$.
Through the mechanism of GAE, the token-level value function enables token-level advantage estimation.

\subsection{Turn-Level Reward Assignment in PPO}

In PPO, turn-level rewards can be readily incorporated through the use of GAE.
We present the reward assignment strategy for the token-level reward signal $r_t$ in GAE under three MDP formulations as illustrated in Figure~\ref{fig:reward-assignment}.
\begin{enumerate}[itemsep=0mm, topsep=0mm]
    \item $\mathcal{M}_1$: the outcome reward is assigned to the final token of the entire trajectory.
    \item $\mathcal{M}_2$: the merged reward from outcome and intermediate signals is assigned to the final token of the entire trajectory.
    \item $\mathcal{M}_3$: the outcome reward is assigned to the final token of the entire trajectory, while intermediate rewards are assigned to the final token of the corresponding turn.
\end{enumerate}
With explicit intermediate rewards, GAE provides fine-grained training signals at each turn. 
We refer to the PPO variants corresponding to three MDP formulations as PPO-OR, PPO-MR, and MT-PPO, respectively.
Formally, the advantage estimate is defined as $A_t = \sum_{l=0}^{T-t-1} (\gamma \lambda)^l \delta_{t+l}$ with TD error
\[
\delta^{\text{PPO}}_t =  
\begin{cases}
    \gamma V_{t+1} - V_{t} & t<T-1 \\
    R^{\text{traj}} - V_{T} & t= T-1
\end{cases}
\]
where for $\mathcal{M}_1$, $R^{\text{traj}}_i=R(x,y_i)$,
and for $\mathcal{M}_2$, $R^{\text{traj}}_i = \sum_{k=1}^K \gamma^k R_{i,k}$.
For $\mathcal{M}_3$, the TD error is modified to incorporate explicit per-turn rewards:
\[
\delta^{\text{MT-PPO}}_t =
\begin{cases}
R^O - V_T, & t = t_{\text{traj}} \\
R^I + \gamma V_{t+1} - V_t, & t = t_{\text{turn}} \\
\gamma V_{t+1} - V_t, & \text{otherwise}
\end{cases}
\]
where $t_{\text{traj}}$ denotes the final token of the entire trajectory and
$t_{\text{turn}}$ denotes the final token of an intermediate turn.

\subsection{Turn-Decomposed PPO}

Inspired by recent work on turn-decomposed training \cite{feng2025group}, we present the turn-decomposed PPO algorithm. Unlike trajectory-based RL algorithms, which rely on full multi-turn rollout trajectories as training data, turn-decomposed RL decomposes each multi-turn rollout into a set of single-turn training instances, as illustrated in Figure~\ref{fig:turn-decomposed}.

In turn-decomposed RL training, each decomposed turn is treated as an independent single-turn data point associated with a reward. The core idea of turn-decomposed PPO is to first decompose multi-turn rollout data into single-turn instances and then apply PPO to each instance using an assigned reward signal. Depending on the reward structure discussed earlier, the reward for each turn-decomposed instance can be derived from terminal rewards, delayed rewards, or explicit per-turn rewards in the original trajectory-based formulation.

In the following experiment section, we compare trajectory-based PPO and turn-decomposed PPO, as well as different reward structures within turn-decomposed training, to study their impact on learning effectiveness and credit assignment.

\section{Experiments on Multi-Turn Search Agents}

In our experiments, we build our codebase upon the open-source project Search-R1 \citep{jin2025search}, which trains LLM agents for multi-turn reasoning-augmented search tasks. More details on experimental settings can be found in Appendix~\ref{sec:exp-detail}.

\subsection{Experiment Setup}

\textbf{Datasets.}  
These datasets are categorized as follows: 
(1) General Question Answering: NQ \citep{karpukhin2020dense}, TriviaQA \citep{joshi2017triviaqa}, and PopQA \citep{mallen2022not}. 
(2) Multi-Hop Question Answering: HotpotQA \citep{yang2018hotpotqa}, 2WikiMultiHopQA \citep{ho2020constructing}, Musique \citep{trivedi2022musique} and Bamboogle \citep{press2023measuring}.

\textbf{Training Details.}
We use Qwen2.5-7B \citep{yang2024qwen2} as the base model, E5 \citep{wang2022text} as the retriever, and 2018 Wikipedia dump \citep{karpukhin2020dense} as the corpus.
We set the number of retrieved passages to 3, and the maximum number of turns $N_{\max}$ to 4.
The system prompt follows that of Search-R1 \citep{jin2025search}. 
We also enable policy loss masking on retrieved tokens. 

\textbf{LLM Judge as PRM.}
We use \textsc{gpt-oss-120b} \citep{openai2025gptoss120bgptoss20bmodel} as the judge model to evaluate the quality of each turn and provide intermediate rewards. The outcome reward is computed based on exact-match answer correctness. The turn-level judge prompt can be found in Appendix~\ref{sec:judge}.

\textbf{Evaluation Metrics.}
We evaluate model performance using three types of rewards: (1) answer correctness (exact match) reward, (2) format correctness reward, and (3) retrieval correctness reward. Each reward is assigned a value of 1.0 if the criterion is satisfied and 0 otherwise.

\subsection{Evaluated Methods}

We compare both training reward dynamics and benchmark performance across different methods.

\textbf{Training Dynamics.}
We evaluate PPO-type algorithms, including PPO-OR, PPO-MR, and MT-PPO.
We omit GRPO training curves since, as reported in~\citep{jin2025search}, GRPO consistently crashes during training.

\textbf{Benchmark Evaluation.}
We benchmark several RL algorithms mentioned in this paper with state-of-the-art methods. These include PPO-based methods (PPO-OR, PPO-MR, and MT-PPO), GRPO-based methods (GRPO-OR and GRPO-MR), OTC trained with both GRPO and PPO \citep{wang2025otc}, StepSearch trained with PPO \citep{wang2025stepsearch}, PPR trained with PPO \citep{xu2025hybrid}, and turn-decomposed PPO.
Since PPO-based baselines often experience training collapse, we report results using either the final checkpoint or the last stable checkpoint prior to collapse.

\subsection{Main Results}

\begin{figure}
    \centering
    \begin{subfigure}[b]{0.23\textwidth}
        \includegraphics[width=\linewidth]{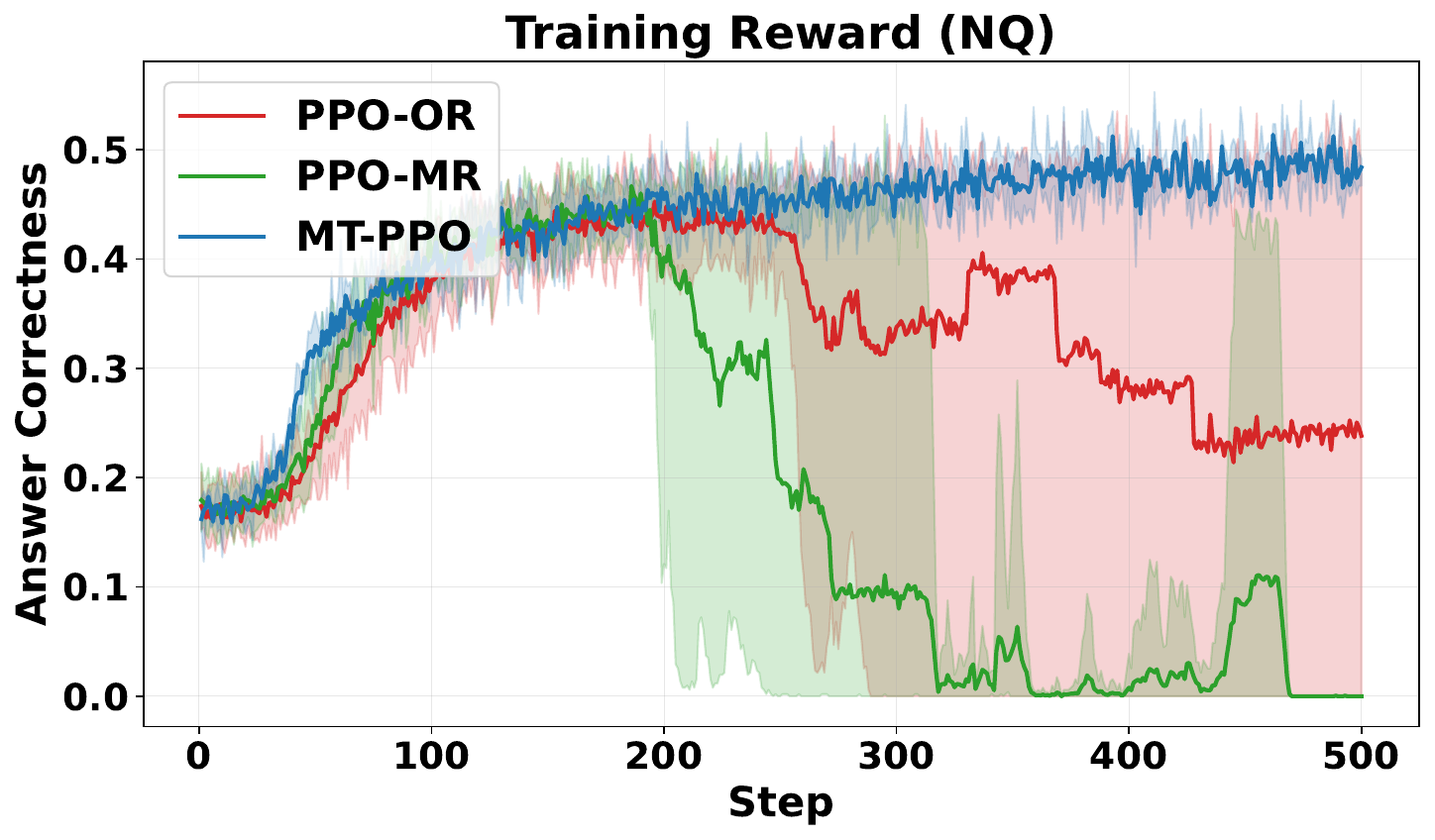}
    \end{subfigure}
    \hfill   
    \begin{subfigure}[b]{0.23\textwidth}
        \includegraphics[width=\linewidth]{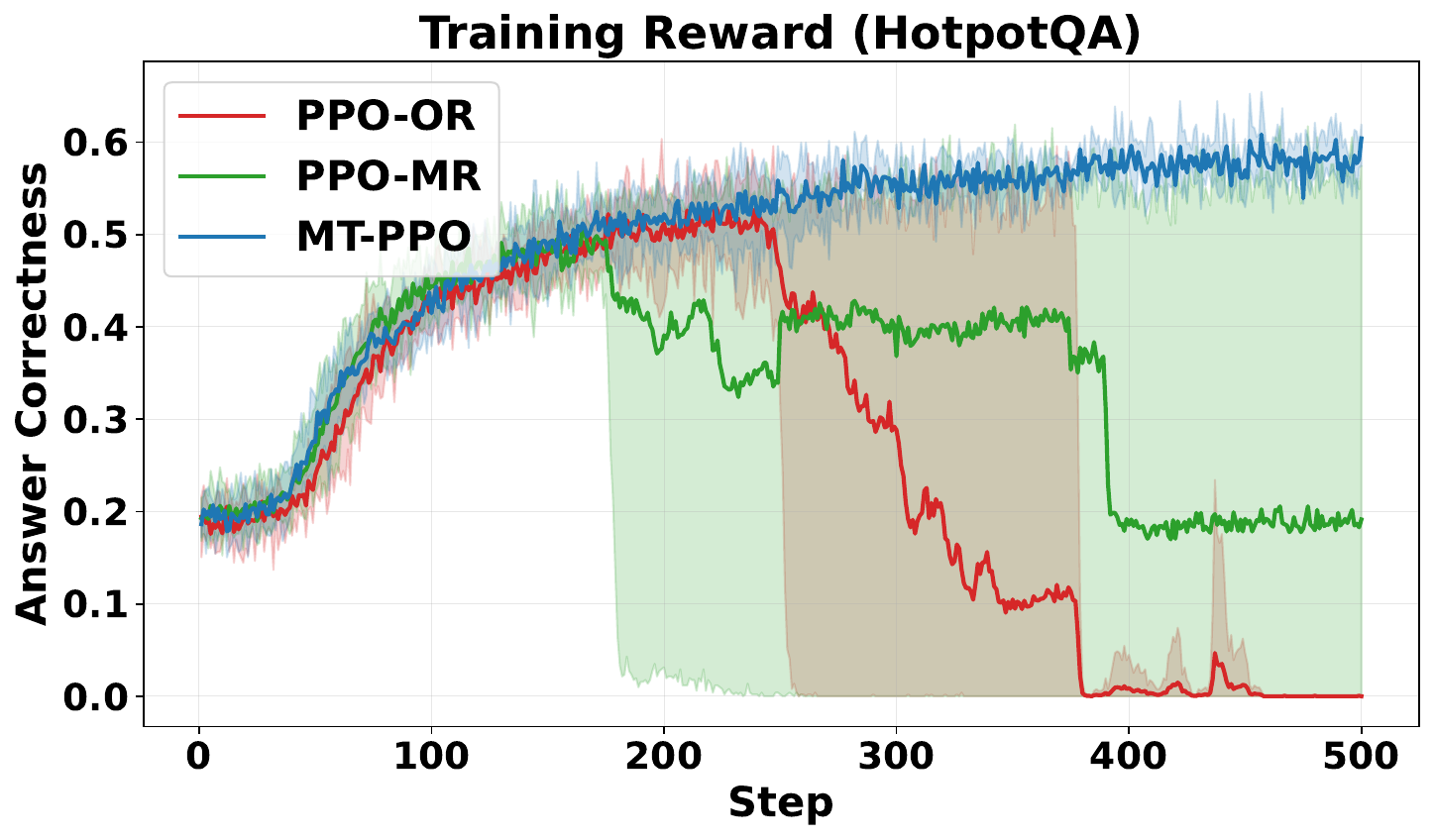}
    \end{subfigure}
    \caption{Training reward (answer correctness) curves recorded during training for PPO-OR, PPO-MR and MT-PPO on the NQ and HotpotQA datasets. Solid lines show mean reward values, while shaded regions indicate variability across five independent runs.}
    \label{fig:exp-ppo-training-reward-1}
\end{figure}

\begin{table*}[t]
\centering
\small
\renewcommand{\arraystretch}{1.25}
\caption{
The performance results of different methods on six datasets. 
Bold numbers indicate the best performance for each dataset and metric. 
\textsuperscript{\textdagger}/\textsuperscript{\textasteriskcentered} denote in-domain/out-of-domain datasets. 
\textsuperscript{\textdaggerdbl} indicates results copied from the original paper.
}
\label{tab:eval}
\resizebox{\textwidth}{!}{
\begin{tabular}{lcccccccc}
\toprule
\multirow{2}{*}{\textbf{Methods}} 
  & \multicolumn{3}{c}{\textbf{General QA}} 
  & \multicolumn{4}{c}{\textbf{Multi-Hop QA}} 
  & \multirow{2}{*}{\textbf{Avg.}} \\
\cmidrule(lr){2-4} \cmidrule(lr){5-8}
 & \textbf{NQ\textsuperscript{\textdagger}} 
 & \textbf{TriviaQA\textsuperscript{\textasteriskcentered}} 
 & \textbf{PopQA\textsuperscript{\textasteriskcentered}} 
 & \textbf{HotpotQA\textsuperscript{\textdagger}} 
 & \textbf{2wiki\textsuperscript{\textasteriskcentered}} 
 & \textbf{Musique\textsuperscript{\textasteriskcentered}} 
 & \textbf{Bamboogle\textsuperscript{\textasteriskcentered}}
 &  \\
\midrule
\rowcolor{Gray!15}
\multicolumn{9}{c}{\textit{Answer Correctness (Exact Match)}} \\
\midrule
Qwen2.5-7B-Base          & 0.177 & 0.319 & 0.181 & 0.160 & 0.167 & 0.040 & 0.120 & 0.174 \\
Qwen2.5-7B-Instruct      & 0.320 & 0.563 & 0.349 & 0.292 & 0.277 & 0.118 & 0.112 & 0.320 \\
GRPO-OR (Search-R1)      & 0.391 & 0.560 & 0.388 & 0.331 & 0.306 & 0.129 & 0.346 & 0.351 \\
GRPO-MR (Search-R1)\textsuperscript{\textdaggerdbl}  
                         & 0.453 & 0.628 & 0.450 & 0.416 & 0.375 & 0.164 & 0.387 & 0.414 \\
PPO-OR (Search-R1)       & 0.483 & 0.639 & 0.456 & 0.435 & 0.382 & 0.199 & 0.432 & 0.432 \\
PPO-MR (Search-R1)\textsuperscript{\textdaggerdbl}   
                         & 0.472 & 0.629 & 0.452 & 0.436 & 0.402 & 0.180 & 0.363 & 0.429 \\
Turn-Decomposed PPO-OR   & 0.425 & 0.598 & 0.448 & 0.322 & 0.324 & 0.102 & 0.125 & 0.335 \\
Turn-Decomposed PPO-MR   & 0.269 & 0.522 & 0.260 & 0.224 & 0.274 & 0.062 & 0.349 & 0.280 \\
Turn-Decomposed MT-PPO   & 0.304 & 0.522 & 0.309 & 0.259 & 0.286 & 0.085 & 0.287 & 0.293 \\
GRPO-OR (OTC)\textsuperscript{\textdaggerdbl}           
                         & 0.444 & 0.597 & 0.431 & 0.366 & 0.311 & 0.130 & 0.025 & 0.380 \\
PPO-OR (OTC)\textsuperscript{\textdaggerdbl}            
                         & 0.446 & 0.623 & 0.425 & 0.383 & 0.363 & 0.152 & 0.391 & 0.399 \\
MT-PPO (StepSearch)         & 0.355 & 0.570 & 0.385 & 0.351 & 0.396 & 0.179 & 0.373 & 0.373 \\
MT-PPO (PPR)\textsuperscript{\textdaggerdbl}             
                         & 0.458 & 0.610 & 0.437 & 0.386 & 0.355 & 0.147 & 0.355 & 0.393 \\
\rowcolor{softred}
MT-PPO (ours)            & \textbf{0.490} & \textbf{0.647} & \textbf{0.459} & \textbf{0.453} & \textbf{0.424} & \textbf{0.209} & \textbf{0.445} & \textbf{0.447} \\
\midrule
\rowcolor{Gray!15}
\multicolumn{9}{c}{\textit{Format Correctness}} \\
\midrule
Qwen2.5-7B-Base          & 0.118 & 0.118 & 0.105 & 0.098 & 0.084 & 0.082 & 0.084 & 0.101 \\
Qwen2.5-7B-Instruct      & 0.183 & 0.267 & 0.067 & 0.109 & 0.037 & 0.071 & 0.064 & 0.122 \\
GRPO-OR (Search-R1)      & 0.706 & 0.685 & 0.597 & 0.513 & 0.376 & 0.328 & 0.564 & 0.534 \\
PPO-OR (Search-R1)       & 0.909 & 0.954 & 0.952 & 0.916 & 0.806 & 0.834 & 0.951 & 0.895 \\
MT-PPO (StepSearch)         & 0.521 & 0.614 & 0.668 & 0.560 & 0.396 & 0.571 & 0.541 & 0.555 \\
\rowcolor{softred}
MT-PPO (ours)            & \textbf{0.999} & \textbf{0.997} & \textbf{0.999} & \textbf{0.998} & \textbf{0.999} & \textbf{0.999} & \textbf{0.999} & \textbf{0.999} \\
\bottomrule
\end{tabular}
}
\end{table*}

\textbf{Training Dynamics on PPO-Type Algorithms.}
Figures~\ref{fig:exp-ppo-training-reward-1}, and~\ref{fig:exp-ppo-training-reward} and~\ref{fig:exp-ppo-val-reward} in Appendix~\ref{sec:ppo-more} show training and validation reward curves for PPO-OR, PPO-MR and MT-PPO. MT-PPO achieves substantially more stable training, converging faster in the early phase (first 100 steps) thanks to intermediate rewards that provide stronger guidance. As training progresses, PPO-OR and PPO-MR exhibit high variance and even performance degradation, especially on HotpotQA, while MT-PPO maintains consistent improvement. MT-PPO attains higher average accuracy than PPO, demonstrating greater robustness.
Format reward curves show that MT-PPO consistently follows the correct output format, while PPO-OR and PPO-MR struggle, especially on HotpotQA, where formatting mistakes prevent correct evaluation. This indicates that turn-level rewards in MT-PPO stabilize training and enforce structural correctness. Retrieval curves further show that MT-PPO achieves more consistent accuracy by leveraging intermediate signals to guide reasoning. 

\textbf{Benchmark Performance on PPO-Type Algorithms.}
Our MT-PPO consistently outperforms PPO-OR and PPO-MR in answer correctness across all datasets. On average, MT-PPO improves over PPO-OR by approximately 3.5\% and over PPO-MR by 4.2\% in exact match accuracy. The gains are more pronounced on multi-hop datasets: MT-PPO achieves a 3.9\% relative improvement on HotpotQA and a 5.5\% improvement on 2Wiki, highlighting the effectiveness of dense turn-level reward structure for long-horizon reasoning.

\textbf{Turn-Decomposed vs. Trajectory-Based Training.}
Turn-decomposed PPO variants consistently underperform standard trajectory-based PPO, indicating that naïvely decomposing multi-turn trajectories can lead to suboptimal credit assignment. In particular, turn-decomposed PPO often achieves high intermediate rewards while yielding low outcome rewards, suggesting that the model over-optimizes local turn-level signals without sufficiently aligning them with final task success.

\textbf{Comparison with Concurrent Work.}
Compared to concurrent MT-PPO approaches, including PPR and StepSearch, our MT-PPO with judge-based turn-level rewards achieves consistently higher answer correctness, demonstrating the advantage of LLM-based process supervision.

\textbf{Overall Performance.}
Our MT-PPO method achieves the best performance over all SOTA method, and near-perfect format correctness across all datasets, approaching 100\%. This result underscores the effectiveness of fine-grained credit assignment and turn-level reward structure in producing not only correct answers but also well-structured and executable outputs.

\section{Experiments on Multi-Turn Game Agents}
\label{sec:game}

In this section, we present an additional case study on multi-turn game agents, focusing on the classic grid-based puzzle Sokoban \citep{SchraderSokoban2018}. In this puzzle, the agent must push all boxes to designated target locations. The environment is represented as a 2D grid, and the action space is discrete (up, down, left, right). The key challenge is that Sokoban is irreversible: boxes can be pushed but not pulled, so a single wrong move may lead to an unrecoverable dead-end. As a result, solving the puzzle requires the agent to reason several steps ahead rather than relying on simple navigation heuristics.

\textbf{Experiment Setup.}
In our experiments, we use Qwen2.5-VL-3B \citep{Qwen-VL} as the base model. The Sokoban environment is configured with a $(6 \times 6)$ grid and allows up to 100 steps per episode. Each puzzle contains one box, and solving it requires at least five actions. During interaction, the agent may take up to three actions per turn, and it can interact with the environment for a maximum of three turns. For reward design, the agent receives a success reward of 10 when all boxes are placed on their target locations in the final state. At each turn, it receives a box-placement reward of 1 for each box pushed onto a target, along with a format reward of 0.5 to encourage proper visual-state reasoning and structured output. A failure penalty of $-0.1$ is applied at each step when the task remains incomplete. We compare  MT-PPO with GRPO-OR, GPRO-MR, PPO-OR, PPO-MR.

\begin{figure}
    \centering
    \begin{subfigure}[b]{0.23\textwidth}
        \includegraphics[width=\linewidth]{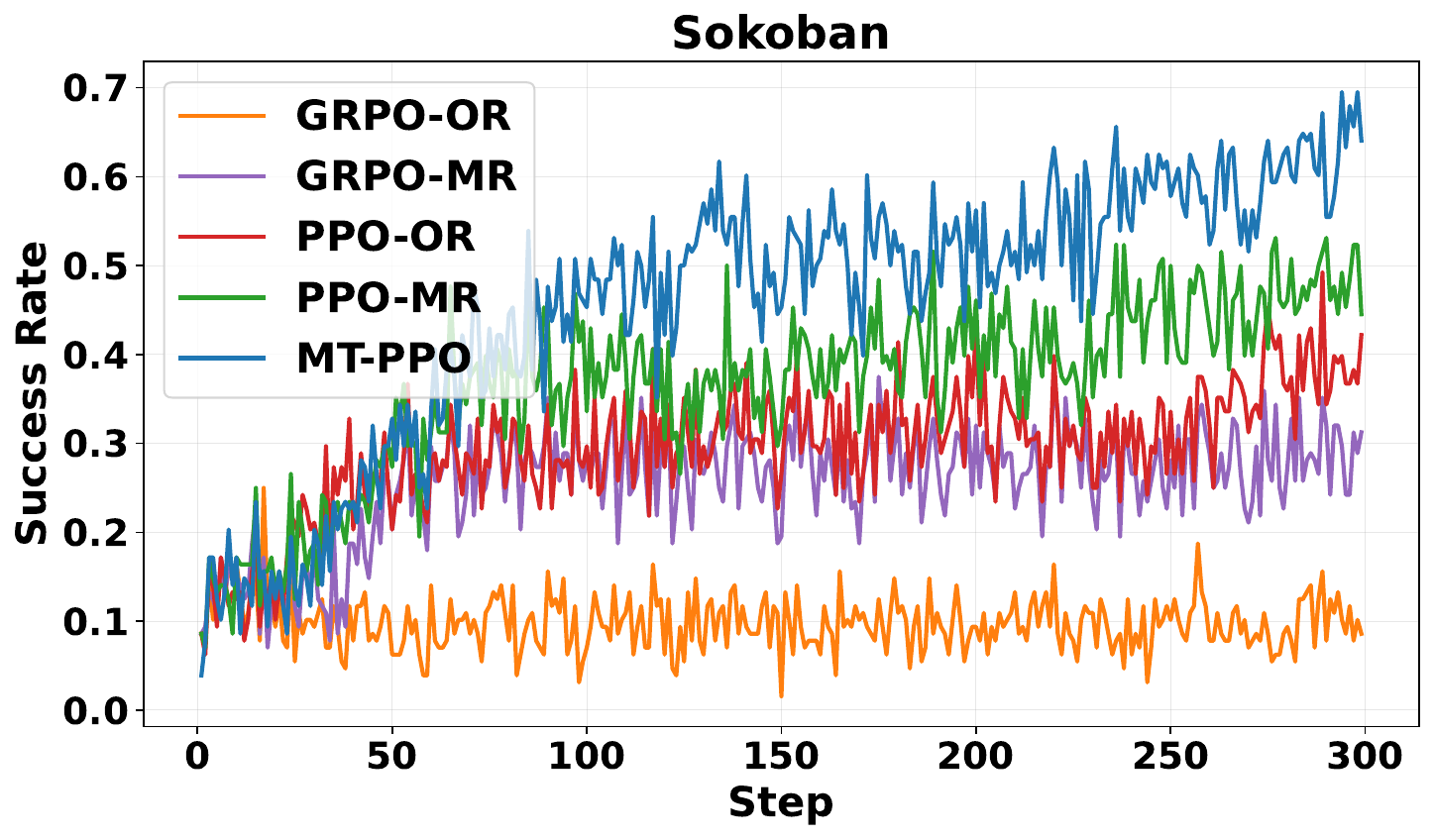}
    \end{subfigure}
    \hfill
    \begin{subfigure}[b]{0.23\textwidth}
        \includegraphics[width=\linewidth]{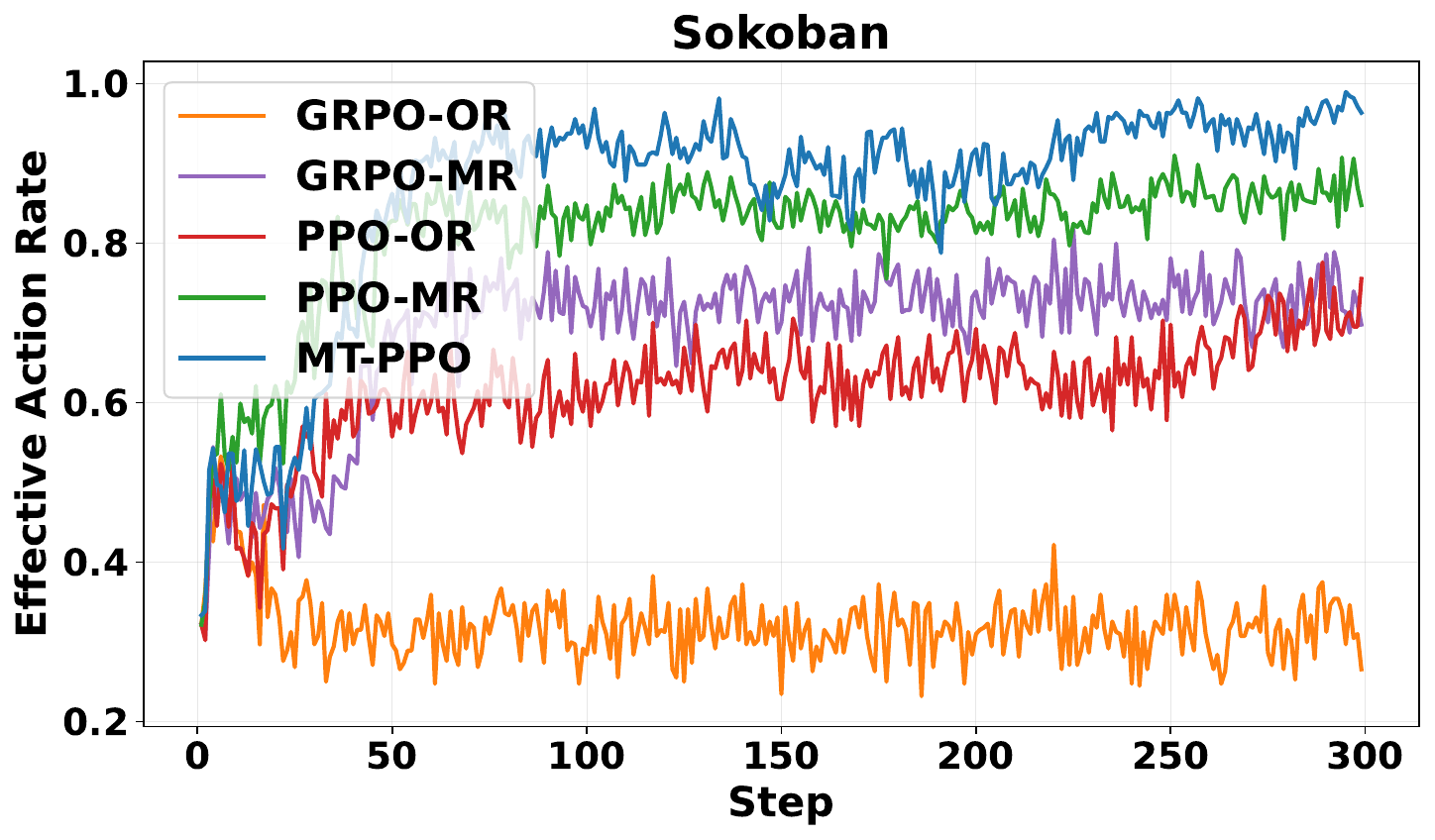}
    \end{subfigure}
    \caption{Training curves on the Sokoban environment. Left: success rate over training steps. Right: effective action rate, measuring the proportion of actions that contribute to valid box movement.}
    \label{fig:sokoban}
\end{figure}

\textbf{Experimental Results.}
The experimental results in Figure~\ref{fig:sokoban} show the training dynamics of different methods on the Sokoban environment. MT-PPO consistently achieves a higher success rate throughout training and maintains a substantially higher effective action rate, demonstrating its ability to cope with the long-horizon and irreversible structure of Sokoban. These curves highlight how MT-PPO learns more stable and purposeful action sequences during optimization.
Table~\ref{tab:sokoban-results} in Appendix~\ref{sec:game} reports the final test performance at step 300. MT-PPO achieves the highest scores on both success rate and effective action rate, outperforming all PPO and GRPO baselines by a significant margin. The results illustrate the advantages of explicit turn-level rewards and fine-grained credit assignment, further validating the generality of our approach beyond language-based tasks.

\section{Conclusion}

In this paper, we investigate how dense, turn-level reward structures can be effectively leveraged within GRPO and PPO, to achieve fine-grained credit assignment. Through extensive experiments on two agent tasks, we demonstrate that for both GRPO and PPO, dense per-turn reward structures consistently outperform sparse terminal and delayed reward designs, leading to more stable training dynamics and superior overall performance.

\clearpage

\section{Impact Statements}

This paper presents work whose goal is to advance the field of machine learning. There are many potential societal consequences of our work, none of which we feel must be specifically highlighted here.

\bibliography{icml2026_ref}
\bibliographystyle{icml2026}

\newpage
\appendix
\onecolumn

\section{PPO Experiments on Search Agents}

\subsection{Details for Experimental Setup (PPO)}
\label{sec:exp-detail}

\subsubsection{Evaluated Methods}

We list all evaluated methods.
\begin{itemize}[itemsep=0mm, topsep=0mm]
    \item PPO-OR~\citep{jin2025search}: vanilla PPO trained with only \underline{o}utcome \underline{r}ewards, where the trajectory-level reward is a binary signal indicating final-answer correctness, corresponding to the terminal-reward MDP \(\mathcal{M}_1\).
    \item PPO-MR~\citep{jin2025empirical}: vanilla PPO trained with \underline{m}erged intermediate and outcome \underline{r}ewards, where the trajectory-level reward combines intermediate rewards (retrieval correctness) and outcome rewards (answer correctness and format correctness), corresponding to the delayed-reward MDP \(\mathcal{M}_2\). The detailed reward design is provided in Section 4.1 of~\citep{jin2025empirical}.
    \item MT-PPO (ours): PPO variant trained with both intermediate and outcome rewards, where the turn-level reward design is described in Section~\ref{sec:judge}, with $\lambda_s = 0.1$ by default, corresponding to the turn-level-reward  MDP \(\mathcal{M}_3\).
    \item GRPO-OR~\citep{jin2025search}: vanilla GRPO trained with only \underline{o}utcome \underline{r}ewards, where the trajectory-level reward is a binary signal indicating final-answer correctness, corresponding to the terminal-reward MDP \(\mathcal{M}_1\).
    \item GRPO-MR~\citep{jin2025empirical}: vanilla GRPO trained with \underline{m}erged intermediate and outcome \underline{r}ewards, where the trajectory-level reward combines intermediate rewards (retrieval correctness) and outcome rewards (answer correctness and format correctness), corresponding to the delayed-reward MDP \(\mathcal{M}_2\). The detailed reward design is provided in Section 4.1 of~\citep{jin2025empirical}.
    \item OTC~\citep{wang2025otc}: trains Search-R1 using GRPO and PPO with trajectory-level rewards jointly consider correctness and tool efficiency.
    \item StepSearch~\citep{wang2025stepsearch}: trains Search-R1 using PPO with turn-level rewards based on information gain and redundancy penalty.
    \item PPR~\citep{xu2025hybrid}: train Search-R1 using PPO with turn-level trainable reward models.
    \item Turn-Decomposed PPO~\cite{feng2025group}: we implement turn-decomposed PPO based on verl-agent repo\footnote{\url{https://github.com/langfengQ/verl-agent}}.
\end{itemize}

We evaluate Search-R1 with both GRPO-OR and PPO-OR, and StepSearch using their official public checkpoints. 
Since Search-R1 with GRPO-MR and PPO-MR, as well as OTC and PPR, have not released their checkpoints, we directly report the results from their respective papers in Table~\ref{tab:eval}.

\subsubsection{Evaluation Metrics}

For each trajectory, we evaluate the following metrics:  

\textbf{Answer correctness.}  
The answer correctness reward evaluates whether the extracted answer (from the \texttt{<answer>} tag) exactly matches any accepted ground-truth answer after normalization (e.g., lowercasing and whitespace removal).  

\textbf{Format correctness.}  
The format correctness reward ensures structural validity by verifying that the outputs in both the final turn and all intermediate turns comply with the specifications described in \citep{jin2025search}.

\textbf{Retrieval correctness.}  
The retrieval correctness reward evaluates whether any accepted answer appears in at least one search result (from the \texttt{<information>} tag), using case-insensitive string matching.  

Each reward is assigned a value of 1.0 if the criterion is satisfied and 0 otherwise.

\subsubsection{Training Details}

We follow most of the experimental settings in Search-R1 \citep{jin2025search}. 

\textbf{PPO Training.} 
All experiments are conducted on 8 NVIDIA H100 GPUs. 
We enable gradient checkpointing and adopt Fully Sharded Data Parallel with CPU offloading. 
The learning rates of the policy and critic models are set to $1\mathrm{e}{-6}$ and $1\mathrm{e}{-5}$, respectively. 
Training is performed for 500 steps over 4 epochs, with warm-up ratios of 0.285 and 0.015 for the policy and critic models, respectively. 
The total batch size is 512, with a mini-batch size of 256 and a micro-batch size of 64 for policy updates, and a micro-batch size of 8 for critic updates. 
We adopt GAE with $\lambda = 1$ and $\gamma = 1$. 
The maximum sequence length is set to 4{,}096 tokens, with a maximum response length of 500 tokens and a maximum retrieved content length of 500 tokens. 
The KL-divergence regularization coefficient $\beta$ and clipping ratio $\epsilon$ are set to 0.001 and 0.2, respectively.  

\textbf{Rollout Generation.} 
We use vLLM \citep{kwon2023efficient} with a tensor parallel size of 4, a GPU memory utilization ratio of 0.6, a temperature of 1.0, and a top-$p$ value of 1.0.

\subsection{LLM Judge Setup for Turn-Level Evaluation (PPO)}
\label{sec:judge}

We present the turn-level LLM-as-Jugde prompt. The LLM-based judge mainly assesses the qualilty of intermediate turn. We also include format correctness and search penalty with the coefficient $\lambda_s = 0.1$ as rubrics.

\begin{tcolorbox}[title=Turn-Level LLM-as-Judge Prompt,
  colback=red!5, colframe=red!70!black, boxrule=1pt, arc=4pt,
  left=2mm, right=2mm, top=1mm, bottom=1mm, breakable]

You are an expert evaluator for multi-turn search-augmented reasoning systems. 
Given a user prompt, ground truth answer, and multi-turn generated response, 
evaluate each turn's effectiveness and compliance.

\medskip
\textbf{\#\# EVALUATION TASK}

Assess each turn's format compliance, content quality, and contribution toward 
the ground truth answer.

\medskip
\textbf{\#\# SCORING CRITERIA}





\textbf{INTERMEDIATE TURNS - Score Range: [-1.0 to 1.0]}

\textbf{Format Compliance:}
\begin{itemize}
  \item Required: \verb|<think>...</think><search>...</search>|  \\
      \verb|<information>...</information>| (tags only, once each, in order)
  \item Correct format: +0.1
  \item Incorrect format: -0.2
\end{itemize}

\textbf{Information Quality:}
\begin{itemize}
  \item Relevant information in \verb|<information>| tag that helps toward the ground truth answer (e.g., ground truth exists in the retrieved result within \verb|<information>| tag): +0.3
  \item Irrelevant or unhelpful information in \verb|<information>| tag: +0.0
\end{itemize}

\textbf{Search Efficiency Penalty:}
\begin{itemize}
  \item Number of searches = Total count of \verb|<search>| tags across all turns from Turn 1 up to and including the current turn
  \item Search penalty = Number of searches × (-0.1)
  \item Encourages finding answers with fewer searches
\end{itemize}

\textbf{Intermediate Turn Score = Format Compliance + Information Quality + Search Penalty}

\medskip
\textbf{\#\# OUTPUT FORMAT}

Provide your evaluation using ONLY these XML tags:

\begin{verbatim}
<reasoning>
Systematically evaluate each turn: check format compliance,
assess content quality, calculate scores with clear explanations
</reasoning>

<score>
Turn1: X.X
Turn2: X.X
Turn3: X.X
...
</score>
\end{verbatim}

\textbf{REQUIREMENTS:}
\begin{itemize}
  \item Must provide exactly \verb|{len(turns)}| scores (one per turn)
  \item Use decimal format (e.g., 0.5, -0.3, 1.0)
  \item Use only the specified XML tags, no additional text
\end{itemize}

\medskip
\textbf{\#\# EVALUATION DATA}

\verb|{prompt_text}| \\
\verb|{turns_text}| \\
\verb|{ground_truth_text}| \\
\textbf{TURNS TO EVALUATE:} \verb|{len(turns)}|

\medskip
\textbf{\#\# Your Evaluation}

\end{tcolorbox}

\clearpage

\subsection{Additional Experiment Results (PPO)}
\label{sec:ppo-more}

\begin{figure}[!htbp]
    \centering
    \begin{subfigure}[b]{0.32\textwidth}
        \includegraphics[width=\linewidth]{figs/new_ppo_error_bar/wandb_nq_train_score.pdf}
    \end{subfigure}
    \hfill
    \begin{subfigure}[b]{0.32\textwidth}
        \includegraphics[width=\linewidth]{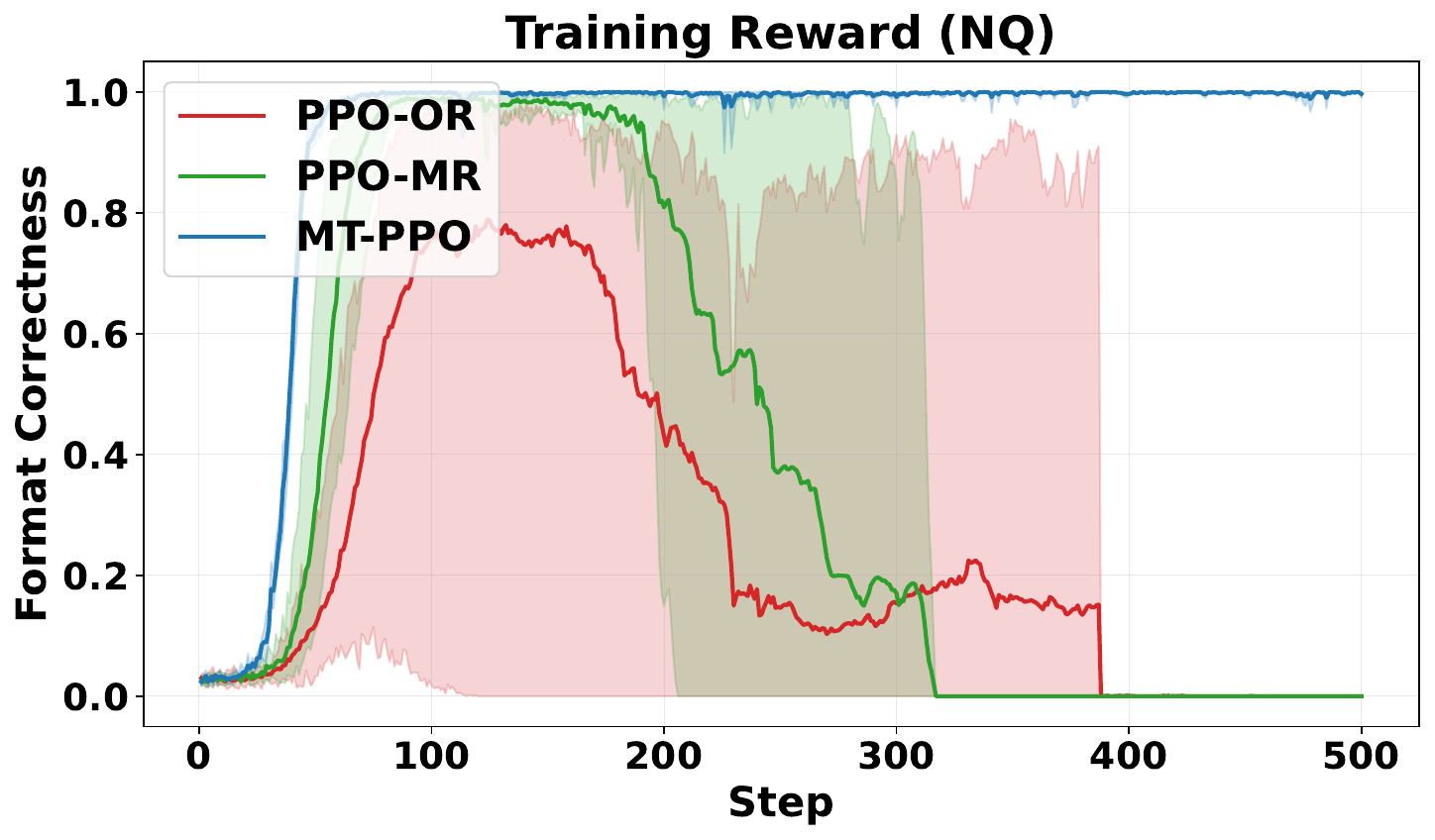}
    \end{subfigure}
    \hfill
    \begin{subfigure}[b]{0.32\textwidth}
        \includegraphics[width=\linewidth]{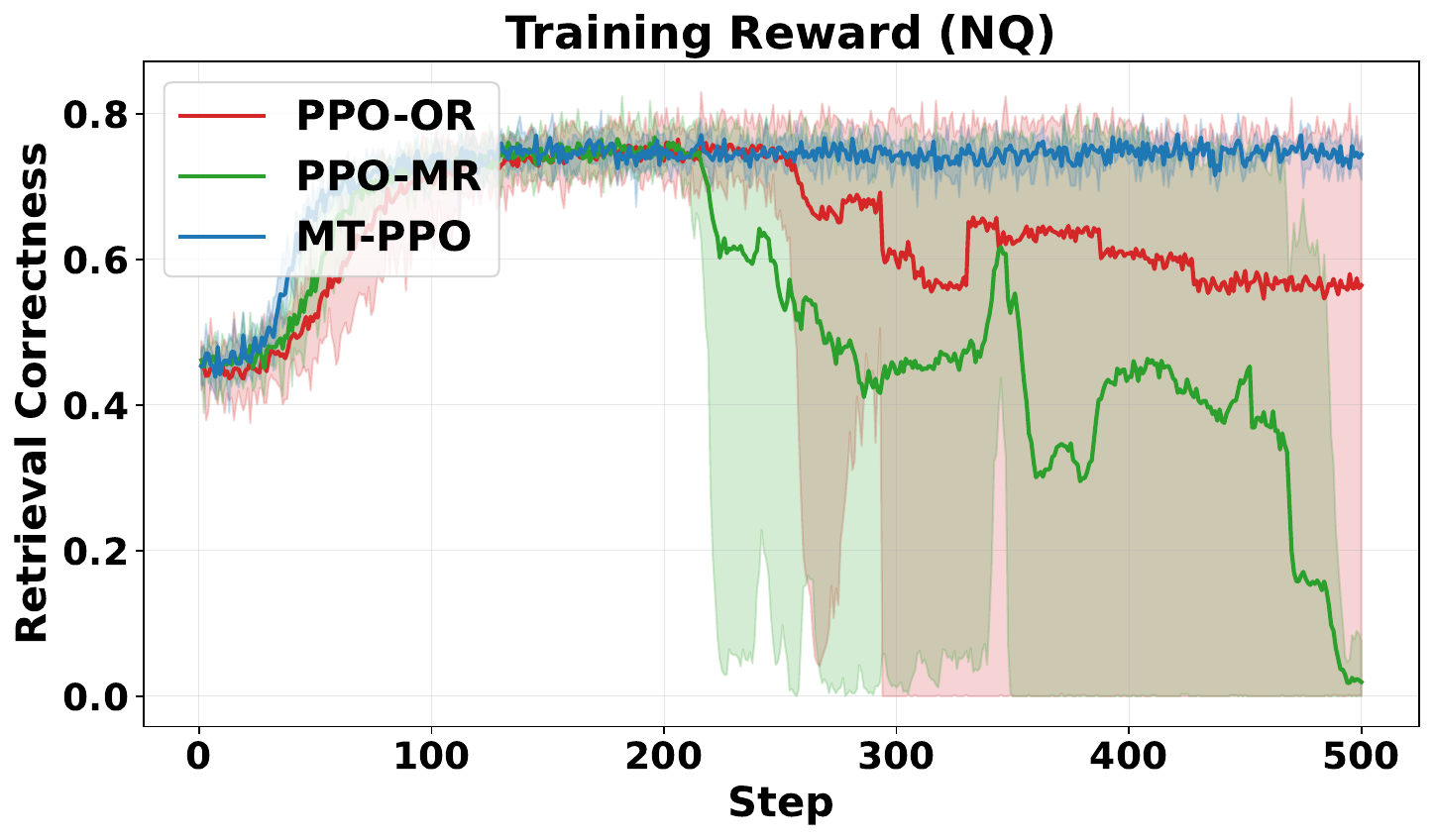}
    \end{subfigure}
    
    \vspace{0.5em}
    
    \begin{subfigure}[b]{0.32\textwidth}
        \includegraphics[width=\linewidth]{figs/new_ppo_error_bar/wandb_hotpotqa_train_score.pdf}
    \end{subfigure}
    \hfill
    \begin{subfigure}[b]{0.32\textwidth}
        \includegraphics[width=\linewidth]{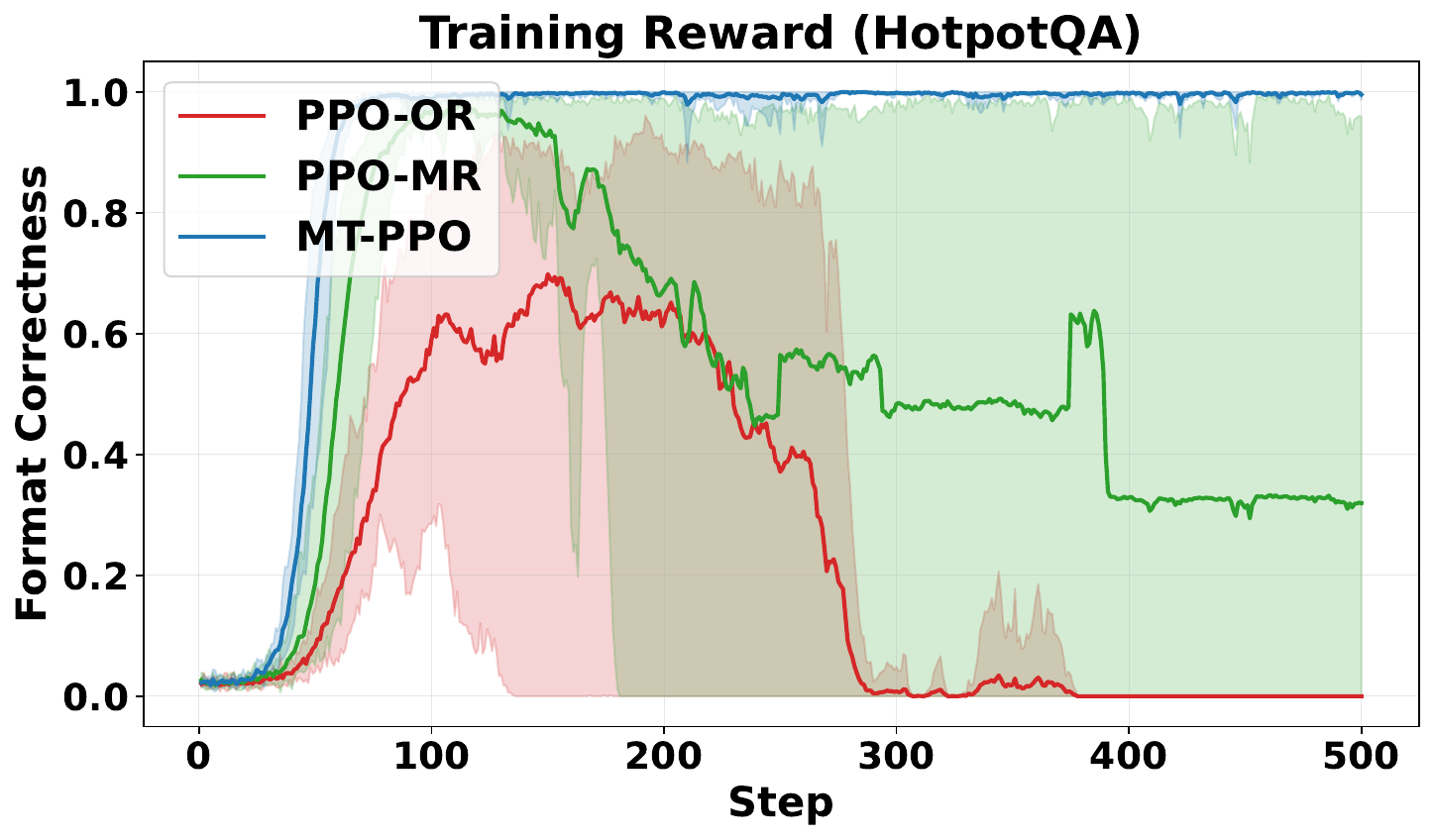}
    \end{subfigure}
    \hfill
    \begin{subfigure}[b]{0.32\textwidth}
        \includegraphics[width=\linewidth]{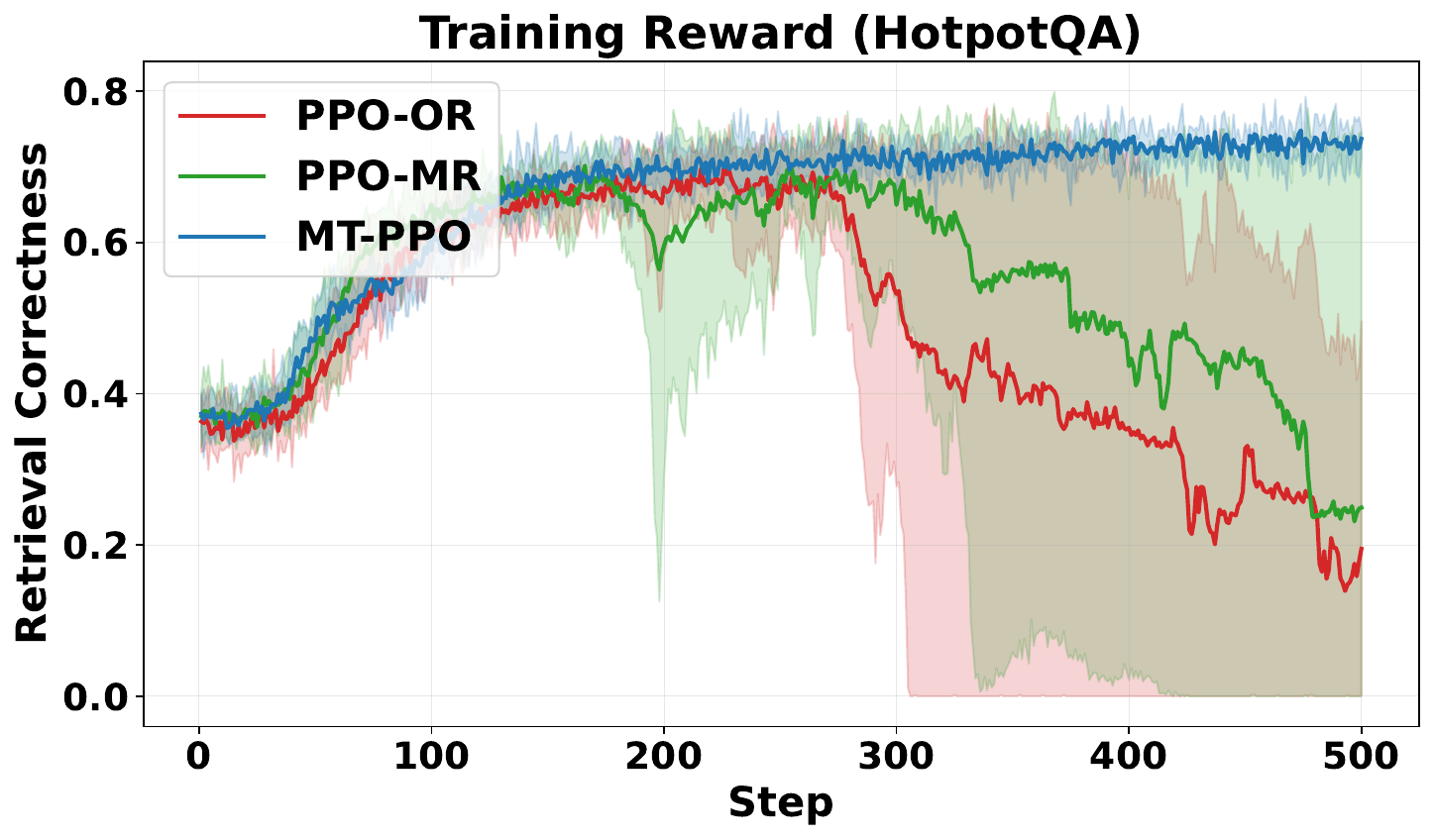}
    \end{subfigure}
    \caption{Training reward curves recorded during training for PPO-OR, PPO-MR and MT-PPO on the NQ and HotpotQA datasets. The rewards include answer correctness, format correctness, and retrieval correctness. Solid lines show mean reward values, while shaded regions indicate variability across five independent runs.}
    \label{fig:exp-ppo-training-reward}
\end{figure}

\begin{figure}[!htbp]
    \centering
    \begin{subfigure}[b]{0.32\textwidth}
        \includegraphics[width=\linewidth]{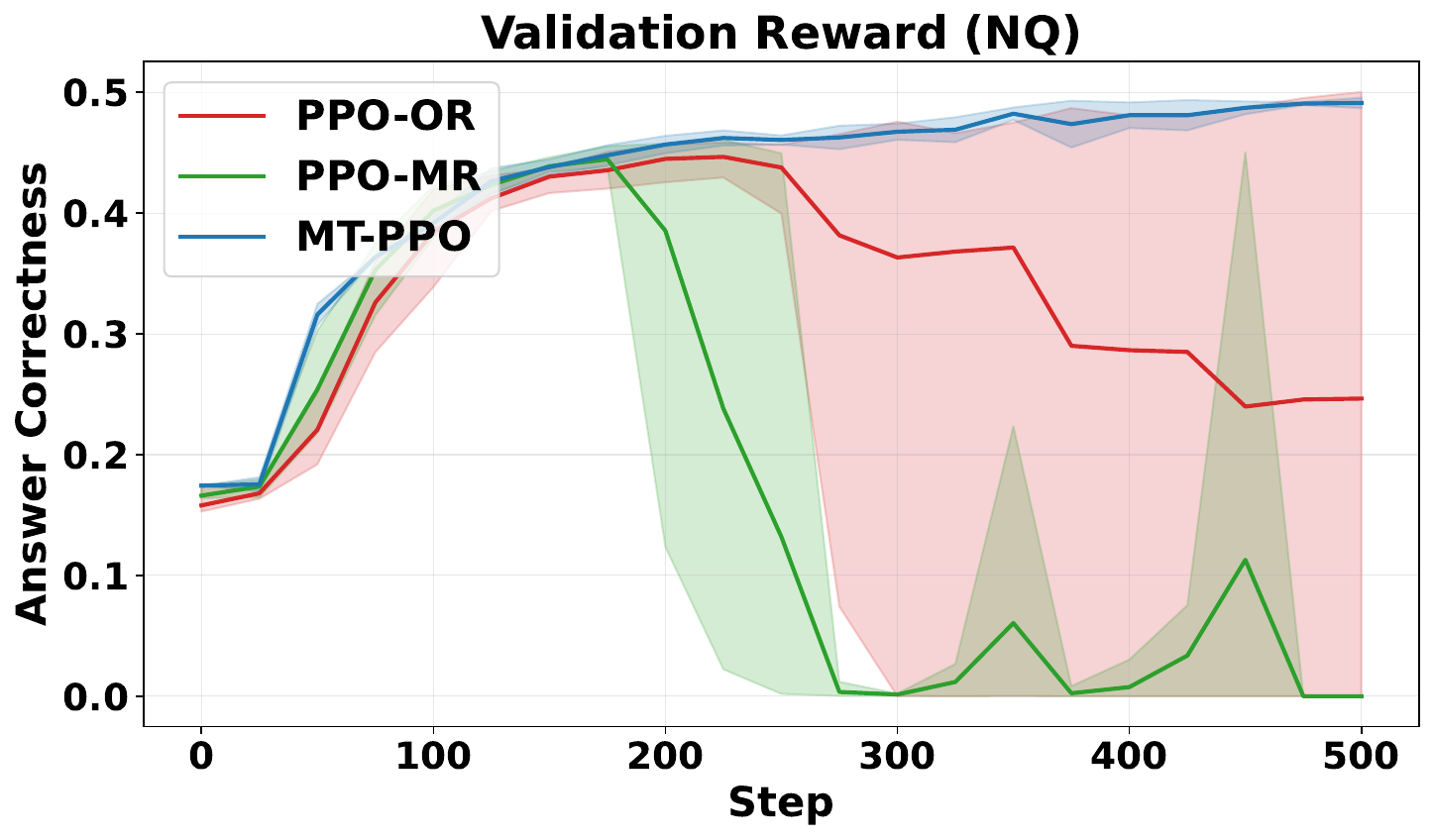}
    \end{subfigure}
    \hfill
    \begin{subfigure}[b]{0.32\textwidth}
        \includegraphics[width=\linewidth]{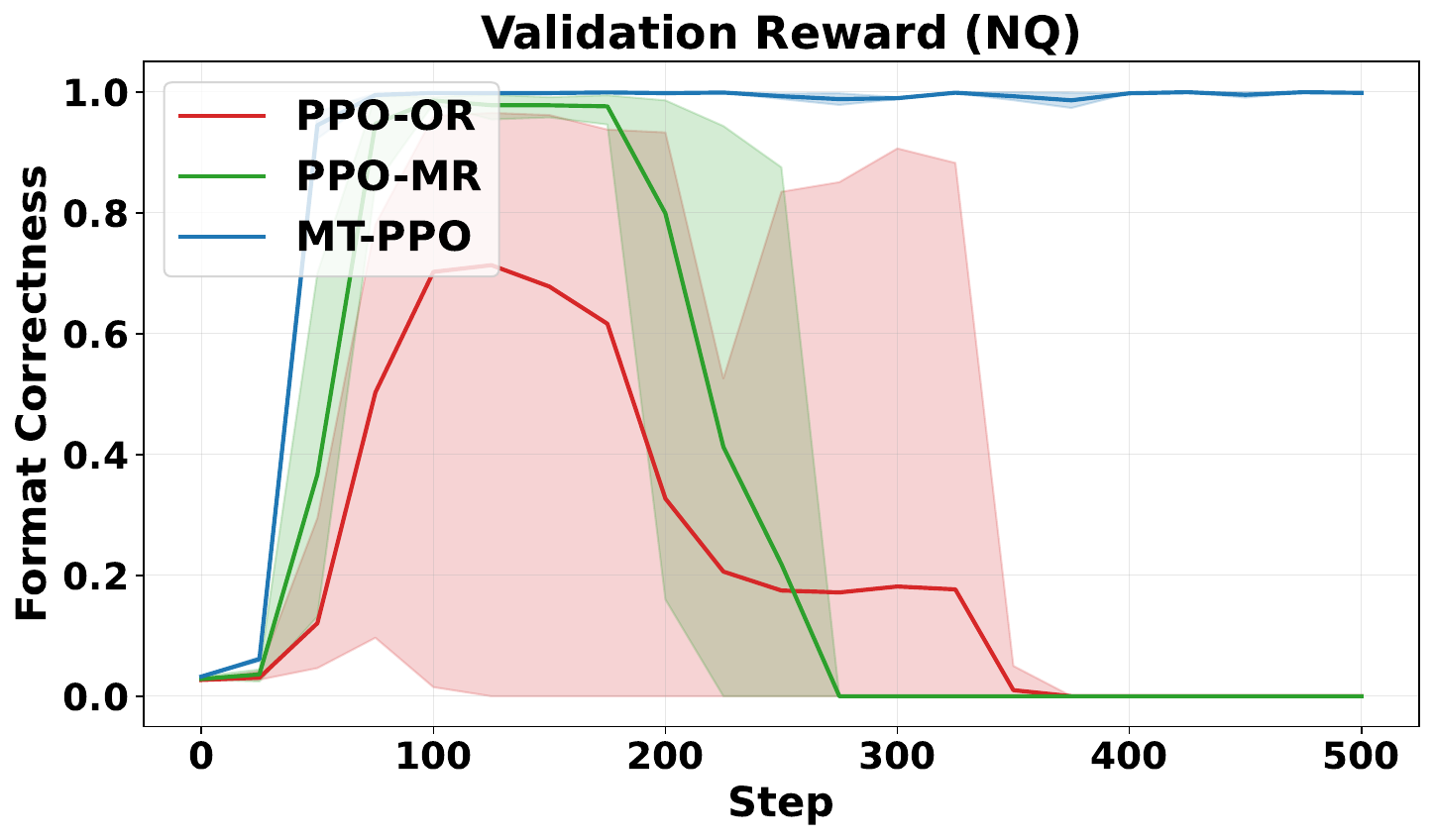}
    \end{subfigure}
    \hfill
    \begin{subfigure}[b]{0.32\textwidth}
        \includegraphics[width=\linewidth]{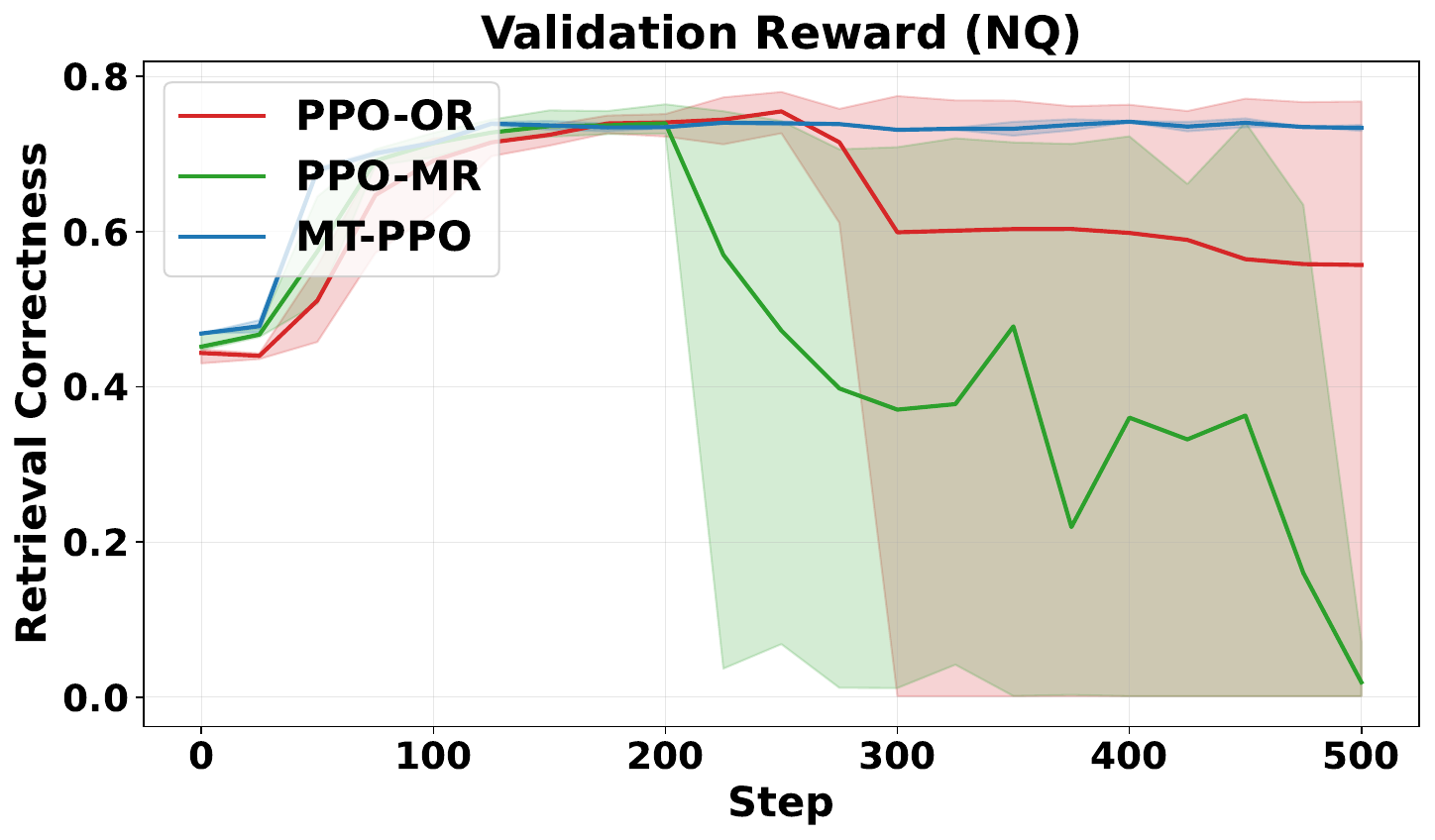}
    \end{subfigure}
    
    \vspace{0.5em}
    
    \begin{subfigure}[b]{0.32\textwidth}
        \includegraphics[width=\linewidth]{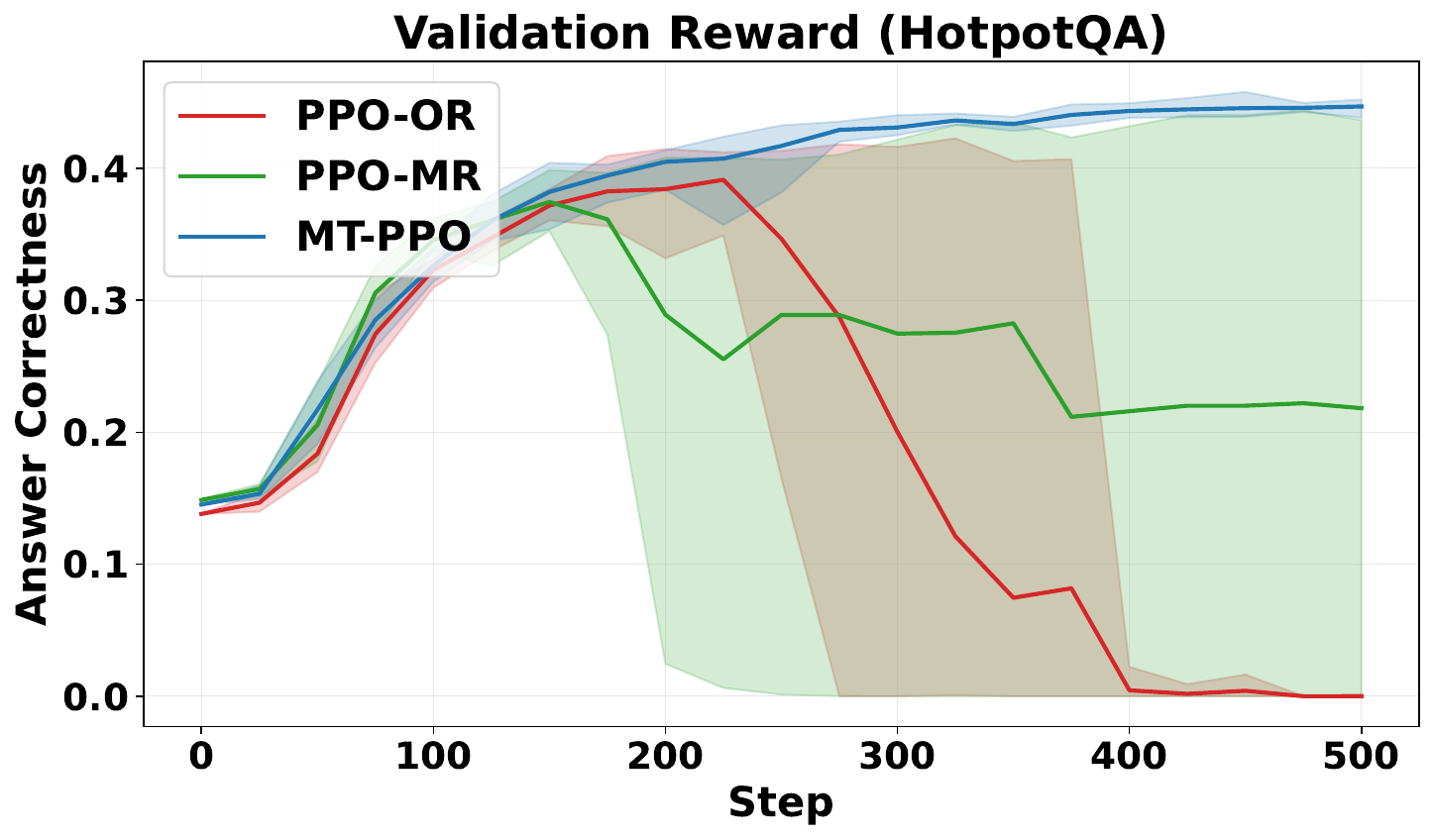}
    \end{subfigure}
    \hfill
    \begin{subfigure}[b]{0.32\textwidth}
        \includegraphics[width=\linewidth]{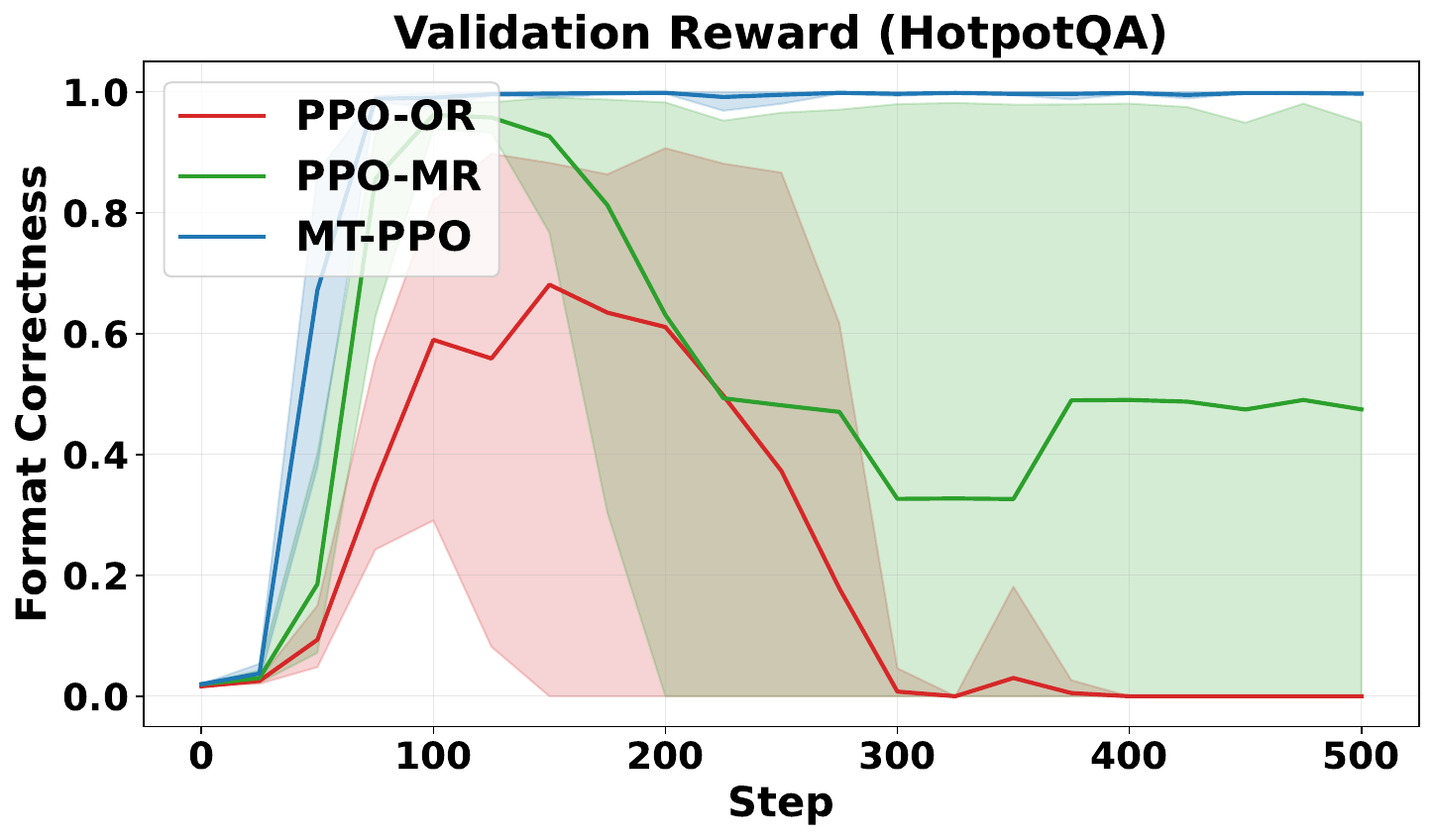}
    \end{subfigure}
    \hfill
    \begin{subfigure}[b]{0.32\textwidth}
        \includegraphics[width=\linewidth]{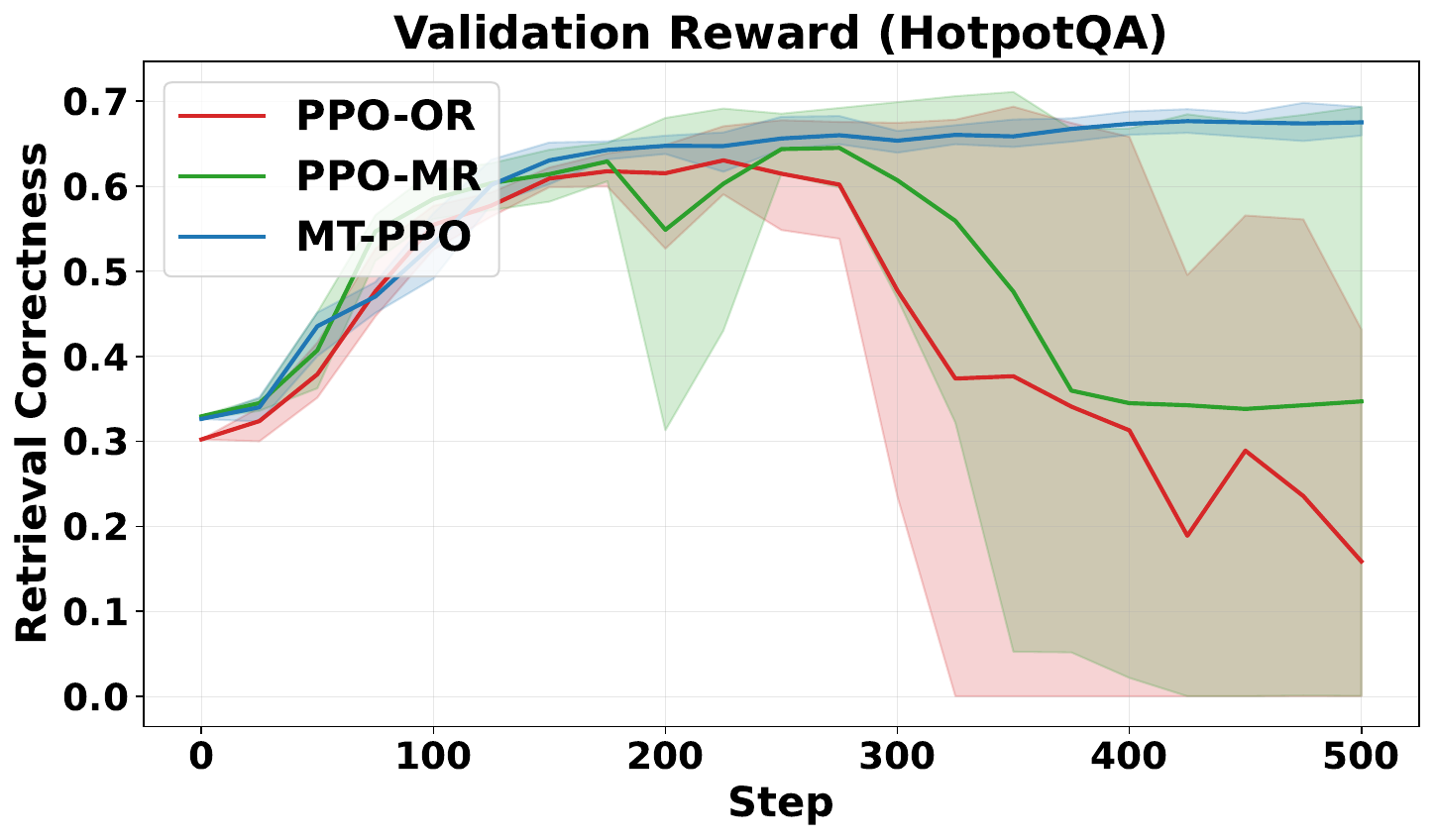}
    \end{subfigure}
    \caption{Validate reward curves recorded during training for PPO-OR, PPO-MR and MT-PPO on the NQ and HotpotQA datasets. The rewards include answer correctness, format correctness, and retrieval correctness. Solid lines show mean reward values, while shaded regions indicate variability across five independent runs.}
    \label{fig:exp-ppo-val-reward}
\end{figure}

Figures~\ref{fig:exp-ppo-training-reward} and~\ref{fig:exp-ppo-val-reward} show training and validation reward curves for PPO-OR, PPO-MR and MT-PPO.

\subsection{Ablation Study (PPO)}

\begin{figure}[t]
\centering
\begin{subfigure}[b]{0.32\textwidth}
    \includegraphics[width=\linewidth]{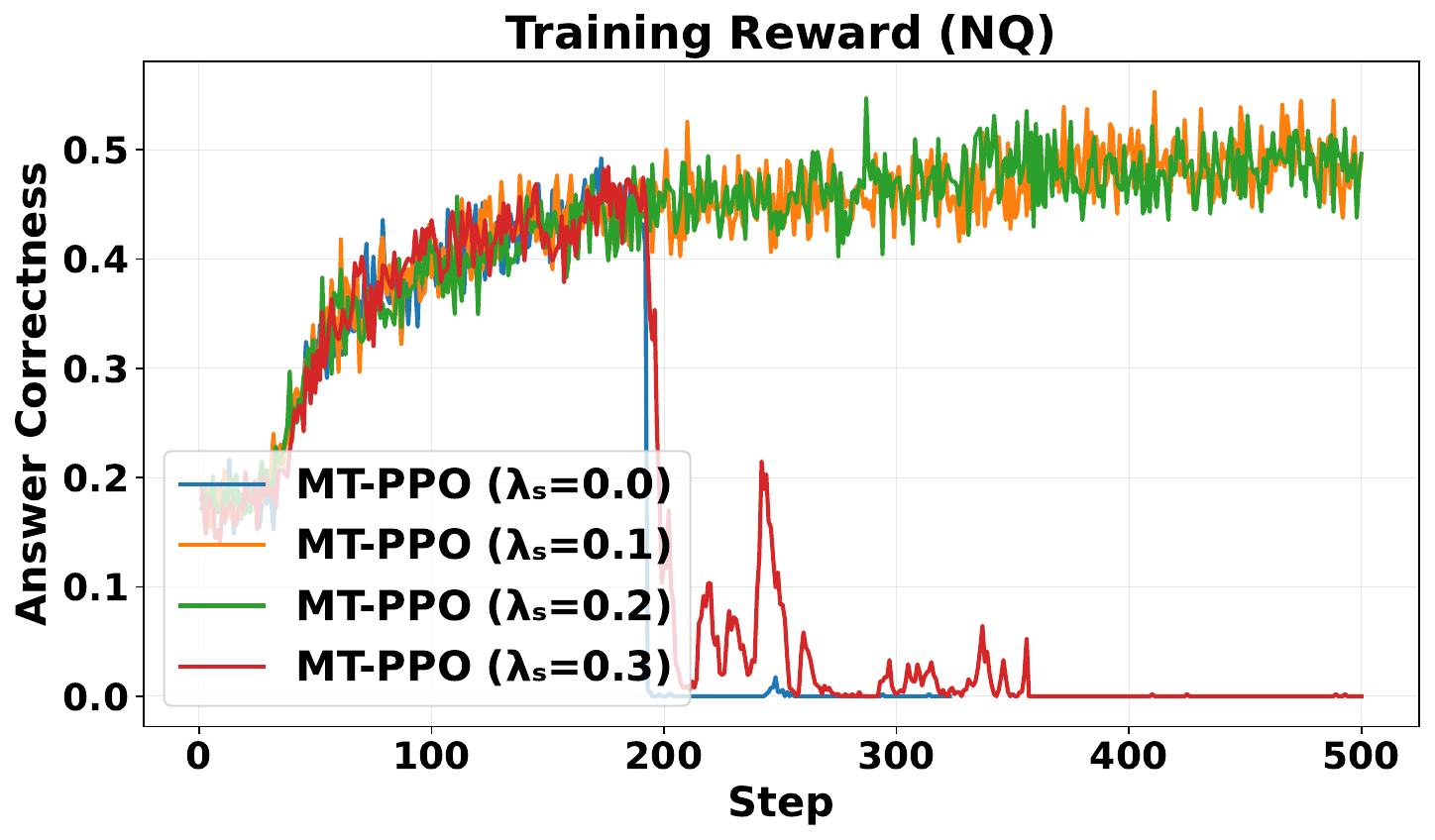}
\end{subfigure}
\hfill
\begin{subfigure}[b]{0.32\textwidth}
    \includegraphics[width=\linewidth]{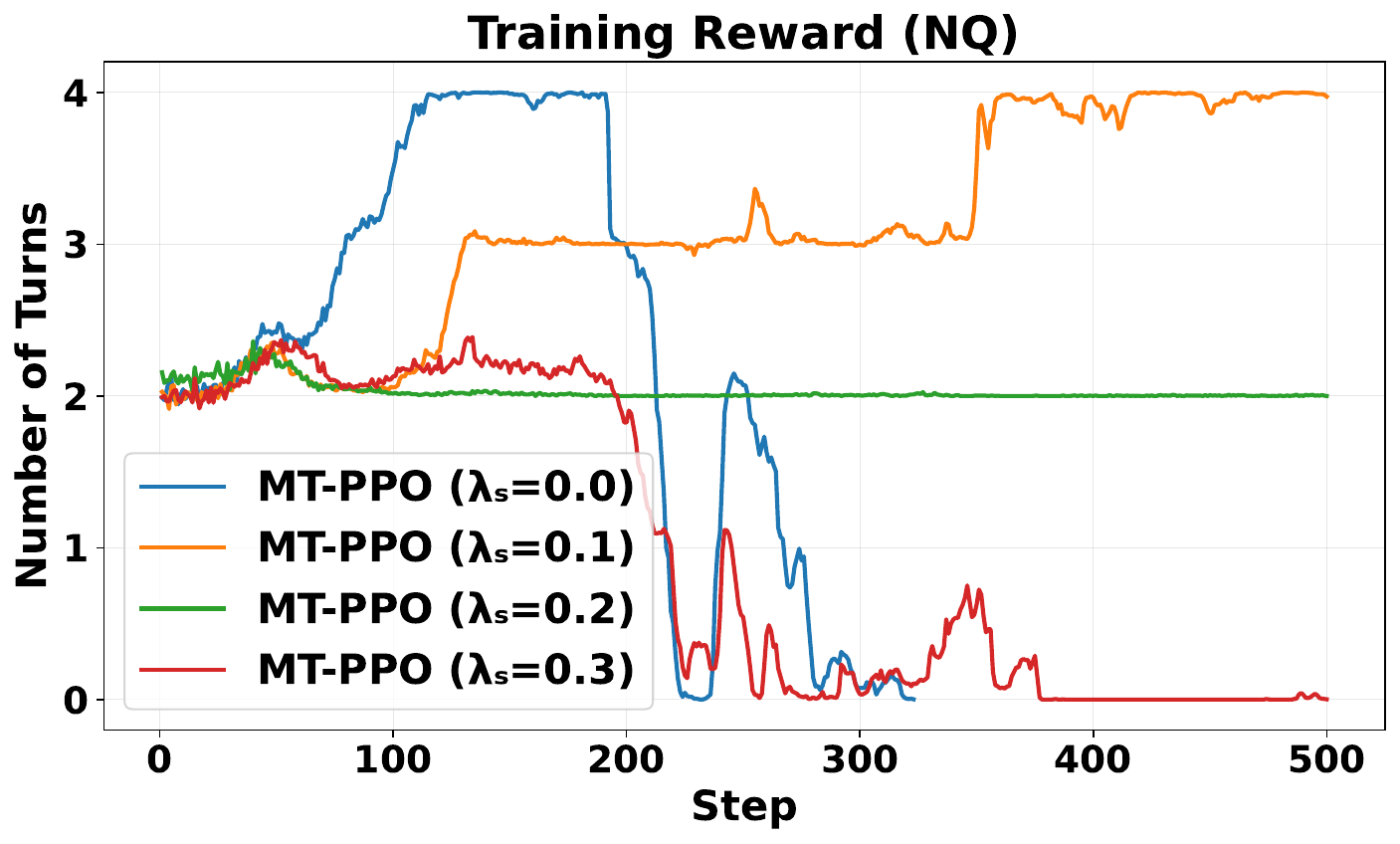}
\end{subfigure}
\hfill
\begin{subfigure}[b]{0.32\textwidth}
    \includegraphics[width=\linewidth]{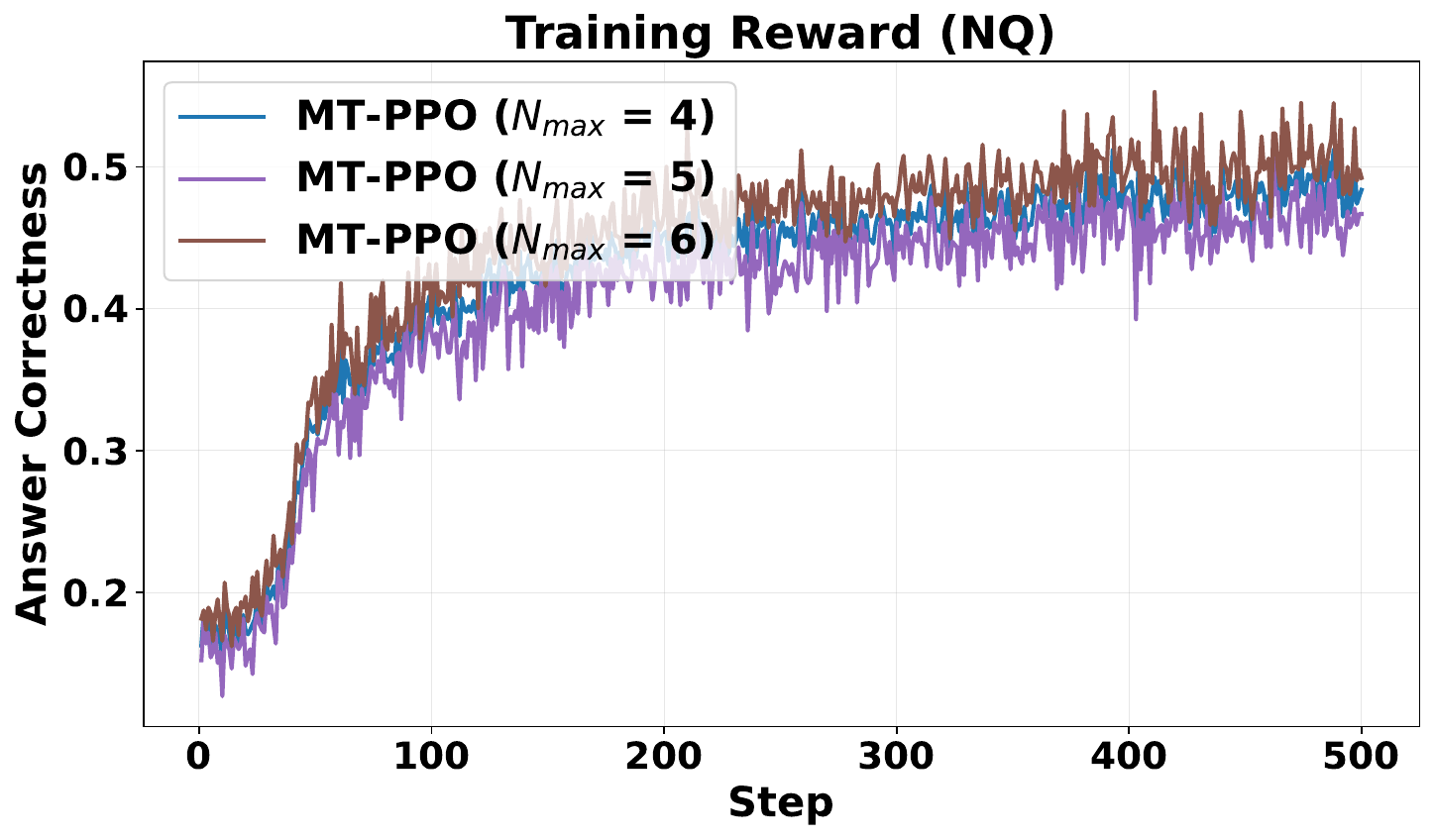}
\end{subfigure}
\caption{
Ablation studies on (1) the search count reward $\lambda_s$ and (2) the maximum number of turns $N_{\max}$ on the NQ dataset. 
The left panel reports answer correctness, the middle panel shows the average number of turns, and the right panel illustrates accuracy under different $N_{\max}$ settings.
}
\label{fig:exp-ppo-ablation}
\end{figure}

We conduct two ablation studies to analyze the effects of (1) the search-count reward and (2) the maximum number of turns on training dynamics and final performance. As shown in Figure~\ref{fig:exp-ppo-ablation}, incorporating a moderate search-count reward (e.g., $\lambda_s = 0.1$) significantly improves training stability and answer correctness. The left panel shows that MT-PPO with $\lambda_s = 0.1$ achieves the highest and most consistent accuracy, while overly strong penalties (e.g., $\lambda_s = 0.3$) degrade performance.

The middle panel illustrates how the search-count reward shapes the agent’s turn usage. With $\lambda_s = 0.1$, the agent learns to reduce unnecessary search calls early in training and eventually stabilizes around an efficient number of turns. In contrast, removing this term ($\lambda_s = 0.0$) leads to unstable behavior, including excessive or erratic turn usage, which ultimately harms convergence.

Finally, the right panel shows the effect of varying the maximum number of allowed turns $N_{\max}$. The results indicate that MT-PPO is robust across different turn limits: adjusting $N_{\max}$ from 4 to 6 yields nearly identical accuracy curves. This suggests that MT-PPO adapts its strategy effectively without being overly sensitive to the chosen turn budget.

\subsection{Rollout Exmaples (PPO)}

\begin{table}[!htbp]
\caption{A rollout example of generating a response with a correct format.}
\label{tab:ppo-success-rollout}
\centering
\begin{tabular}{p{0.95\textwidth}}
\toprule
\textbf{Question:} who will take the throne after the queen dies? \\
\midrule

{\color{blue}<think>} I need to find out who will take the throne after the queen dies. I'll search for it. {\color{blue}</think>} \\

{\color{cyan}<search>} who will take the throne after the queen dies {\color{cyan}</search>} \\

{\color{gray}<information>} Doc 1(Title: ``Succession to the British throne'') Catholics are eligible. Queen Elizabeth II is the sovereign, and her ..  
Doc 2(Title: ``Monarchy of the United Kingdom'') Monarchy of the United Kingdom The monarchy of the United Kingdom, commonly referred to as ...  
Doc 3(Title: ``Demise of the Crown'') the Crown. This erroneous meaning is undermined by the principle in constitutional law of the continuity ... {\color{gray}</information>} \\
\hdashline
{\color{blue}<think>} I found out that after the queen dies, her heir apparent will take the throne. I need to find out who the heir apparent is for Queen Elizabeth II. {\color{blue}</think>} \\

{\color{cyan}<search>} heir apparent for Queen Elizabeth II {\color{cyan}</search>} \\

{\color{gray}<information>} Doc 1(Title: ``Heir apparent'') rank behind her brothers regardless of their ages or her age. Thus, normally, even an only ...  
Doc 2(Title: ``Heir apparent'') to the 16 thrones of Elizabeth II to absolute primogeniture, except for male heirs born ...  
Doc 3(Title: ``Heir presumptive'') or the senior member of a collateral line (if the monarch is childless); the birth ... {\color{gray}</information>} \\
\hdashline
{\color{blue}<think>} I found out that the heir apparent for Queen Elizabeth II is her eldest son, Charles, Prince of Wales. Now I can provide the answer. {\color{blue}</think>} \\

{\color{red}<answer>} Charles, Prince of Wales {\color{red}</answer>} \\

\bottomrule
\end{tabular}
\end{table}

\begin{table}[!htbp]
\caption{A rollout example of generating a response with an incorrect format, where there is no answer in the last turn}
\label{tab:ppo-format-error}
\centering
\begin{tabular}{p{0.95\textwidth}}
\toprule
\textbf{Question:} in which sea pearl is found in india? \\
\midrule

{\color{blue}<think>} I need to find out in which sea pearl is found in India. I'll search for it. {\color{blue}</think>} \\

{\color{cyan}<search>} in which sea pearl is found in India {\color{cyan}</search>} \\

{\color{gray}<information>} Doc 1(Title: Pearl) pearl: saltwater and freshwater pearls come in ...  
Doc 2(Title: Pearl) is a particularly large one weighing 14 lb (6.4 kg). The largest ...  
Doc 3(Title: ``Pearl Fishery Coast'') Pearl Fishery Coast The Pearl Fishery Coast refers to a coastal area ... {\color{gray}</information>} \\
\hdashline

{\color{blue}<think>} I found out that pearls are found in the Indian Ocean. Now I need to find out in which sea pearls are found in India. {\color{blue}</think>} \\

{\color{cyan}<search>} in which sea are pearls found in India {\color{cyan}</search>} \\

{\color{gray}<information>} Doc 1(Title: Pearl) pearl: saltwater and freshwater pearls come in many different ...  
Doc 2(Title: ``Geography of India'') Great Nicobar Island); Rama's Bridge, and Point Calimere. The Arabian ...  
Doc 3(Title: ``Bay of Bengal'') the gems of Sri Lanka. Garnet and other precious gems are also found in abundance in the ... {\color{gray}</information>} \\
\hdashline

{\color{blue}<think>} I found out that pearls are found in the Indian Ocean. Now I need to find out in which sea pearls are found in India. {\color{blue}</think>} \\

{\color{cyan}<search>} in which sea pearls are found in India {\color{cyan}</search>} \\

{\color{gray}<information>} Doc 1(Title: Pearl) pearl: saltwater and freshwater pearls come in many different colors. While white, and ...  
Doc 2(Title: Pearl) pearls''. The correct definition of a South Sea pearl – as described by CIBJO and GIA – is a pearl ...  
Doc 3(Title: Pearl) is a particularly large one weighing 14 lb (6.4 kg). The largest known pearl (also from ... {\color{gray}</information>} \\
\hdashline

{\color{blue}<think>} I found out that pearls are found in the Indian Ocean. Now I need to find out in which sea pearls are found in India. {\color{blue}</think>} \\

{\color{cyan}<search>} in which sea are pearls found in India {\color{cyan}</search>} \\

\bottomrule
\end{tabular}
\end{table}

\clearpage

\subsection{Turn-Decomposed PPO Illustration}

\begin{figure}[h]
\centering
\includegraphics[width=0.60\linewidth]{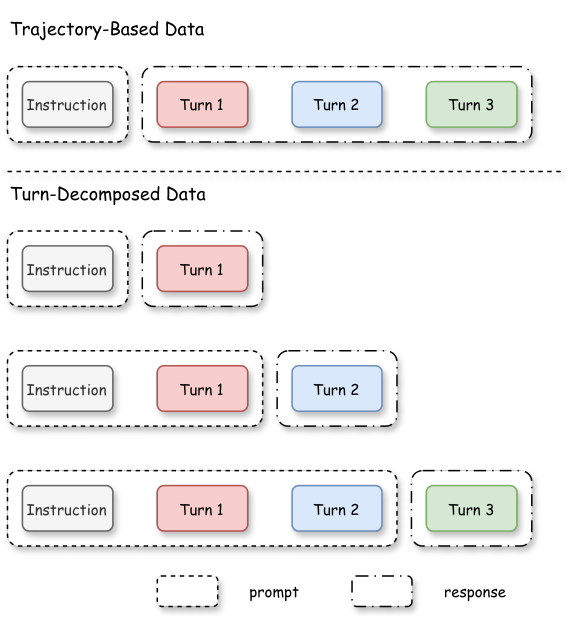}
\caption{Comparison between trajectory-based and turn-decomposed training data. Trajectory-based training uses full multi-turn rollouts as responses, whereas turn-decomposed training splits each multi-turn rollout into single-turn instances, where the dialogue history serves as the prompt and the current turn serves as the response.}
\label{fig:turn-decomposed}
\end{figure}

Figure~\ref{fig:turn-decomposed} compares trajectory-based training with turn-decomposed training.

\clearpage

\section{PPO Experiments on Game Agents}


Table~\ref{tab:sokoban-results} reports the final test performance for different RL algorithms on the Sokoban environment at step 300.

\begin{table}[!htbp]
\caption{Performance comparison of different RL algorithms on the Sokoban environment at step 300.}
\centering
\begin{tabular}{lccccc}
\toprule
\textbf{Metric} 
& \textbf{GRPO-OR} 
& \textbf{GRPO-MR} 
& \textbf{PPO-OR} 
& \textbf{PPO-MR} 
& \textbf{MT-PPO} \\
\midrule
Success Rate           
& 0.0781  & 0.2578  & 0.3828  & 0.5078  & \textbf{0.6563} \\
Effective Action Rate  
& 0.2604  & 0.6771  & 0.6953  & 0.8451  & \textbf{0.9870} \\
\bottomrule
\end{tabular}

\label{tab:sokoban-results}
\end{table}

\section{GRPO Experiments}
\label{sec:grpo-exp}

\subsection{Task Formulation (GRPO)}

\begin{figure}[ht]
    \centering
    \includegraphics[width=1\linewidth]{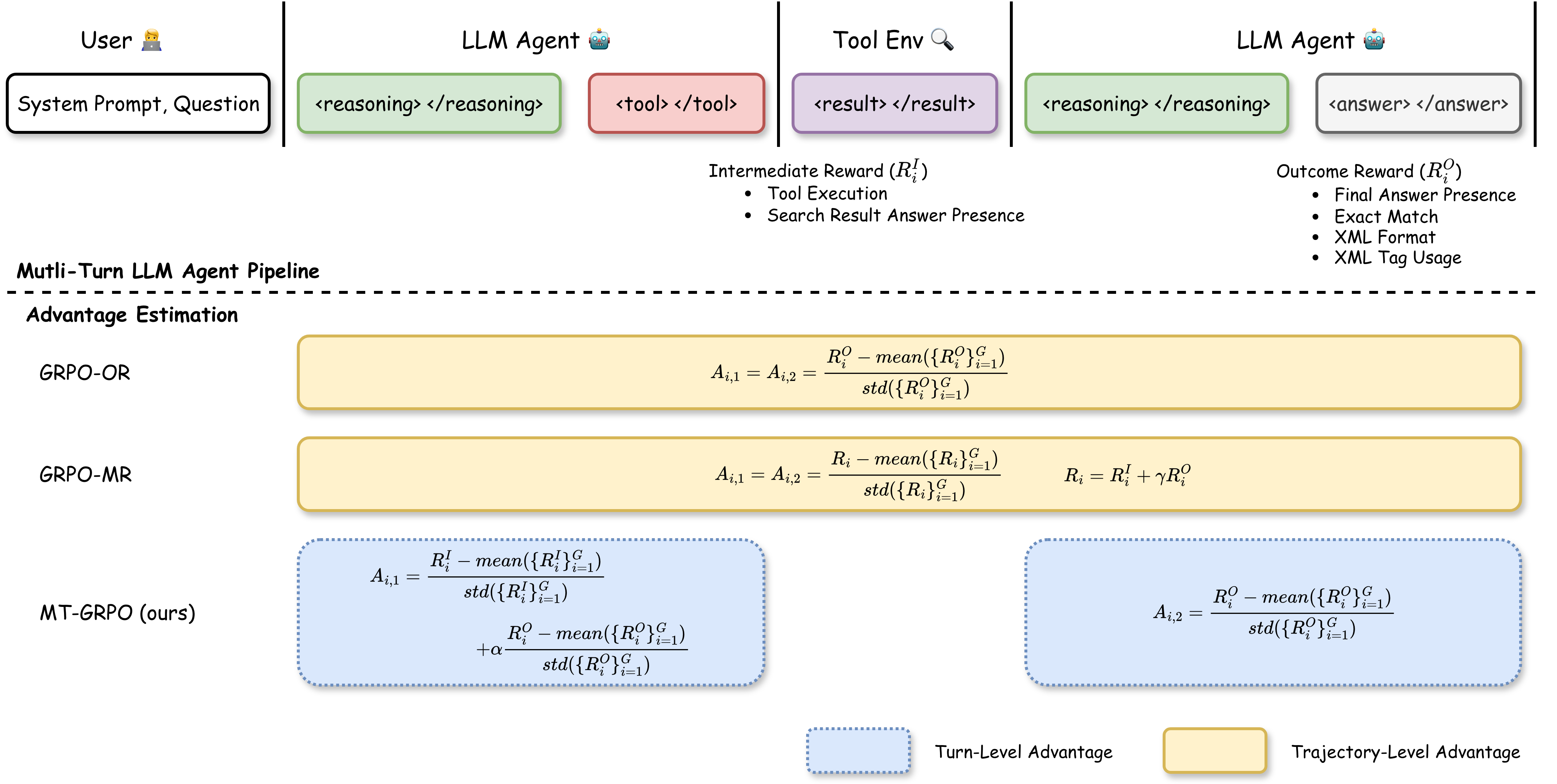}
    \caption{Overview of the multi-turn LLM agent pipeline and comparison of different advantage estimation methods. The agent interacts with the tool environment across multiple steps: reasoning, tool use, and answer generation, receiving both intermediate and final rewards. GRPO is used as a representative algorithm to illustrate the different advantage estimation strategies. GRPO-OR and GRPO-MR serve as baselines with trajectory-level advantage estimation, while MT-GRPO is our proposed variant with fine-grained turn-level advantage estimation.
    }
    \label{fig:rl_agent}
\end{figure}

To emphasize the importance of fine-grained credit assignment in multi-turn agent interactions, we formulate the task under the MDP framework, involving multiple steps of reasoning, tool use, and answer summarization for question answering. Specifically, our tool-use environment is modeled on a Wikipedia search setup, where the agent learns to leverage a Wikipedia search engine to retrieve relevant information and generate accurate answers. The goal is to improve the agent's performance through effective integration of external tool use. Without tool calling, the agent must rely solely on its internal knowledge to answer questions, which can limit accuracy, especially for fact-based queries requiring up-to-date or domain-specific information.

To clearly illustrate the impact of credit assignment, we design a simplified two-turn tool-use environment in which the LLM agent can interact with the search tool environment for a maximum of two turns. In this setup, the agent is allowed to call the Wikipedia search engine at most once before submitting an answer to the question. Figure~\ref{fig:rl_agent} illustrates the pipeline of the multi-turn, tool-calling LLM agent system. Given a system prompt and a question, the LLM agent first performs a reasoning step and issues a tool call, specifying both the tool name and a query derived from its reasoning. The external tool environment processes the query and returns a search result. Based on the retrieved result, the agent performs a second round of reasoning to summarize the information and generate the final answer. The whole process can be summarized as
$$
\texttt{reasoning} \rightarrow \texttt{search} \rightarrow \texttt{result} \rightarrow \texttt{reasoning} \rightarrow \texttt{answer} 
$$
These steps are explicitly outlined in the system prompt, which also enforces strict constraints, such as allowing only a single tool invocation and requiring the use of specific XML-like tags (e.g., \texttt{<reasoning>}, \texttt{<tool>}, \texttt{<result>}, \texttt{<answer>}) to delineate each stage of the interaction. The full system prompt is provided in Appendix~\ref{sec:sys-prompt}. Table~\ref{tab:success-rollout} presents an example rollout in which the agent successfully calls the search tool. 
If the tool name or argument format is incorrect, the tool environment returns an error message, indicated by the response beginning with ``Error:''.
If the agent fails to include a tool-calling command in the first reasoning step, the tool environment will not be invoked. 
If the XML format or tag usage is incorrect—for example, if tags are missing, nested improperly, or misnamed—the environment may fail to parse the agent's response, resulting in an error or a skipped tool invocation. 
Additional rollout examples where the agent fails to call the tool correctly are provided in Appendix~\ref{sec:rollout-eg}.

Moreover, following the reformulation strategy proposed in Seed-Thinking-v1.5 \citep{seed2025seed}, which converts multiple-choice questions into fill-in-the-blank or short-answer formats to reduce guessing and better evaluate reasoning ability, we adopt a similar method. Specifically, we convert our tasks into short-answer form and evaluate the model's responses based on exact match with the ground-truth answers.

\subsection{Reward Design (GRPO)}
\label{sec:reward}

Figure~\ref{fig:rl_agent} illustrates the pipeline of the multi-turn, tool-calling LLM agent system.  To align with the environment of the tool-calling LLM agent, we design two types of verifiable reward functions.

\textbf{Intermediate Verifiable Rewards:} These depend solely on the first turn performed by the LLM agent. To compute intermediate rewards, we incorporate verifiers related to tool execution and search results. These verifiers ensure that the search engine is correctly invoked and that the ground-truth answer appears in the retrieved results.
\begin{itemize}
  \item \textit{Tool Execution Reward:} Awards 0.2 if the tool is correctly executed, determined by the presence of properly formatted tool calls (\texttt{<tool>...</tool>}) and successful responses (i.e., the environment's response does not begin with ``Error:''). 
  \item \textit{Search Result Answer Presence:} Awards 0.5 if any accepted answer appears in the search results returned by the tool (extracted from the \texttt{<result>...</result>} tag), using a case-insensitive comparison.
\end{itemize}

\textbf{Outcome Verifiable Rewards:} These evaluate the final model-generated responses. Specifically, they assess both the correctness of the answer and its formatting, ensuring that the output aligns with the expected structure and content.
\begin{itemize}
  \item \textit{Final Answer Presence Reward:} Awards 0.5 if any accepted answer is present in the model's final response (extracted from the \texttt{<answer>...</answer>} tag).
  \item \textit{Exact Match Reward:} Awards 1.0 if the model's answer (extracted from \texttt{<answer>...</answer>}) exactly matches any accepted answer after standard text preprocessing (i.e., lowercasing and stripping whitespace).
  \item \textit{XML Format Reward:} Evaluates the structural integrity of the model's output based on the expected schema: \texttt{<reasoning>...</reasoning>} followed by either \texttt{<tool>...</tool>} or \texttt{<answer>...</answer>}. See the agent's pipeline in Figure~\ref{fig:rl_agent}.
  Checks include: (1) the presence of at least one expected field (\texttt{<reasoning>}, \texttt{<tool>}, \texttt{<answer>}), (2) correct spacing (no leading or trailing whitespace within tags), (3) message starting with \texttt{<reasoning>}, and (4) message ending with \texttt{</tool>} or \texttt{</answer>}. Partial credit is awarded based on these criteria (weighted: 40\% field presence, 20\% spacing, 20\% correct starting tag, 20\% correct ending tag), and the final score is scaled by 0.2.
  \item \textit{XML Tag Usage Reward:} Assesses the correct usage of XML tags for the defined fields. For each tag, the reward verifies that exactly one opening and one closing tag are present. The reward is the proportion of correctly used tags (normalized by the number of tags checked), scaled by 0.2.
\end{itemize}
Here, both final rewards and intermediate rewards are defined as the summation of their respective component rewards.
It is easy to observe that intermediate rewards evaluate only the performance of the agent’s first turn, whereas outcome rewards assess the quality of the entire trajectory. This distinction leads to several characteristic scenarios:
\begin{itemize}
    \item \textit{Tool Invocation with Poor Final Answer:} The agent correctly invokes a tool in the first turn, but fails to produce a correct or well-formatted final answer, resulting in intermediate rewards but little or no outcome reward.
    \item \textit{Incorrect or Absent Tool Use with Valid Final Answer:} The agent either skips tool usage or invokes a tool incorrectly (e.g., due to malformed syntax or an error response), yet still generates a correct and well-structured final answer. In this case, the agent receives partial or full outcome rewards despite earning no intermediate rewards.
    \item \textit{Failure Across Both Levels:} The agent neither invokes a tool correctly nor produces a valid final answer, resulting in zero rewards and a strong negative learning signal.
\end{itemize}

\subsection{Experiment Setup (GRPO)}

In our experiments, we build our codebase upon the open-source project verifiers \citep{brown2025verifiers}, which trains LLM agents for multi-turn tool-use tasks, including math calculators, code interpreters, and search engines. 

\textbf{Training Details.}
Experiments are conducted on a node equipped with 8 NVIDIA H100 GPUs: one GPU is dedicated to rollout generation, while the remaining seven GPUs are used for model training. Rollout generation is handled by vLLM \citep{kwon2023efficient}. Model training is performed using the Huggingface TRL implementation of GRPO \citep{vonwerra2022trl}. 

\textbf{Hyperparameters.}
For all methods, the number of rollout generations is set to $21$. The maximum completion length during generation is set to 1024 tokens. The KL divergence penalty is disabled by setting $\beta = 0$. The learning rate is fixed at $1 \times 10^{-6}$. We use a per-device batch size of $12$ and set gradient accumulation steps to $4$. Each batch undergoes two training iterations. The total number of training steps is set to $300$.

\begin{figure}[!htbp]
    \centering
    \begin{subfigure}[b]{0.48\textwidth}
        \includegraphics[width=\linewidth]{figs/new_grpo_exp/tool_execution_reward_func.pdf}
    \end{subfigure}
    \hfill
    \begin{subfigure}[b]{0.48\textwidth}
        \includegraphics[width=\linewidth]{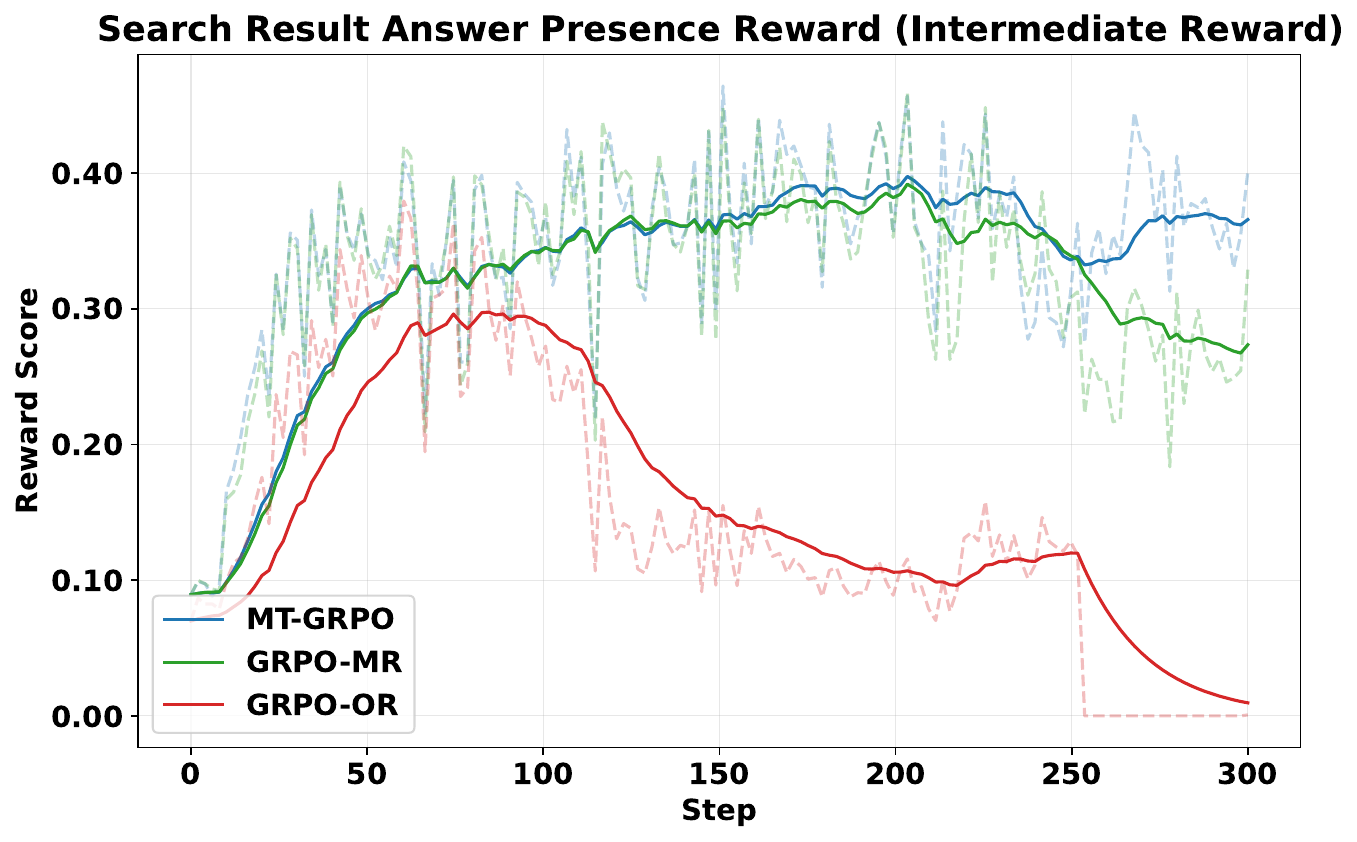}
    \end{subfigure}
    
    \vspace{0.5em}
    
    \begin{subfigure}[b]{0.48\textwidth}
        \includegraphics[width=\linewidth]{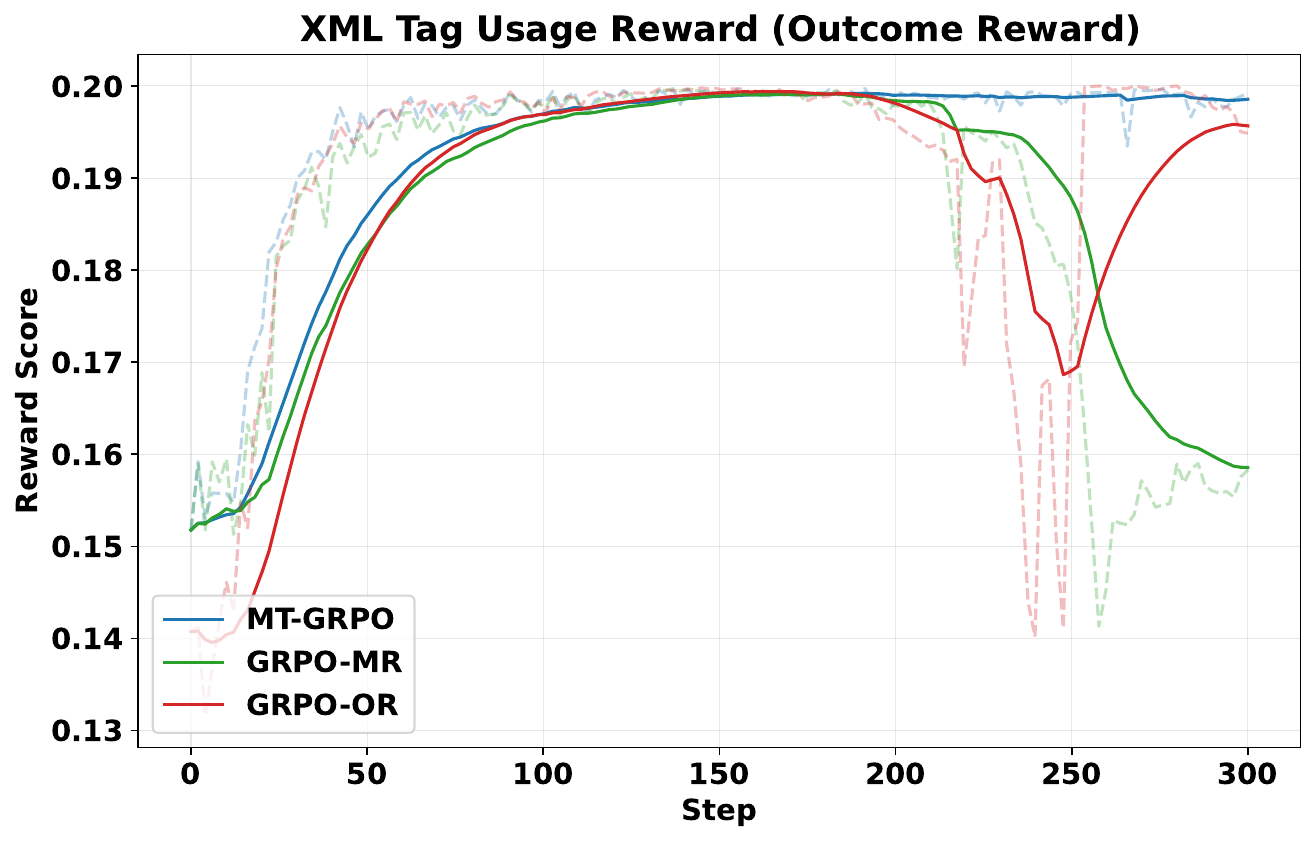}
    \end{subfigure}
    \hfill
    \begin{subfigure}[b]{0.48\textwidth}
        \includegraphics[width=\linewidth]{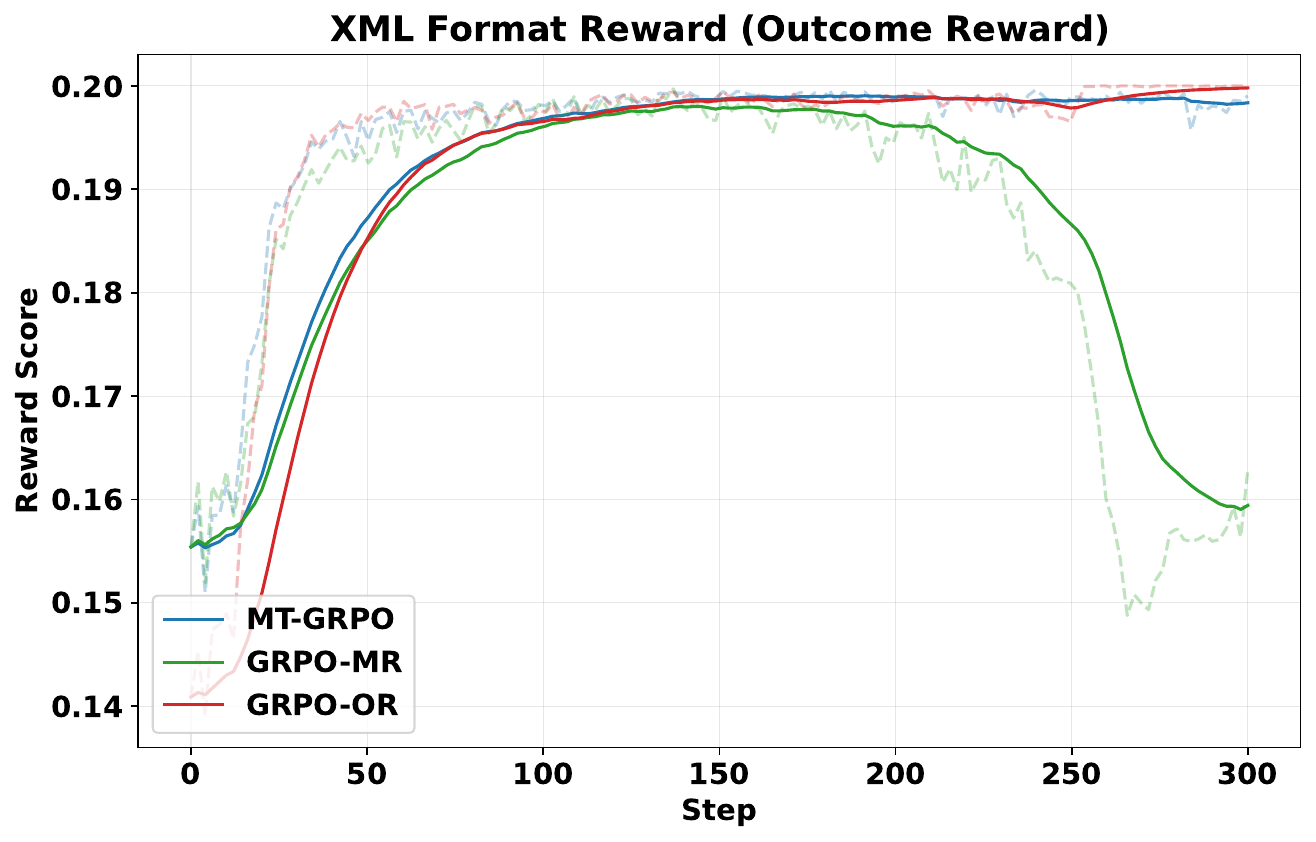}
    \end{subfigure}
    
    \vspace{0.5em}
    
    \begin{subfigure}[b]{0.48\textwidth}
        \includegraphics[width=\linewidth]{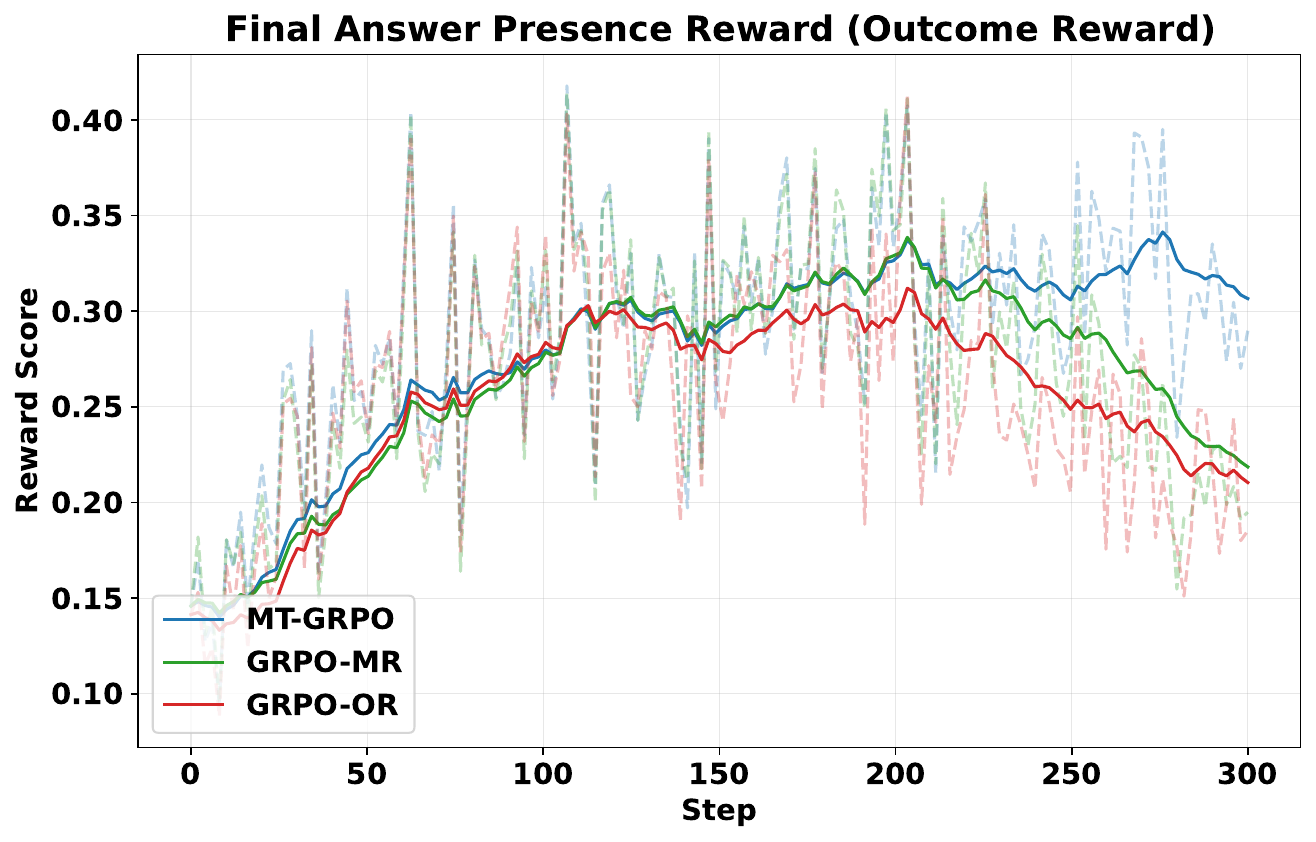}
    \end{subfigure}
    \hfill
    \begin{subfigure}[b]{0.48\textwidth}
        \includegraphics[width=\linewidth]{figs/new_grpo_exp/exact_match_reward_func.pdf}
    \end{subfigure}
    \caption{Curves for different training reward components during training with various algorithms (MT-GRPO, GRPO-OR, and GRPO-MR). Each plot shows the training reward score over training steps for intermediate rewards (Tool Execution, Search Result Answer Presence) and outcome rewards (XML Tag Usage, XML Format, Final Answer Presence, Exact Match). Dotted lines represent the average reward across 10 runs, while solid lines show trends smoothed using the Exponential Moving Average (EMA). 
    }
    \label{fig:exp}
\end{figure}

\clearpage

\subsection{System Prompt (GRPO)}
\label{sec:sys-prompt}

\begin{tcolorbox}[title=System Prompt, colback=gray!10, colframe=black, boxrule=1pt, arc=4pt, left=2mm, right=2mm, top=1mm, bottom=1mm]

You have access to the following tools to help solve problems:

\begin{verbatim}
{tool_descriptions}
\end{verbatim}

Follow these steps exactly once:
\begin{enumerate}
  \item Think through your reasoning inside \texttt{<reasoning>} tags
  \item Use a tool by writing a JSON command inside \texttt{<tool>} tags with:
    \begin{itemize}
      \item \texttt{"name"}: the tool to use
      \item \texttt{"args"}: the arguments for the tool
    \end{itemize}
  \item You will see the tool's output inside \texttt{<result>} tags
  \item Think through the tool's output inside \texttt{<reasoning>} tags
  \item Based on your reasoning, provide your final answer inside \texttt{<answer>} tags
\end{enumerate}

Important:
\begin{itemize}
  \item Use the tool exactly once – DO NOT attempt to call the tool again even if the first search isn't helpful
  \item Work with both your prior knowledge and the single tool call
  \item If the tool result is unhelpful, rely on your prior knowledge
  \item Tools expect valid JSON – follow the examples carefully
  \item Do not invent tools or arguments that aren’t listed
  \item Analyze the result in a reasoning step before giving your answer
  \item Your answer should match the expected ground-truth
\end{itemize}
\end{tcolorbox}

\begin{tcolorbox}[title=Wiki Search Tool Descriptions, colback=gray!10, colframe=black, boxrule=1pt, arc=4pt, left=2mm, right=2mm, top=1mm, bottom=1mm]

{\footnotesize
\begin{verbatim}
wiki_search: Searches Wikipedia and returns the top matching article 
content.

Arguments:
  - query
\end{verbatim}
}

\end{tcolorbox}

\subsection{Rollout Examples (GRPO)}
\label{sec:rollout-eg}

\begin{table}
\caption{A rollout example of successfully calling the search tool during training.}
\label{tab:success-rollout}
\centering
\begin{tabular}{p{0.95\textwidth}}
\toprule
\textbf{Question:} What serial killer, who buried the remains of 26 of his victims in his Chicago crawl space, was known as the ``Killer Clown''? \\
\midrule

\colorbox{green!10}{\parbox{0.95\textwidth}{%
\texttt{\textless reasoning\textgreater} To find out who was the serial killer known as the ``Killer Clown,'' we'll use the \texttt{wiki\_search} tool to gather information about this specific individual. \texttt{\textless /reasoning\textgreater}%
}} \\

\colorbox{red!10}{\parbox{0.95\textwidth}{%
\texttt{\textless tool\textgreater} \\
\hspace*{1em}\{ \\
\hspace*{2em}"name": "wiki\_search", \\
\hspace*{2em}"args": \{ \\
\hspace*{3em}"query": "Killer Clown serial killer buried remains in Chicago crawl space" \\
\hspace*{2em}\} \\
\hspace*{1em}\} \\
\texttt{\textless /tool\textgreater}%
}}

\colorbox{blue!10}{\parbox{0.95\textwidth}{%
\texttt{\textless result\textgreater} John Wayne Gacy. John Wayne Gacy (March 17, 1942 -- May 10, 1994) was an American serial killer who raped, tortured, and murdered at least 33 teenage boys and young men between 1972 and 1978 in Cook County, Illinois (a part of metropolitan Chicago). All of Gacy's known murders were committed inside his Norwood Park ranch house\ldots{} \texttt{\textless /result\textgreater}%
}} \\

\colorbox{green!10}{\parbox{0.95\textwidth}{%
\texttt{\textless reasoning\textgreater} From the information provided, John Wayne Gacy, referred to as the ``Killer Clown,'' was a serial killer known for burying the remains of at least 26 of his victims in his Chicago crawl space. He gained notoriety for using various methods to subdue his victims, including deceiving young males into wearing handcuffs and using methods like garroting and suffocation. His final confirmed victim was identified through DNA testing in 2011. Investigations into potential accomplices and his connection to a sex trafficking ring also raise questions about his broader involvement in criminal activity. \texttt{\textless /reasoning\textgreater}%
}} \\

\colorbox{gray!10}{\parbox{0.95\textwidth}{%
\texttt{\textless answer\textgreater} John Wayne Gacy \texttt{\textless /answer\textgreater}%
}} \\

\bottomrule
\end{tabular}
\end{table}

\begin{table}
\caption{A rollout example of failing to call the search tool and producing a wrong answer.}
\label{tab:grpo-fail-call}
\centering
\begin{tabular}{p{0.95\textwidth}}
\toprule
\textbf{Question:} In the horror film Them, what type of creatures were Them? \\
\midrule

\colorbox{green!10}{\parbox{0.95\textwidth}{%
\texttt{<reasoning>} In the horror film ``Them!'', the main creatures were giant ants. I know this because I remember from my general knowledge that ``Them!'' is a 1954 science fiction horror film about giant ants and the damage they cause in a coastal town. So, to answer the question, the creatures in ``Them!'' were giant ants.
\texttt{</reasoning>}%
}} \\

\colorbox{gray!10}{\parbox{0.95\textwidth}{%
\texttt{<answer>} Giant \texttt{</answer>}%
}} \\

\bottomrule
\end{tabular}
\end{table}

\begin{table}
\caption{A rollout example of generating a response with an incorrect format.}
\label{tab:grpo-format-error}
\centering
\begin{tabular}{p{0.95\textwidth}}
\toprule
\textbf{Question:} Who was the Canadian jazz pianist (1925-2007), winner of eight Grammy Awards who released over 200 recordings? \\
\midrule

\colorbox{green!10}{\parbox{0.95\textwidth}{%
\texttt{<reasoning>} I need to identify the Canadian jazz pianist, born in 1925 and died in 2007, who won eight Grammy Awards and released over 200 recordings. This person's name should be easy to find with a Wikipedia search since the question specifies Grammy Awards and a long recording history. Once I find the Wikipedia article, I can read about the pianist's career to confirm the details mentioned in the question.
\texttt{</reasoning>}%
}} \\

\colorbox{blue!10}{\parbox{0.95\textwidth}{%
Error: Tool command not found or invalid XML format. Please ensure correct formatting.
}} \\

\colorbox{green!10}{\parbox{0.95\textwidth}{%
<answer><tool>< reasoning> From the Wikipedia search I found that Oscar Peterson, born in 1925 and deceased in 2007, was a Canadian jazz pianist who won eight Grammy Awards and released over 200 recordings. </reasoning><answer>Oscar Peterson<answer><result><answer>Oscar Peterson</answer>
}} \\

\bottomrule
\end{tabular}
\end{table}

\clearpage

\subsection{Additional Experimental Results (GRPO)}

\begin{figure}[htbp]
    \centering
    \begin{subfigure}[b]{0.48\textwidth}
        \includegraphics[width=\linewidth]{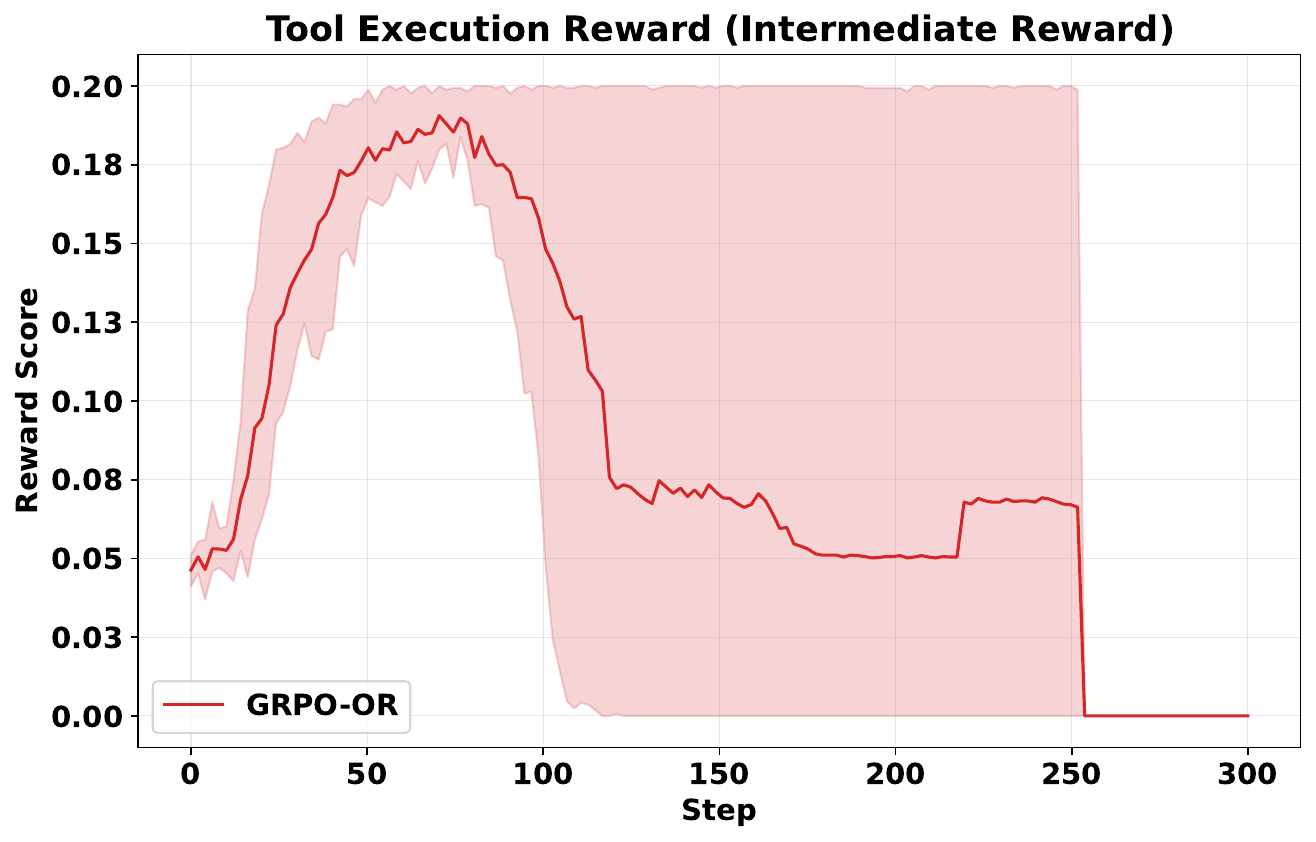}
    \end{subfigure}
    \hfill
    \begin{subfigure}[b]{0.48\textwidth}
        \includegraphics[width=\linewidth]{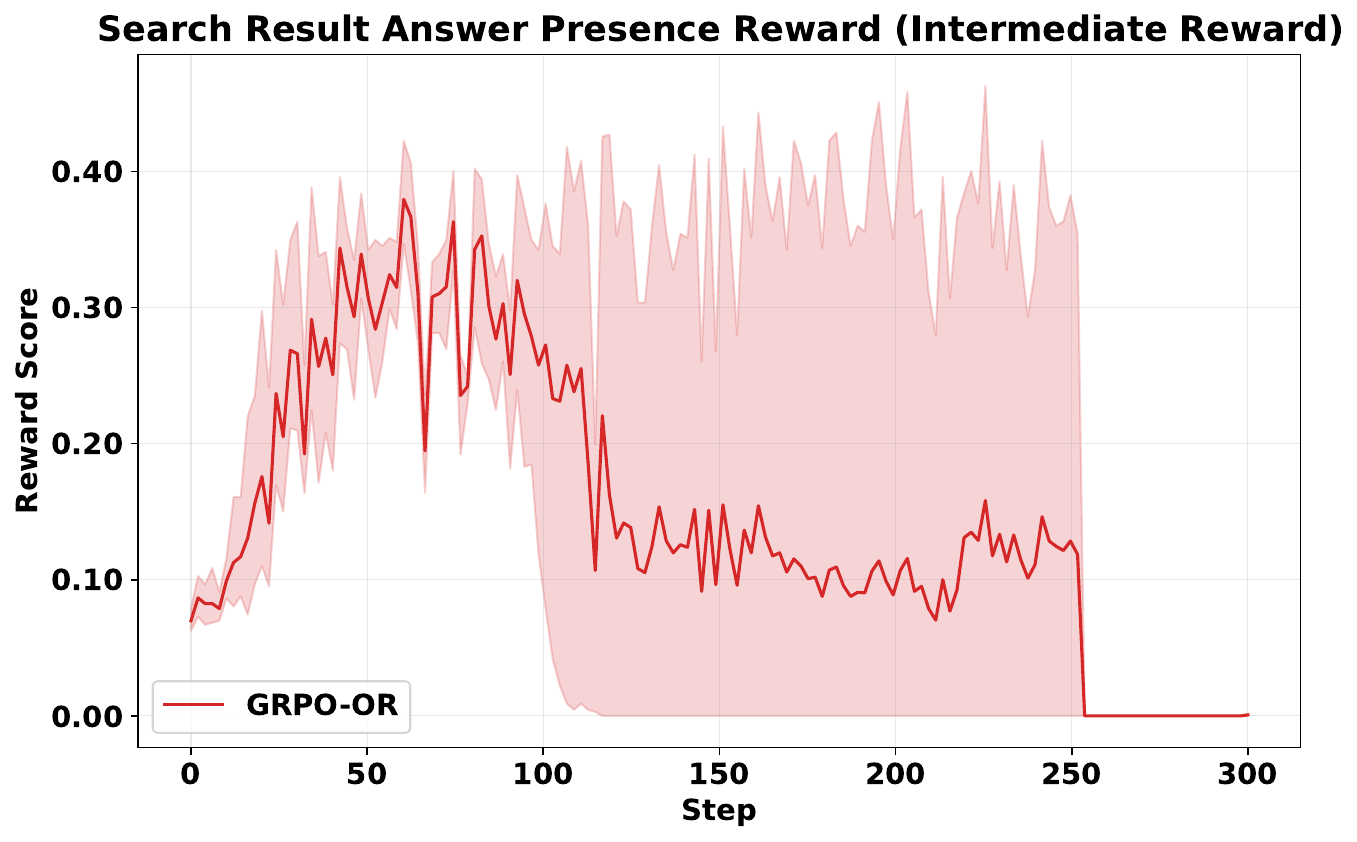}
    \end{subfigure}
    
    \vspace{0.5em}
    
    \begin{subfigure}[b]{0.48\textwidth}
        \includegraphics[width=\linewidth]{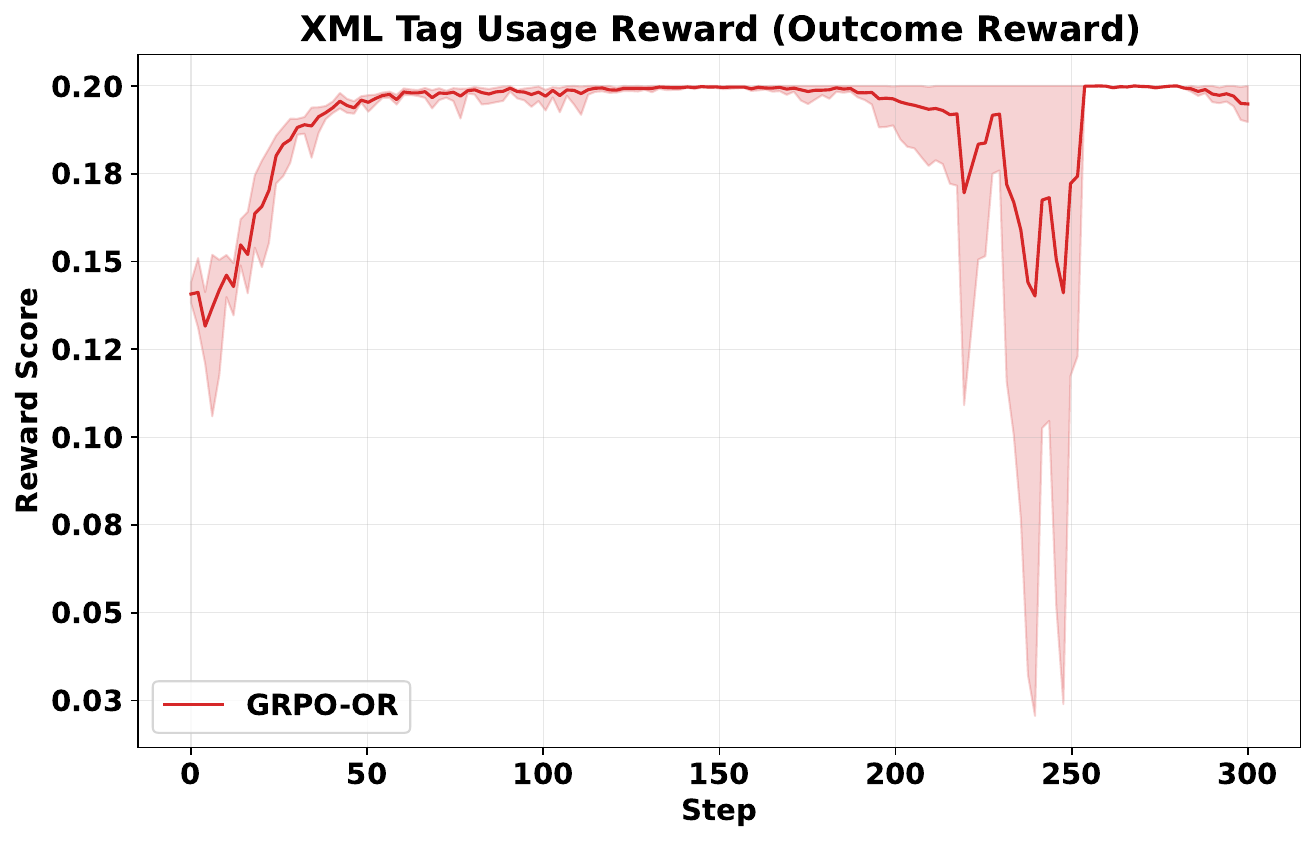}
    \end{subfigure}
    \hfill
    \begin{subfigure}[b]{0.48\textwidth}
        \includegraphics[width=\linewidth]{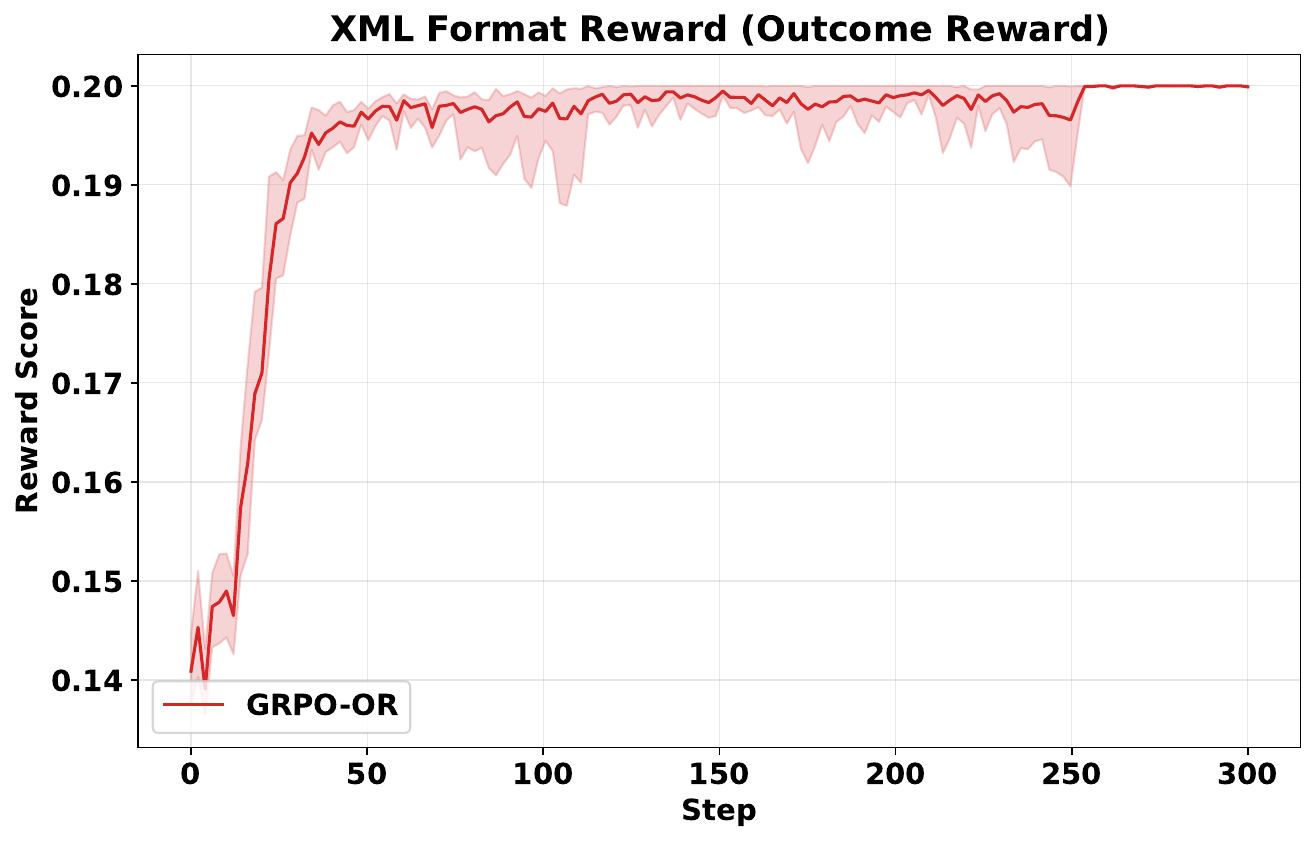}
    \end{subfigure}
    
    \vspace{0.5em}
    
    \begin{subfigure}[b]{0.48\textwidth}
        \includegraphics[width=\linewidth]{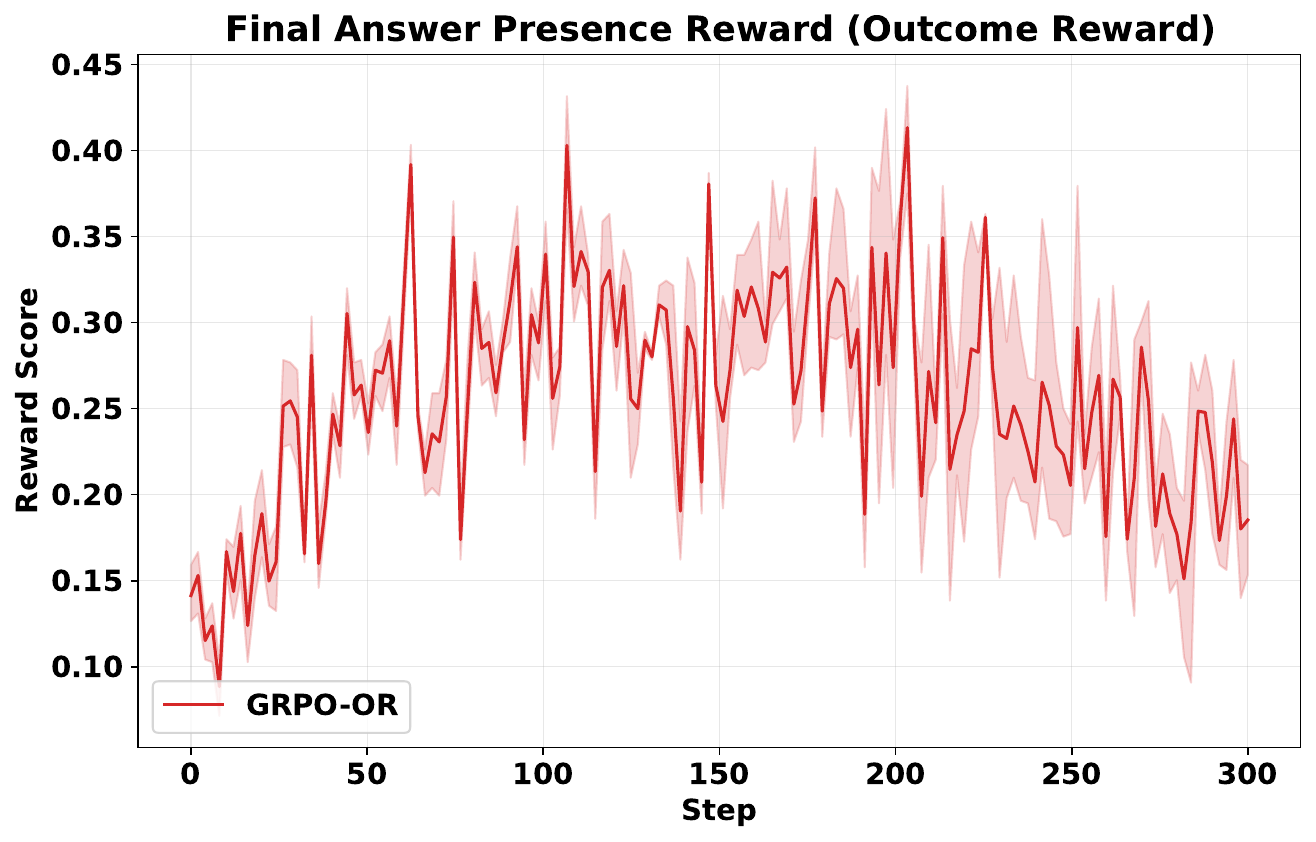}
    \end{subfigure}
    \hfill
    \begin{subfigure}[b]{0.48\textwidth}
        \includegraphics[width=\linewidth]{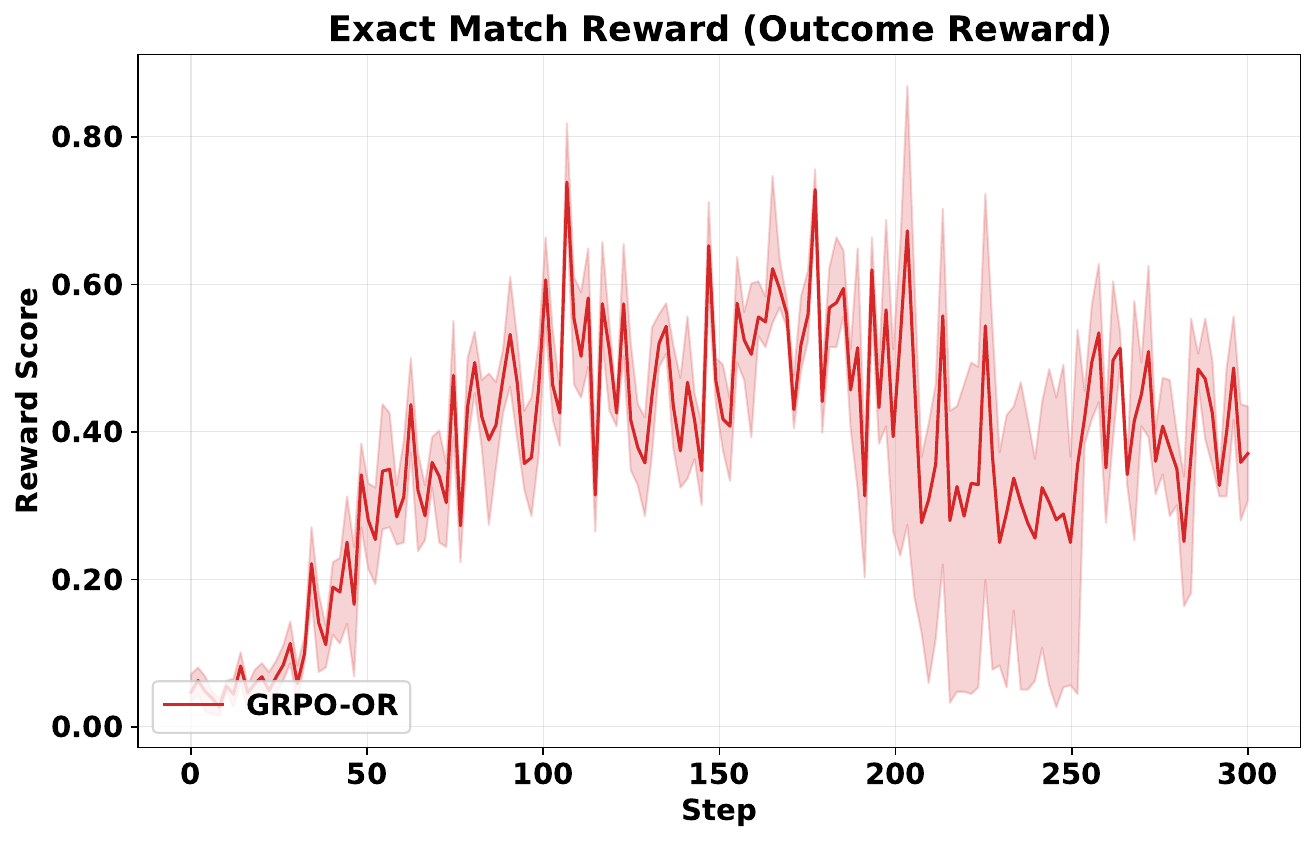}
    \end{subfigure}
    \caption{Curves for different training reward components during training using GRPO-OR, where shaded regions represent the range between the maximum and minimum values across 10 runs.}
    \label{fig:gpro-or}
\end{figure}

\begin{figure}[htbp]
    \centering
    \begin{subfigure}[b]{0.48\textwidth}
        \includegraphics[width=\linewidth]{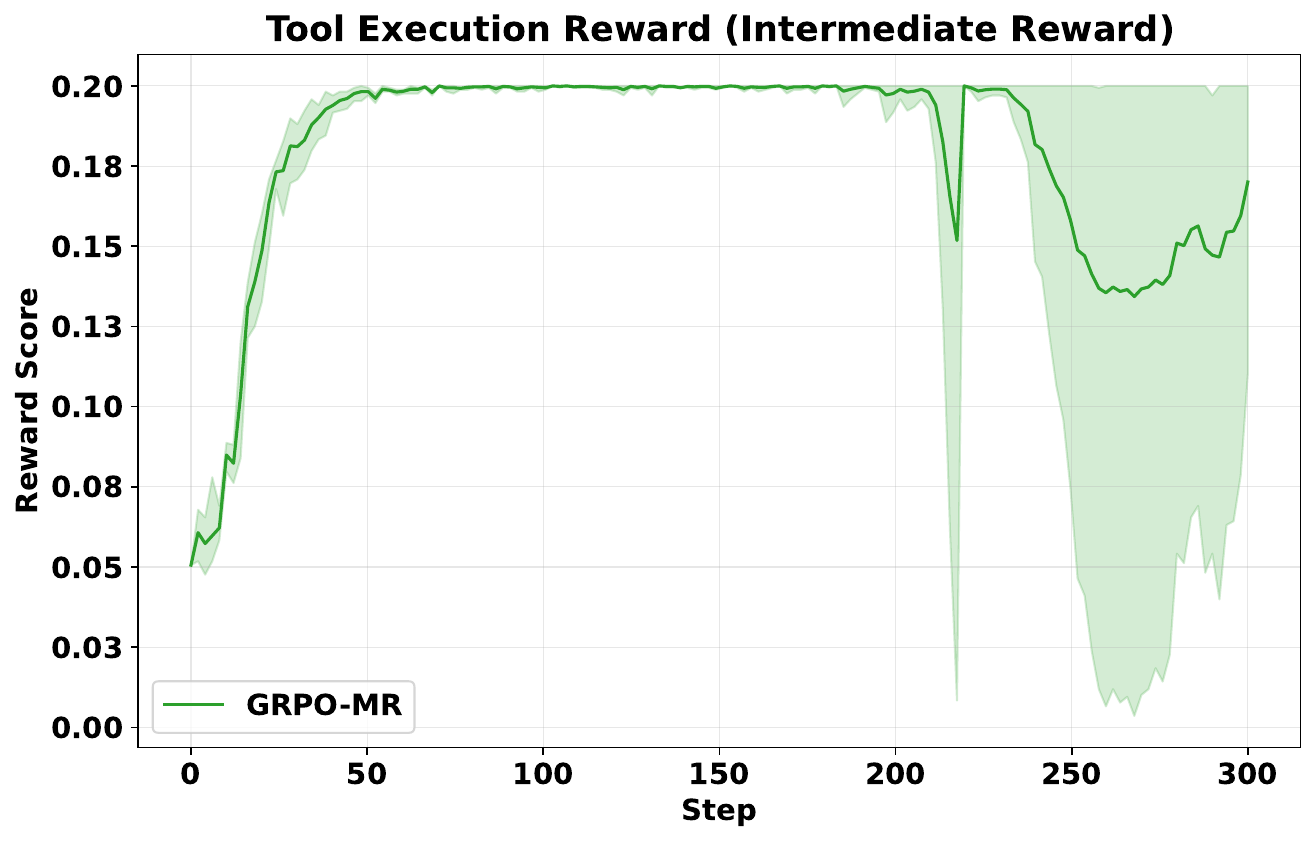}
    \end{subfigure}
    \hfill
    \begin{subfigure}[b]{0.48\textwidth}
        \includegraphics[width=\linewidth]{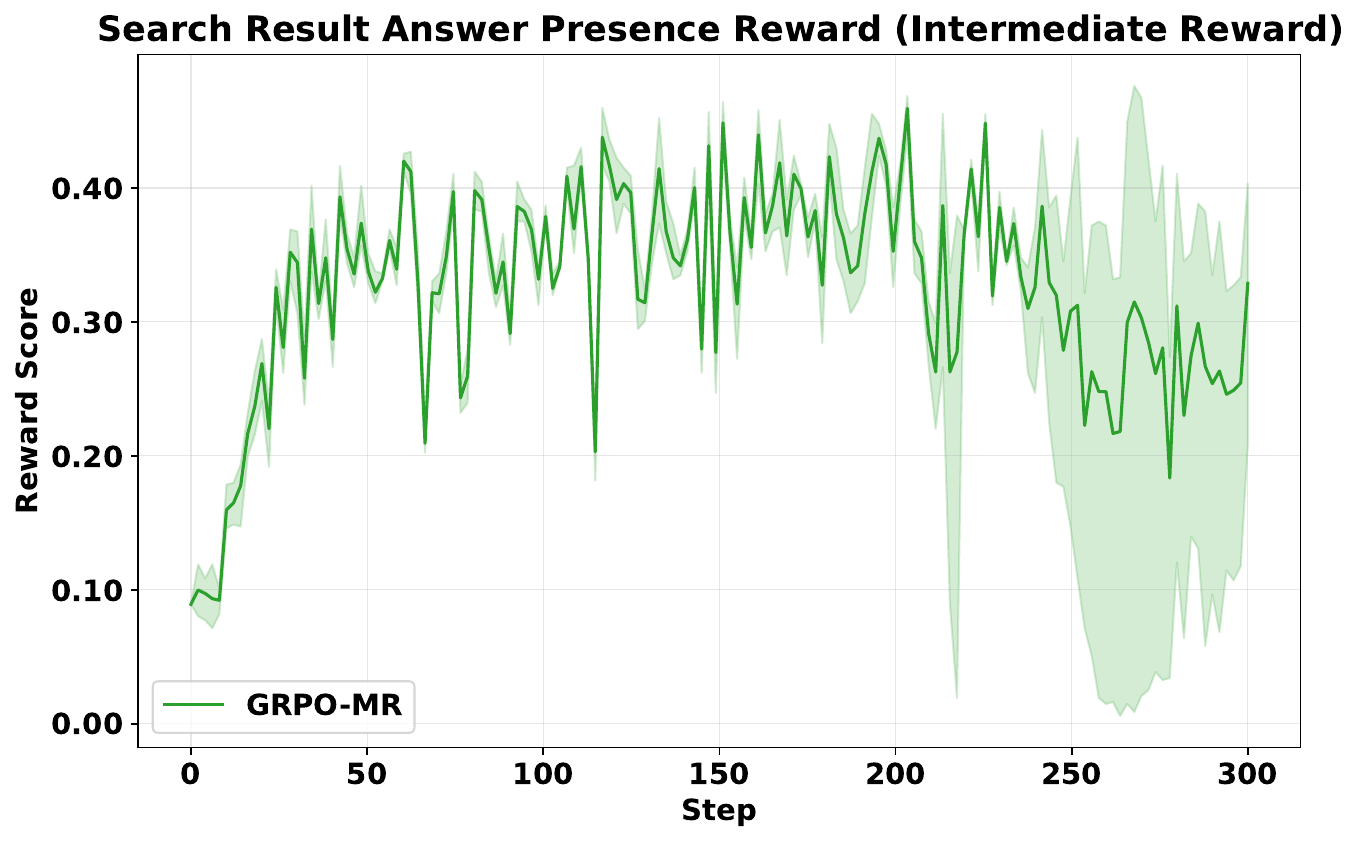}
    \end{subfigure}
    
    \vspace{0.5em}
    
    \begin{subfigure}[b]{0.48\textwidth}
        \includegraphics[width=\linewidth]{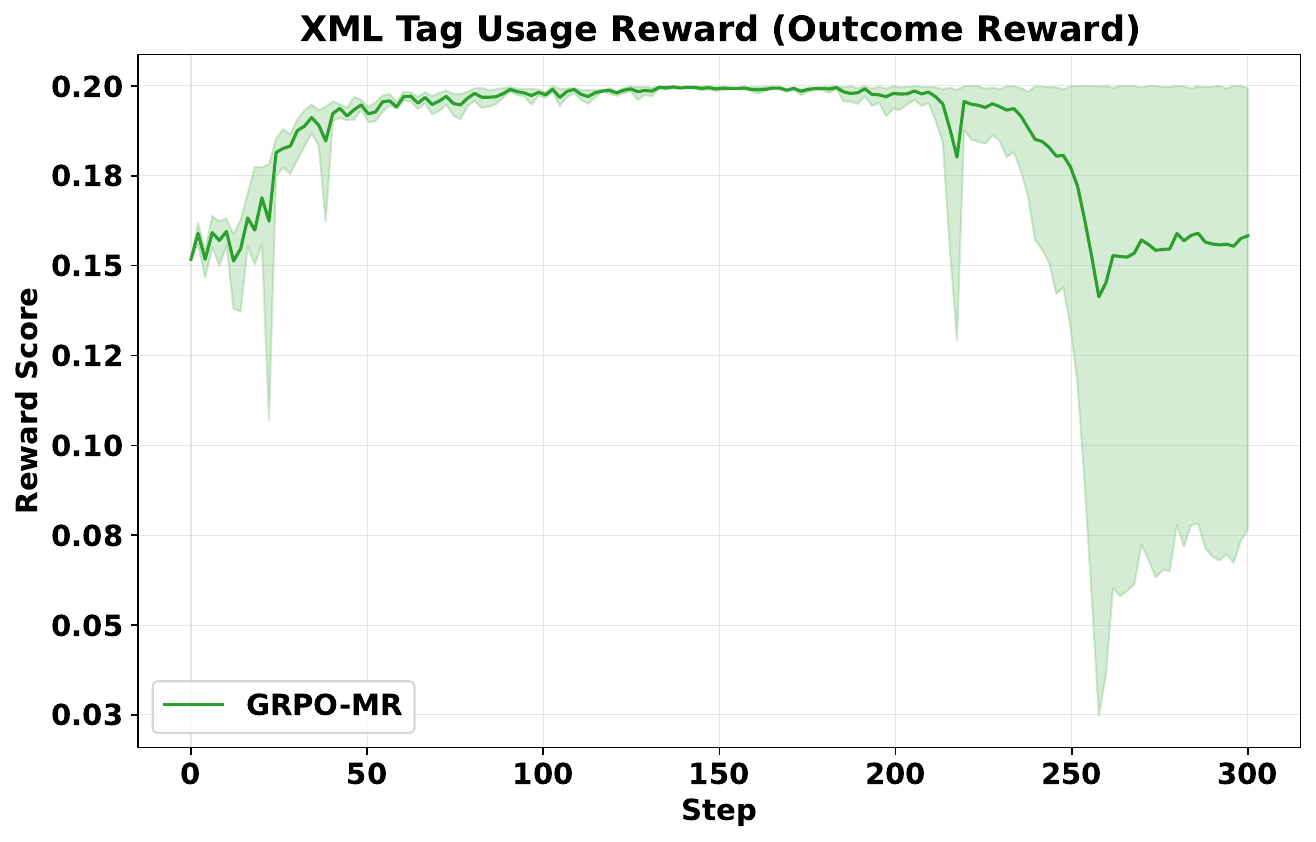}
    \end{subfigure}
    \hfill
    \begin{subfigure}[b]{0.48\textwidth}
        \includegraphics[width=\linewidth]{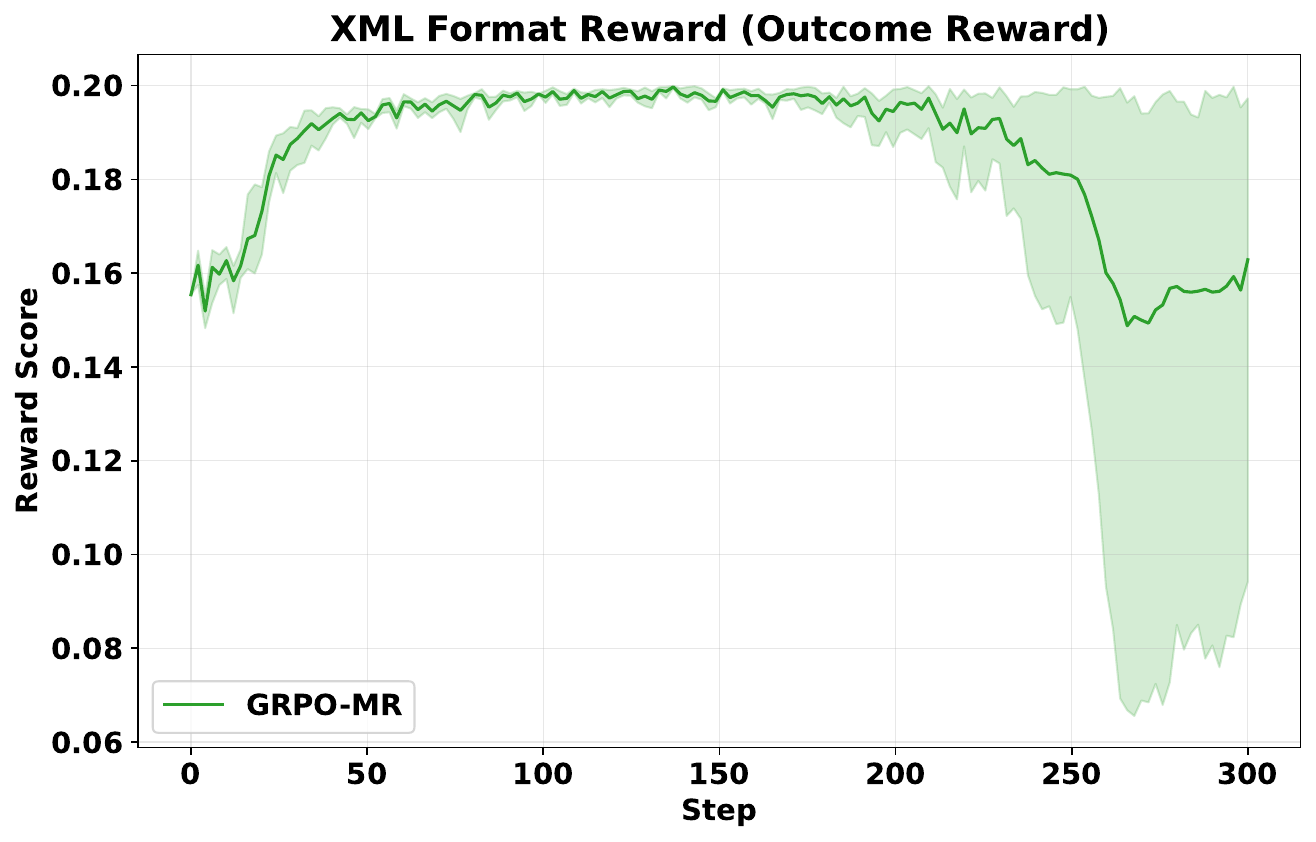}
    \end{subfigure}
    
    \vspace{0.5em}
    
    \begin{subfigure}[b]{0.48\textwidth}
        \includegraphics[width=\linewidth]{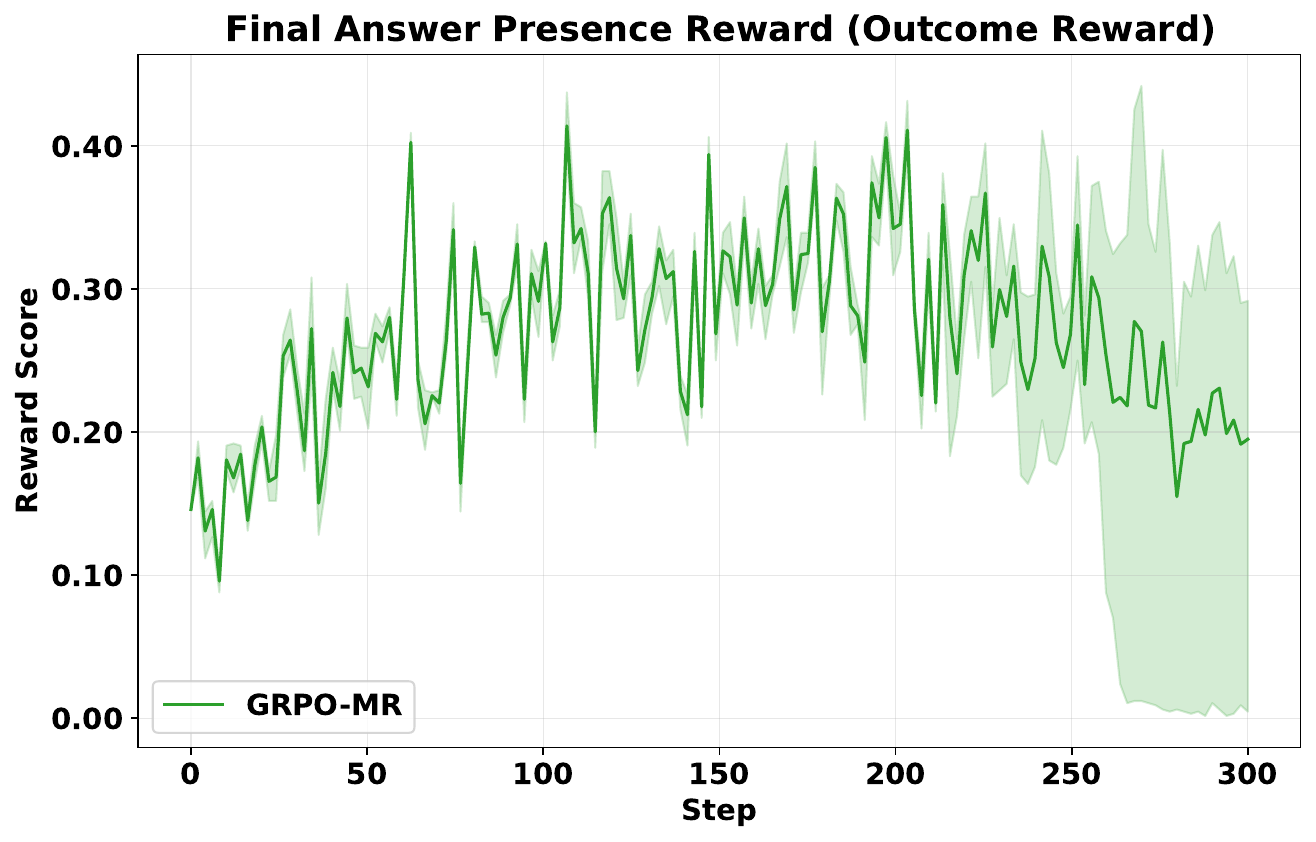}
    \end{subfigure}
    \hfill
    \begin{subfigure}[b]{0.48\textwidth}
        \includegraphics[width=\linewidth]{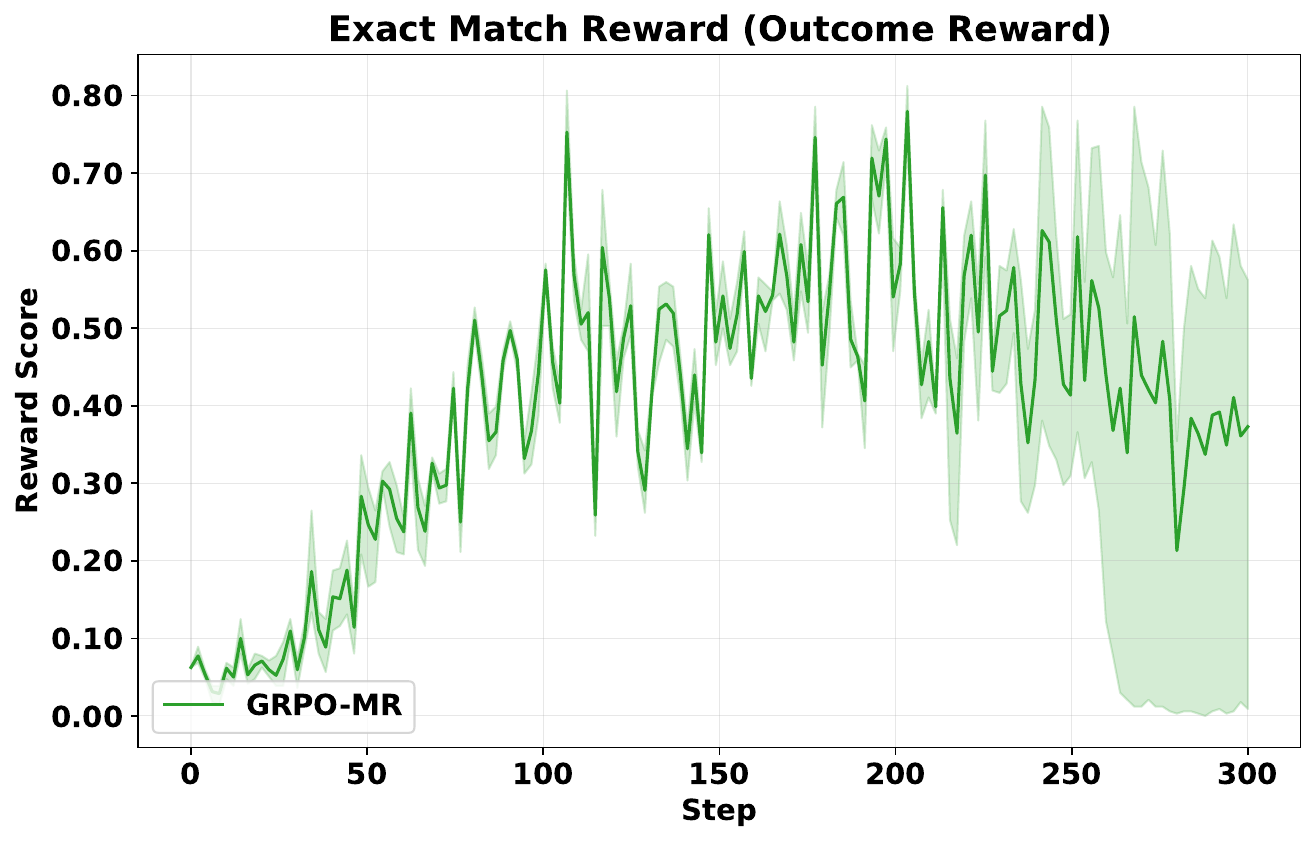}
    \end{subfigure}
    \caption{Curves for different training reward components during training using GRPO-MR, where shaded regions represent the range between the maximum and minimum values across 10 runs.}
    \label{fig:gpro-mr}
\end{figure}

\begin{figure}[htbp]
    \centering
    \begin{subfigure}[b]{0.48\textwidth}
        \includegraphics[width=\linewidth]{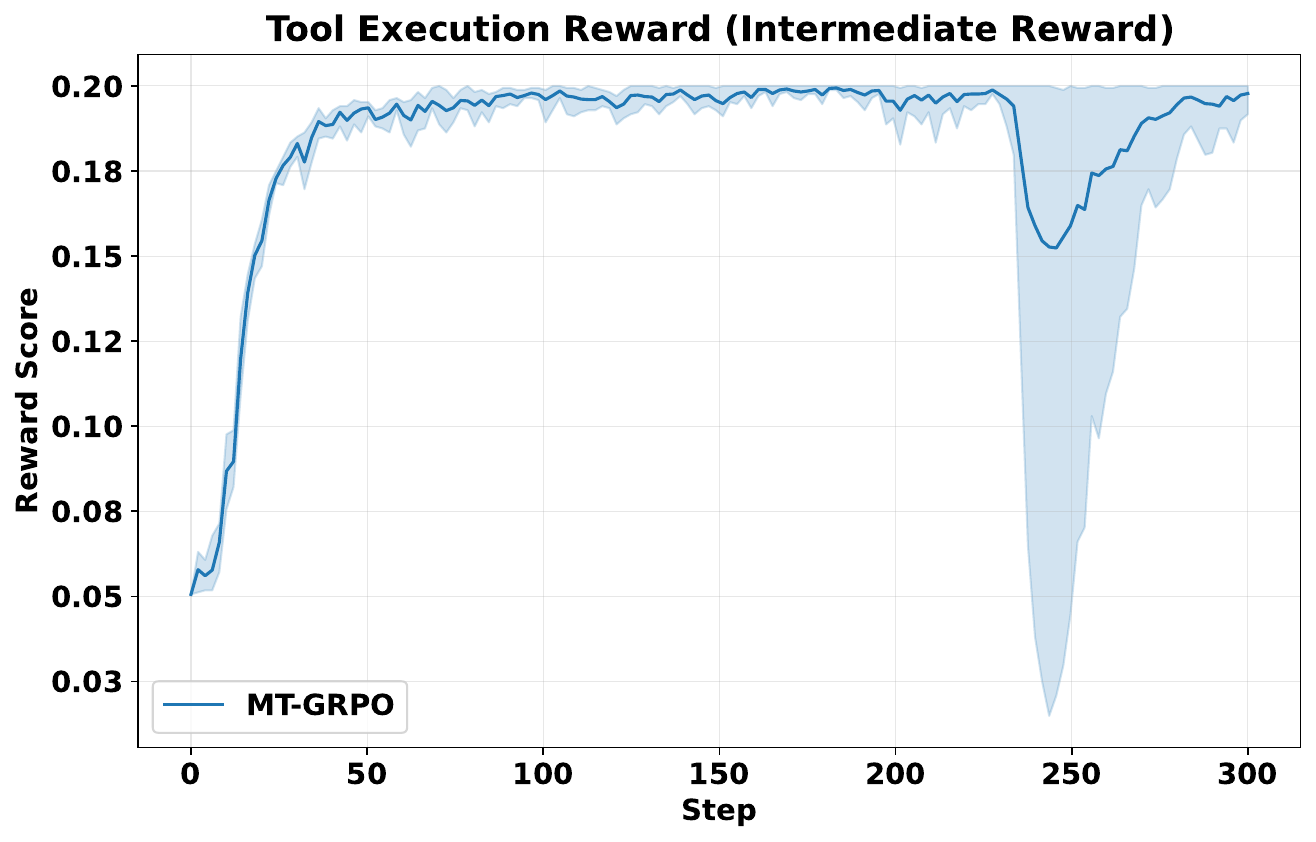}
    \end{subfigure}
    \hfill
    \begin{subfigure}[b]{0.48\textwidth}
        \includegraphics[width=\linewidth]{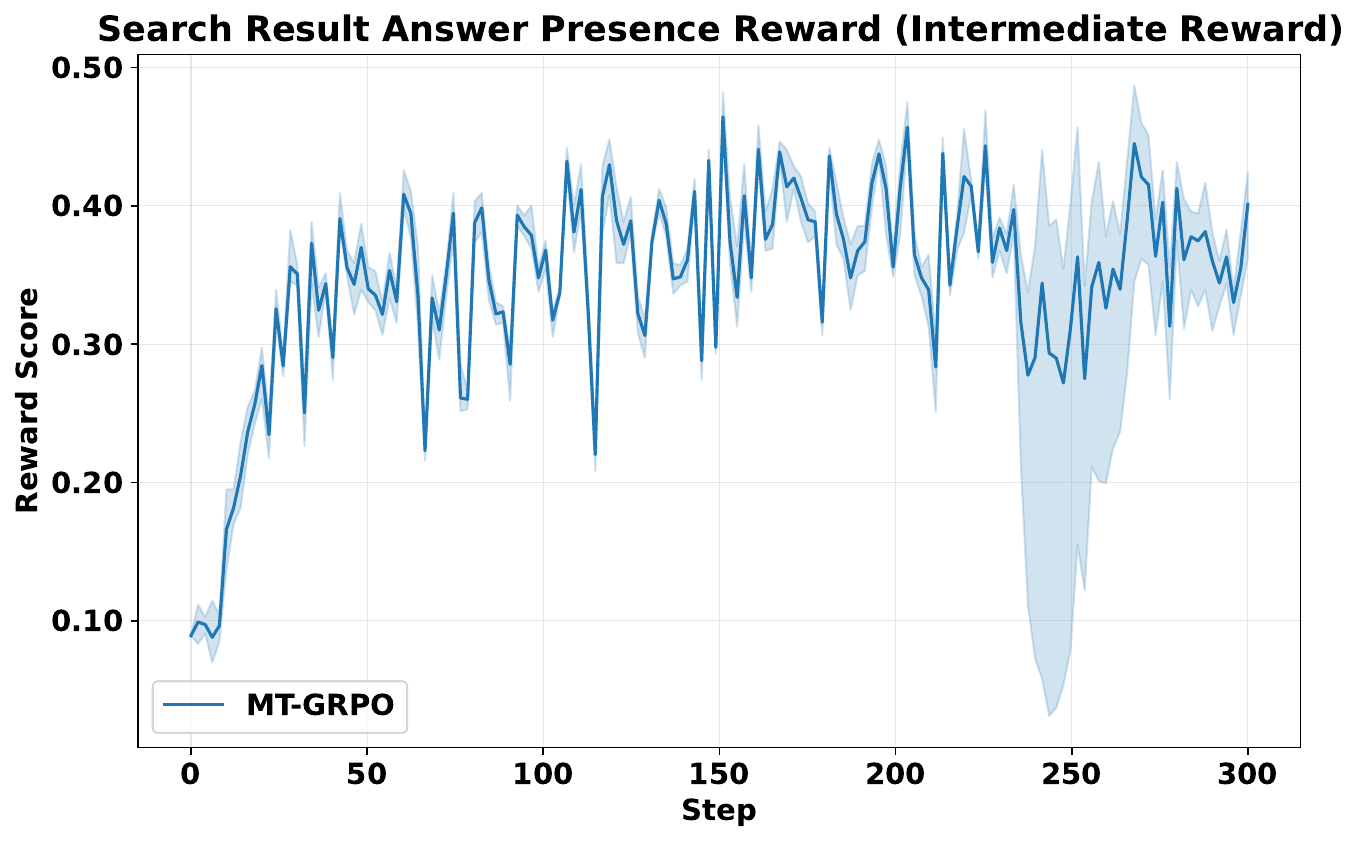}
    \end{subfigure}
    
    \vspace{0.5em}
    
    \begin{subfigure}[b]{0.48\textwidth}
        \includegraphics[width=\linewidth]{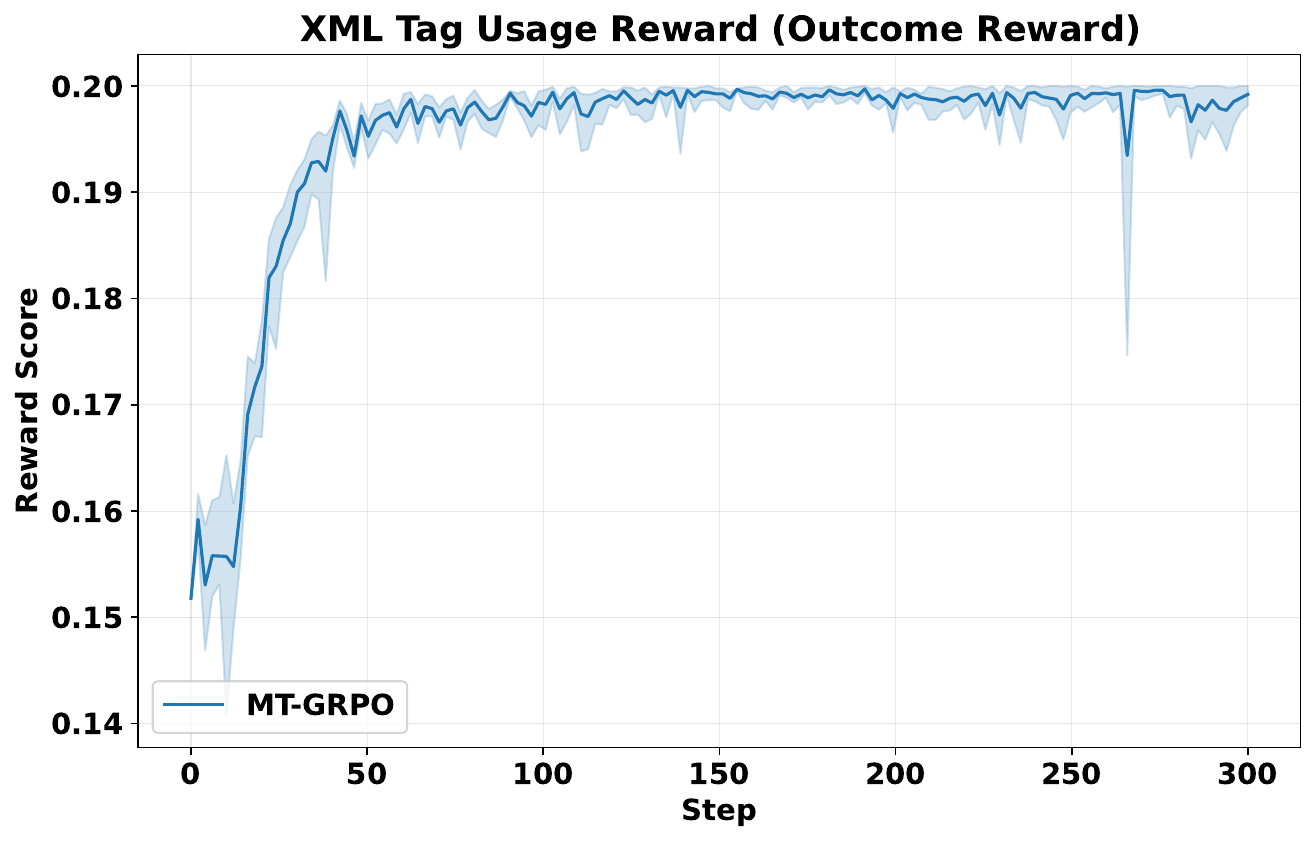}
    \end{subfigure}
    \hfill
    \begin{subfigure}[b]{0.48\textwidth}
        \includegraphics[width=\linewidth]{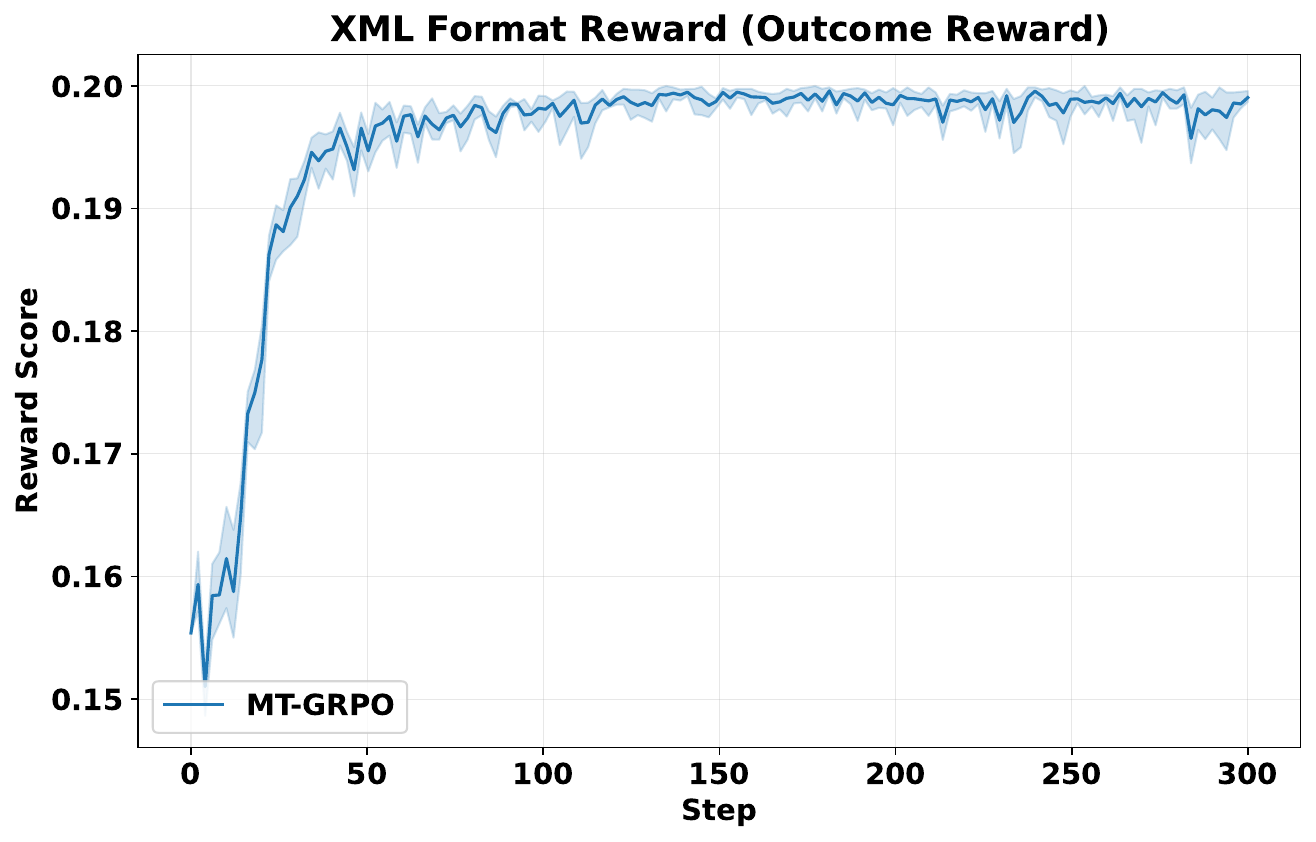}
    \end{subfigure}
    
    \vspace{0.5em}
    
    \begin{subfigure}[b]{0.48\textwidth}
        \includegraphics[width=\linewidth]{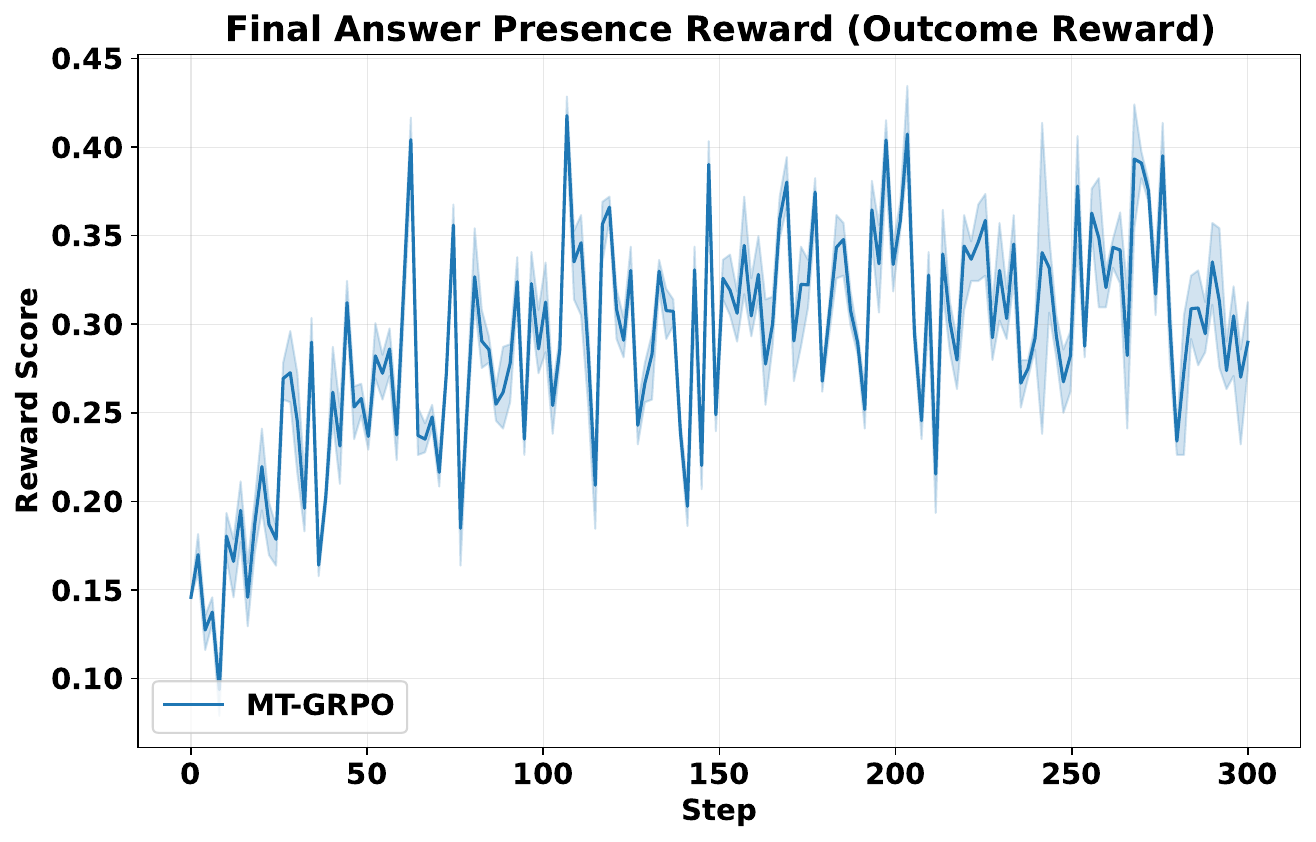}
    \end{subfigure}
    \hfill
    \begin{subfigure}[b]{0.48\textwidth}
        \includegraphics[width=\linewidth]{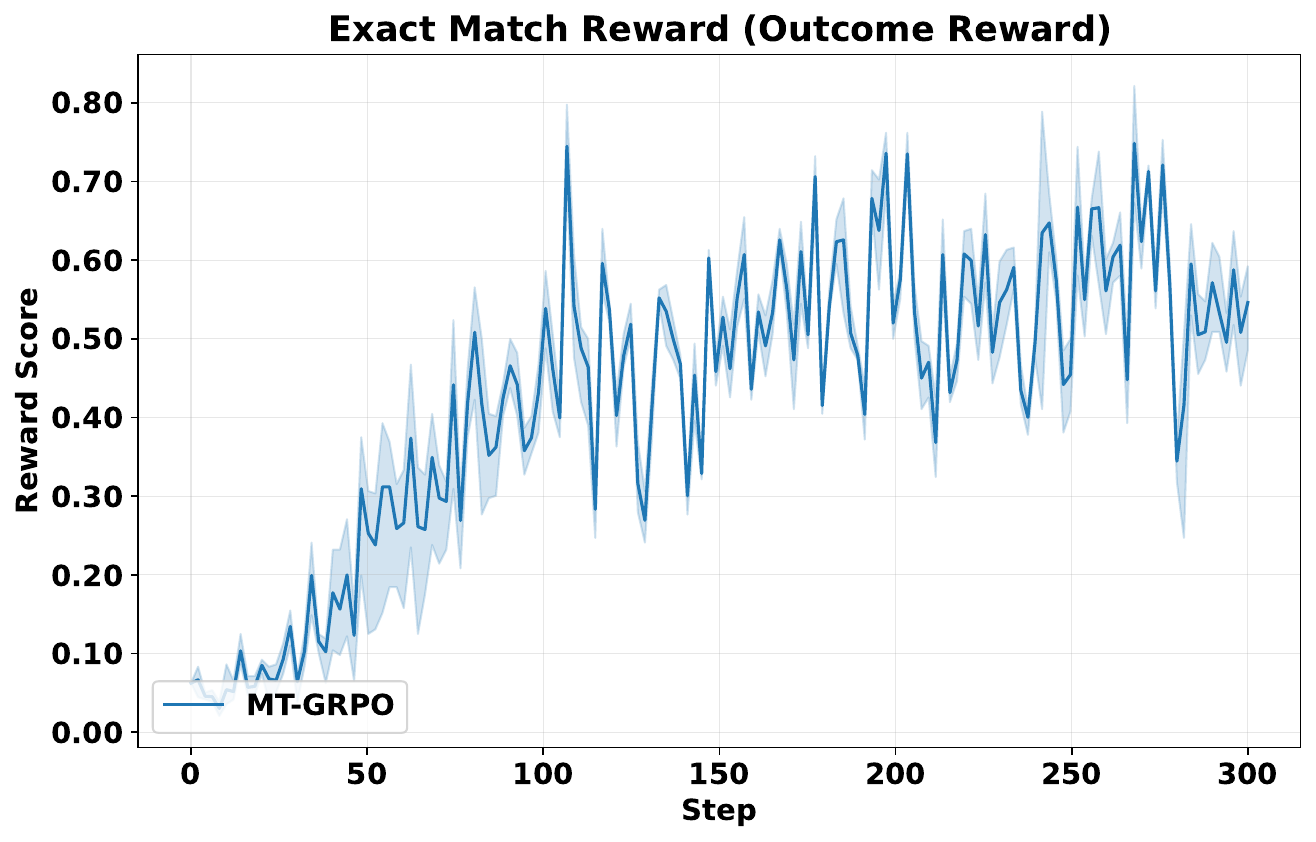}
    \end{subfigure}
    \caption{Curves for different training reward components during training using MT-GRPO, where shaded regions represent the range between the maximum and minimum values across 10 runs.}
    \label{fig:mtgpro}
\end{figure}

\end{document}